\pdfoutput=1
\documentclass{article}

\PassOptionsToPackage{numbers,compress}{natbib}

\usepackage[final,main]{neurips_2026}

\usepackage[utf8]{inputenc}
\usepackage[T1]{fontenc}
\usepackage{url}
\usepackage{booktabs}
\usepackage{amsmath}
\usepackage{amsfonts}
\usepackage{nicefrac}
\usepackage{microtype}
\usepackage{xcolor}

\usepackage{graphicx}
\usepackage{tikz}
\usepackage{tikz-3dplot}
\usetikzlibrary{arrows.meta, positioning, shapes.symbols, calc}
\usepackage{pgfplots}
\pgfplotsset{compat=1.18}
\usepgfplotslibrary{statistics}
\usepackage{subcaption}
\usepackage{trimclip}
\usepackage{array}
\usepackage{enumitem}
\usepackage{multirow}
\usepackage{wrapfig}
\usepackage{pifont}
\usepackage{colortbl}

\pgfdeclarelayer{background}
\pgfsetlayers{background,main}

\definecolor{lavblue}{RGB}{204,204,255}
\definecolor{ltorange}{RGB}{240,247,255}

\definecolor{grayline}{HTML}{B0B0B0}
\definecolor{graytext}{HTML}{505050}
\definecolor{sciblue}{HTML}{4C78A8}
\definecolor{scired}{HTML}{E15759}
\definecolor{scigreen}{HTML}{54A24B}
\definecolor{sciorange}{HTML}{F58518}
\definecolor{scipurple}{HTML}{B279A2}
\definecolor{sciteal}{HTML}{72B7B2}
\definecolor{scigray}{HTML}{7F7F7F}

\usepackage{xspace}

\usepackage[colorlinks=true,
            linkcolor=red,
            citecolor=blue,
            urlcolor=magenta]{hyperref}
\usepackage{cleveref}
\makeatletter
\DeclareRobustCommand\onedot{\futurelet\@let@token\@onedot}
\def\@onedot{\ifx\@let@token.\else.\null\fi\xspace}
\def\eg{\emph{e.g}\onedot}

\def\ie{\emph{i.e}\onedot}

\makeatother

\definecolor{srcblue}{HTML}{4472C4}
\definecolor{transorange}{HTML}{ED7D31}
\definecolor{outgreen}{HTML}{70AD47}
\definecolor{netpurp}{HTML}{7030A0}
\definecolor{mogepurp}{HTML}{7B68EE}

\definecolor{axisred}{HTML}{E88080}
\definecolor{axisgreen}{HTML}{7CC47F}
\definecolor{axisblue}{HTML}{6BA4D9}
\definecolor{camteal}{HTML}{4DBFAF}

\newcommand{\cvComputeRotation}[4]{%
  \pgfmathsetmacro{\cvRaa}{1-2*(#2*#2+#3*#3)}%
  \pgfmathsetmacro{\cvRab}{2*(#1*#2-#3*#4)}%
  \pgfmathsetmacro{\cvRac}{2*(#1*#3+#2*#4)}%
  \pgfmathsetmacro{\cvRba}{2*(#1*#2+#3*#4)}%
  \pgfmathsetmacro{\cvRbb}{1-2*(#1*#1+#3*#3)}%
  \pgfmathsetmacro{\cvRbc}{2*(#2*#3-#1*#4)}%
  \pgfmathsetmacro{\cvRca}{2*(#1*#3-#2*#4)}%
  \pgfmathsetmacro{\cvRcb}{2*(#2*#3+#1*#4)}%
  \pgfmathsetmacro{\cvRcc}{1-2*(#1*#1+#2*#2)}%
}

\newcommand{\cvTransformPoint}[4]{%
  \pgfmathsetmacro{\cvTWx}{\cvPx+\cvRaa*(#1)+\cvRab*(#2)+\cvRac*(#3)}%
  \pgfmathsetmacro{\cvTWy}{\cvPy+\cvRba*(#1)+\cvRbb*(#2)+\cvRbc*(#3)}%
  \pgfmathsetmacro{\cvTWz}{\cvPz+\cvRca*(#1)+\cvRcb*(#2)+\cvRcc*(#3)}%
  \coordinate (#4) at (\cvTWx, \cvTWz, \cvTWy);%
}

\newcommand{\cvComputeVisibility}{%
  \pgfpointanchor{cvFO}{center}\pgfgetlastxy{\cvSxO}{\cvSyO}%
  \pgfpointanchor{cvFBL}{center}\pgfgetlastxy{\cvSxBL}{\cvSyBL}%
  \pgfpointanchor{cvFBR}{center}\pgfgetlastxy{\cvSxBR}{\cvSyBR}%
  \pgfpointanchor{cvFTR}{center}\pgfgetlastxy{\cvSxTR}{\cvSyTR}%
  \pgfpointanchor{cvFTL}{center}\pgfgetlastxy{\cvSxTL}{\cvSyTL}%
  \pgfmathsetmacro{\cvVisBOT}{%
    (\cvSxBL-\cvSxO)*(\cvSyBR-\cvSyO)-(\cvSyBL-\cvSyO)*(\cvSxBR-\cvSxO)}%
  \pgfmathsetmacro{\cvVisRIG}{%
    (\cvSxBR-\cvSxO)*(\cvSyTR-\cvSyO)-(\cvSyBR-\cvSyO)*(\cvSxTR-\cvSxO)}%
  \pgfmathsetmacro{\cvVisTOP}{%
    (\cvSxTR-\cvSxO)*(\cvSyTL-\cvSyO)-(\cvSyTR-\cvSyO)*(\cvSxTL-\cvSxO)}%
  \pgfmathsetmacro{\cvVisLEF}{%
    (\cvSxTL-\cvSxO)*(\cvSyBL-\cvSyO)-(\cvSyTL-\cvSyO)*(\cvSxBL-\cvSxO)}%
  \pgfmathsetmacro{\cvVisFRO}{%
    -((\cvSxBR-\cvSxBL)*(\cvSyTR-\cvSyBL)-(\cvSyBR-\cvSyBL)*(\cvSxTR-\cvSxBL))}%
}

\newcommand{\cvEdge}[5]{%
  \pgfmathparse{int((#4 < 0) * (#5 < 0))}%
  \ifnum\pgfmathresult=1\relax
    \draw[color=#1, semithick, densely dashed, opacity=0.4] (#2)--(#3);%
  \else
    \draw[color=#1, semithick] (#2)--(#3);%
  \fi
}

\newcommand{\drawCamera}[9]{%
  \pgfmathsetmacro{\cvPx}{#1}%
  \pgfmathsetmacro{\cvPy}{#2}%
  \pgfmathsetmacro{\cvPz}{#3}%
  \cvComputeRotation{#4}{#5}{#6}{#7}%
  \pgfmathsetmacro{\cvFw}{0.12*(#9)}%
  \pgfmathsetmacro{\cvFh}{0.09*(#9)}%
  \pgfmathsetmacro{\cvFd}{0.25*(#9)}%
  \pgfmathsetmacro{\cvUh}{0.09*(#9)}%
  \pgfmathsetmacro{\cvUw}{0.08*(#9)}%
  \pgfmathsetmacro{\cvLens}{0.025*(#9)}%
  \cvTransformPoint{0}{0}{0}{cvFO}%
  \cvTransformPoint{-\cvFw}{-\cvFd}{-\cvFh}{cvFBL}%
  \cvTransformPoint{\cvFw}{-\cvFd}{-\cvFh}{cvFBR}%
  \cvTransformPoint{\cvFw}{-\cvFd}{\cvFh}{cvFTR}%
  \cvTransformPoint{-\cvFw}{-\cvFd}{\cvFh}{cvFTL}%
  \cvTransformPoint{-\cvUw}{-\cvFd}{\cvFh}{cvFUA}%
  \cvTransformPoint{\cvUw}{-\cvFd}{\cvFh}{cvFUB}%
  \cvTransformPoint{0}{-\cvFd}{\cvFh+\cvUh}{cvFUC}%
  \cvComputeVisibility
  \fill[#8, opacity=0.08]
    (cvFO)--(cvFBL)--(cvFBR)--cycle;%
  \fill[#8, opacity=0.08]
    (cvFO)--(cvFBR)--(cvFTR)--cycle;%
  \fill[#8, opacity=0.08]
    (cvFO)--(cvFTR)--(cvFTL)--cycle;%
  \fill[#8, opacity=0.08]
    (cvFO)--(cvFTL)--(cvFBL)--cycle;%
  \fill[#8, opacity=0.15]
    (cvFBL)--(cvFBR)--(cvFTR)--(cvFTL)--cycle;%
  \cvEdge{#8}{cvFO}{cvFBL}{\cvVisBOT}{\cvVisLEF}%
  \cvEdge{#8}{cvFO}{cvFBR}{\cvVisBOT}{\cvVisRIG}%
  \cvEdge{#8}{cvFO}{cvFTR}{\cvVisRIG}{\cvVisTOP}%
  \cvEdge{#8}{cvFO}{cvFTL}{\cvVisTOP}{\cvVisLEF}%
  \cvEdge{#8}{cvFBL}{cvFBR}{\cvVisBOT}{\cvVisFRO}%
  \cvEdge{#8}{cvFBR}{cvFTR}{\cvVisRIG}{\cvVisFRO}%
  \cvEdge{#8}{cvFTR}{cvFTL}{\cvVisTOP}{\cvVisFRO}%
  \cvEdge{#8}{cvFTL}{cvFBL}{\cvVisLEF}{\cvVisFRO}%
  \fill[#8] (cvFO) circle[radius=\cvLens];%
  \fill[color=#8]
    (cvFUA)--(cvFUB)--(cvFUC)--cycle;%
}

\newcommand{\drawCameraAxes}[8]{%
  \pgfmathsetmacro{\cvPx}{#1}%
  \pgfmathsetmacro{\cvPy}{#2}%
  \pgfmathsetmacro{\cvPz}{#3}%
  \cvComputeRotation{#4}{#5}{#6}{#7}%
  \cvTransformPoint{0}{0}{0}{cvAO}%
  \cvTransformPoint{#8}{0}{0}{cvAX}%
  \cvTransformPoint{0}{-#8}{0}{cvAF}%
  \cvTransformPoint{0}{0}{-#8}{cvAZ}%
  \draw[->, axisred, very thick] (cvAO)--(cvAX);%
  \draw[->, axisgreen, very thick] (cvAO)--(cvAZ);%
  \draw[->, axisblue, very thick] (cvAO)--(cvAF);%
}

\newcommand{\drawImagePlane}[9]{%
  \pgfmathsetmacro{\cvPx}{#1}%
  \pgfmathsetmacro{\cvPy}{#2}%
  \pgfmathsetmacro{\cvPz}{#3}%
  \cvComputeRotation{#4}{#5}{#6}{#7}%
  \pgfmathsetmacro{\cvIHw}{(#8)/2}%
  \pgfmathsetmacro{\cvIHh}{(#9)/2}%
  \cvTransformPoint{-\cvIHw}{-\cvIHh}{0}{cvIBL}%
  \cvTransformPoint{\cvIHw}{-\cvIHh}{0}{cvIBR}%
  \cvTransformPoint{-\cvIHw}{\cvIHh}{0}{cvITL}%
  \cvTransformPoint{\cvIHw}{\cvIHh}{0}{cvITR}%
  \cvDrawImageContent%
}
\newcommand{\cvDrawImageContent}[1]{%
  \begin{scope}
    \pgftransformreset
    \pgftransformtriangle
      {\pgfpointanchor{cvIBL}{center}}
      {\pgfpointanchor{cvIBR}{center}}
      {\pgfpointanchor{cvITL}{center}}
    \pgflowlevelsynccm
    \node[anchor=south west, inner sep=0pt] at (0,0)
      {\includegraphics[width=1pt, height=1pt]{#1}};
  \end{scope}%
  \path (cvIBL)--(cvIBR)--(cvITR)--(cvITL)--cycle;%
}

\newcommand{\drawGroundGrid}[5]{%
  \pgfmathsetmacro{\cvGxs}{#1}%
  \pgfmathsetmacro{\cvGxe}{#2}%
  \pgfmathsetmacro{\cvGzs}{#3}%
  \pgfmathsetmacro{\cvGze}{#4}%
  \pgfmathsetmacro{\cvGst}{#5}%
  \pgfmathsetmacro{\cvGxn}{\cvGxs+\cvGst}%
  \pgfmathsetmacro{\cvGzn}{\cvGzs+\cvGst}%
  \foreach \i in {\cvGxs,\cvGxn,...,\cvGxe}{%
    \draw[gray!40, thin] (\i, \cvGzs, 0) -- (\i, \cvGze, 0);%
  }%
  \foreach \j in {\cvGzs,\cvGzn,...,\cvGze}{%
    \draw[gray!40, thin] (\cvGxs, \j, 0) -- (\cvGxe, \j, 0);%
  }%
}

\newcommand{\drawInfiniteGroundGrid}[1]{%
  \coordinate (cvBBsw) at (current bounding box.south west);%
  \coordinate (cvBBne) at (current bounding box.north east);%
  \begin{pgfonlayer}{background}
    \begin{scope}
      \clip (cvBBsw) rectangle (cvBBne);%
      \drawGroundGrid{-15}{15}{-15}{15}{#1}%
    \end{scope}
  \end{pgfonlayer}%
}

\newcommand{\setRenderingCamera}[6]{%
  \tdplotsetmaincoords{#4}{#5}%
  \def\cvRenderCamTx{#1}%
  \def\cvRenderCamTy{#2}%
  \def\cvRenderCamTz{#3}%
  \def\cvRenderCamRoll{#6}%
}

\def\cvRenderCamTx{0}
\def\cvRenderCamTy{0}
\def\cvRenderCamTz{0}
\def\cvRenderCamRoll{0}

\newenvironment{renderingCameraView}{%
  \begin{scope}
    \pgftransformrotate{\cvRenderCamRoll}%
    \pgftransformshift{\pgfpointxyz
      {-\cvRenderCamTx}{-\cvRenderCamTz}{-\cvRenderCamTy}}%
}{%
  \end{scope}%
}

\newif\ifshowcomments
\showcommentsfalse

\newcommand{\nicolas}[1]{}
\newcommand{\mname}{OVIE\xspace}
\newcommand{\myparagraph}[1]{\vspace{1.5pt}\noindent\emph{#1}}
\newcommand{\up}[1]{\,{\scriptsize\textcolor{green!60!black}{#1}}}
\newcommand{\dn}[1]{\,{\scriptsize\textcolor{red!70!black}{#1}}}

\renewcommand{\acksection}{\section*{Acknowledgments}}

\newcommand{\fitwidth}[1]{%
  \resizebox{\ifdim\width>\linewidth\linewidth\else\width\fi}{!}{#1}}

\title{One View Is Enough: In-the-Wild Monocular Pretraining for Novel View Generation}

\author{%
  Adrien Ramanana Rahary$^{1,2}$ \quad
  Nicolas Dufour$^{1}$ \quad
  Patrick Pérez$^{1}$ \quad
  David Picard$^{2}$ \\[2pt]
  \normalfont $^{1}$Kyutai \qquad $^{2}$LIGM, ENPC, IP Paris, CNRS, UGE \\[2pt]
  \normalfont \texttt{\{adrienrr, nicolas.dufour, patrick\}@kyutai.org} \quad \texttt{david.picard@enpc.fr}
}

\begin{document}
\raggedbottom

\maketitle
\vspace{-8mm}

\newlength{\teaserImgW}\setlength{\teaserImgW}{2.4cm}
\newlength{\teaserGap}\setlength{\teaserGap}{1.5mm}

\newcommand{\teaserImg}[2]{%
  \tikz[baseline=(img.center)]{%
    \node[inner sep=0pt, draw=#1, line width=2.5pt] (img)
      {\includegraphics[width=\teaserImgW]{#2}};%
  }%
}

\newcommand{\teaserCamLabel}[4]{%
  \node[fill=white, fill opacity=0.9, text opacity=1, text=#1,
        font=\scriptsize\bfseries, inner sep=1.5pt, rounded corners=1pt]
    at ([xshift=#3, yshift=#4]cvFO) {#2};%
}

\newcommand{\teaserScene}[2]{%
  \resizebox*{!}{\dimexpr\teaserImgW+1.5pt\relax}{\clipbox{0pt}{%
  \begin{tikzpicture}[tdplot_main_coords, scale=2,
      baseline=(current bounding box.center)]
    \begin{renderingCameraView}
    \drawImagePlane{0}{0}{2.75}{0}{0}{0}{1}{1.5}{1.5}%
      {#1}
    \drawCamera{0}{0}{0}%
      {-0.7071068}{0}{0}{0.7071067}%
      {gray}{1.8}
    \drawCameraAxes{0}{0}{0}%
      {-0.7071068}{0}{0}{0.7071067}%
      {0.3}
    #2
    \path let
      \p1 = ([shift={(-2mm,-2mm)}]current bounding box.south west),
      \p2 = ([shift={(2mm,2mm)}]current bounding box.north east),
      \n{w} = {\x2 - \x1},
      \n{h} = {\y2 - \y1},
      \n{cx} = {(\x1 + \x2) / 2},
      \n{cy} = {(\y1 + \y2) / 2},
      \n{tw} = {max(\n{w}, 1.5 * \n{h})},
      \n{th} = {max(\n{h}, \n{w} / 1.5)}
    in
      coordinate (cvFrameSW) at (\n{cx} - \n{tw}/2, \n{cy} - \n{th}/2)
      coordinate (cvFrameNE) at (\n{cx} + \n{tw}/2, \n{cy} + \n{th}/2);
    \useasboundingbox (cvFrameSW) rectangle (cvFrameNE);
    \draw[gray!60, very thick] (cvFrameSW) rectangle (cvFrameNE);
    \drawInfiniteGroundGrid{0.5}
    \end{renderingCameraView}
  \end{tikzpicture}}}%
}

\begin{figure}[h!]
\centering
\resizebox{0.94\textwidth}{!}{%
\setlength{\tabcolsep}{0pt}%
\begin{tabular}{
  @{}
  >{\small\raggedleft\arraybackslash}m{1cm}
  @{\hspace{\teaserGap}}c
  @{\hspace{\teaserGap}}c
  @{\hspace{\teaserGap}}c
  @{}c
  @{}c
  @{}
}

&
\small Camera
&
\small Source
&
{\small\color{transorange} View 1}
&
{\small\color{mogepurp} View 2}
&
{\small\color{camteal} View 3}
\\[\teaserGap]

Indoor &
\setRenderingCamera{0.3}{-0.2}{1.0}{55}{130}{0}%
\teaserScene{images/teaser/indoor_quali/source}{%
  \drawCamera{1.3405}{0}{1.0555}%
    {-0.6638}{0.2437}{0.2437}{0.6638}{transorange}{1.8}%
  \teaserCamLabel{transorange}{1}{-4pt}{0pt}%
  \drawCamera{0}{-0.85}{-0.0145}%
    {-0.7908}{0}{0}{0.6121}{mogepurp}{1.8}%
  \teaserCamLabel{mogepurp}{2}{-3pt}{0pt}%
  \drawCamera{-0.6147}{0.2429}{0.8402}%
    {-0.7632}{-0.1177}{-0.2146}{0.5980}{camteal}{1.8}%
  \teaserCamLabel{camteal}{3}{8pt}{10pt}%
} &
\teaserImg{gray}{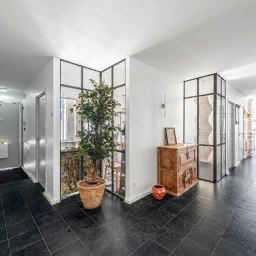} &
\teaserImg{transorange}{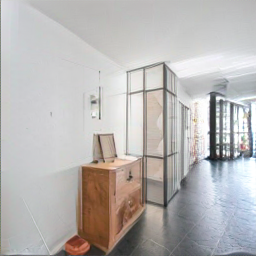} &
\teaserImg{mogepurp}{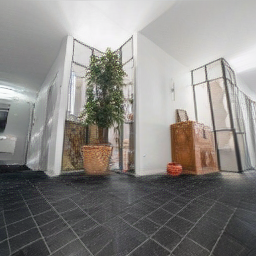} &
\teaserImg{camteal}{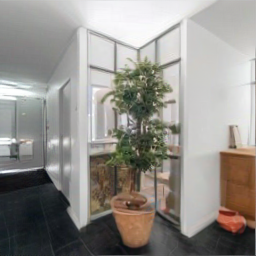}
\\[1mm]

Outdoor &
\setRenderingCamera{0}{0}{1.0}{50}{310}{0}%
\teaserScene{images/teaser/outdoor_quali/source}{%
  \drawCamera{0.0411}{0}{1.8966}%
    {-0.6752}{0.2099}{0.2099}{0.6752}{transorange}{1.8}%
  \teaserCamLabel{transorange}{1}{-5pt}{-8pt}%
  \drawCamera{0.6736}{0.1857}{0.7087}%
    {-0.7381}{-0.2272}{-0.2991}{0.5605}{camteal}{1.8}%
  \teaserCamLabel{camteal}{3}{0pt}{10pt}%
  \drawCamera{0.9875}{0.0209}{-0.1157}%
    {-0.6604}{0.2897}{0.3172}{0.6159}{mogepurp}{1.8}%
  \teaserCamLabel{mogepurp}{2}{10pt}{0pt}%
} &
\teaserImg{gray}{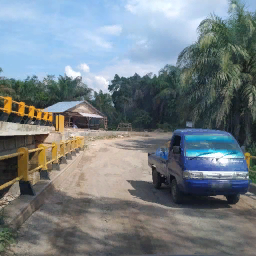} &
\teaserImg{transorange}{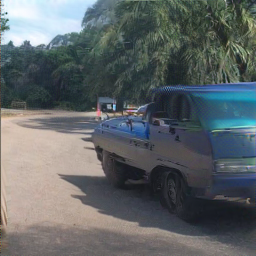} &
\teaserImg{mogepurp}{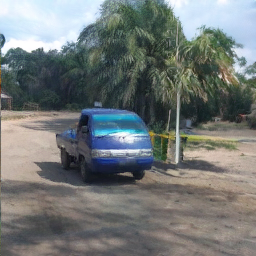} &
\teaserImg{camteal}{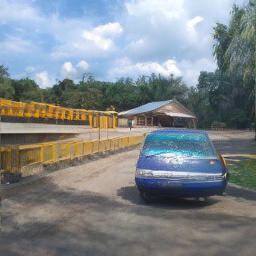}
\\[1mm]

Subject-centric &
\setRenderingCamera{0}{1}{-3}{45}{30}{0}%
\teaserScene{images/teaser/object_quali/source}{%
  \drawCamera{0}{0.027}{1.2355}%
    {-0.8526}{0}{0}{0.5226}{transorange}{1.8}%
  \teaserCamLabel{transorange}{1}{5pt}{-6pt}%
  \drawCamera{-0.699}{0.853}{0.5115}%
    {-0.5394}{0.2914}{0.2501}{0.7494}{mogepurp}{1.8}%
  \teaserCamLabel{mogepurp}{2}{-5pt}{4pt}%
  \drawCamera{-2.33}{1.2}{0.885}%
    {-0.4901}{0.3647}{0.3620}{0.7041}{camteal}{1.8}%
  \teaserCamLabel{camteal}{3}{-5pt}{0pt}%
} &
\teaserImg{gray}{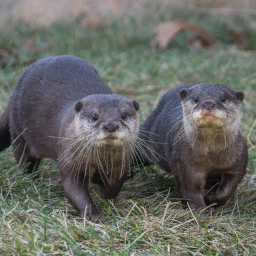} &
\teaserImg{transorange}{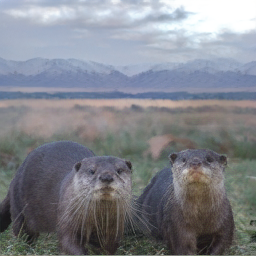} &
\teaserImg{mogepurp}{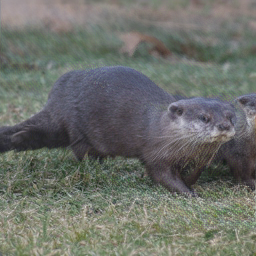} &
\teaserImg{camteal}{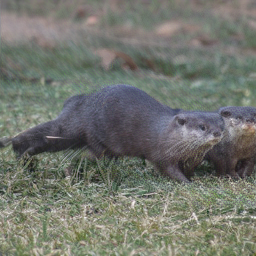}
\\[1mm]

Non-natural &
\setRenderingCamera{0}{0.1}{1.0}{55}{130}{0}%
\teaserScene{images/teaser/art_quali/source}{%
  \drawCamera{-0.9895}{0.27}{0.4765}%
    {-0.8246}{-0.1081}{-0.3193}{0.4543}{transorange}{1.8}%
  \teaserCamLabel{transorange}{1}{-5pt}{0pt}%
  \drawCamera{0}{0.405}{1.0695}%
    {-0.8943}{0}{0}{0.4474}{mogepurp}{1.8}%
  \teaserCamLabel{mogepurp}{2}{0pt}{10pt}%
  \drawCamera{1.3285}{-0.214}{-0.0505}%
    {-0.6328}{0.3458}{0.3694}{0.5861}{camteal}{1.8}%
  \teaserCamLabel{camteal}{3}{8pt}{-8pt}%
} &
\teaserImg{gray}{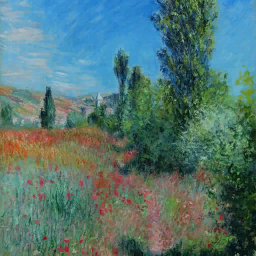} &
\teaserImg{transorange}{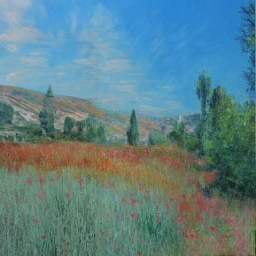} &
\teaserImg{mogepurp}{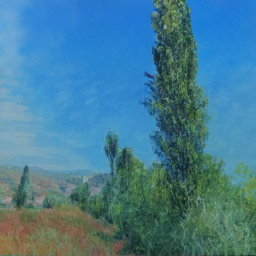} &
\teaserImg{camteal}{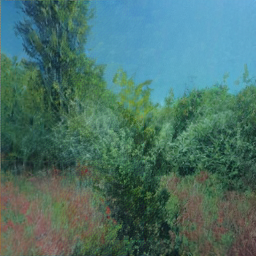}

\end{tabular}%
}%
\caption{\textbf{\mname generates novel views from a single image in real time} across diverse domains
given a source image (gray) and target poses (colored), regardless of content or style.}
\label{fig:teaser}
\end{figure}

\vspace{-1mm}

\begin{abstract}
Monocular novel-view synthesis has long required multi-view image pairs for supervision, limiting training to a narrow set of purpose-built datasets. We propose in-the-wild monocular pretraining: a frozen depth estimator lifts each source image into 3D and reprojects under sampled poses to yield pseudo-target views; masked losses restrict supervision to valid regions and an adversarial objective covers disoccluded areas. Scaled to 30 million uncurated images, this produces OVIE, requiring only a source image and target pose at inference. Prior work trains without multi-view data but needs a depth estimator at inference, or drops this dependency but requires multi-view training pairs; OVIE is the first to require neither. Without multi-view supervision, OVIE rivals in-domain baselines on RealEstate10K and surpasses all on DL3DV, producing the most multi-view-consistent trajectories of any geometry-free method; brief multi-view fine-tuning outperforms all geometry-free methods on their training domain. At 116 FPS, it is over 600$\times$ faster than the fastest baseline. Code and pretrained models are at \url{https://github.com/kyutai-labs/ovie}; video results are on the project page, \url{https://kyutai.org/blog/2026-04-14-ovie/}.
\end{abstract}

\section{Introduction}
\label{sec:intro}
Trillions of photographs document the world, yet monocular novel view synthesis, the task of generating what a scene looks like from a new camera pose given a single image, trains on a tiny fraction of them. The bottleneck is a data-type constraint: the task requires large volumes of video from static environments to extract posed multi-view pairs, a condition most of the visual world cannot meet. Paintings, historical photographs, and any dynamic scene simply have no multi-view counterpart. The community has converged on a handful of purpose-built datasets, such as RealEstate10K~\cite{re10k} and DL3DV~\cite{dl3dv}, which cover only a narrow set of visual domains. Models trained on this data generalize within their training domains but fail under domain shift, and the reason is structural.

The natural approach to sidestep multi-view requirements synthesizes training pairs from single images via depth estimation and reprojection under a sampled pose. Such pseudo-targets omit disoccluded regions, making naive supervision harmful. Methods that take this route, such as Infinite Nature-Zero~\cite{infinitenaturezero}, still require a depth model at inference, while geometry-free models~\cite{geogpt, photonvs, vivid} drop this dependency by training on posed multi-view datasets. No prior work achieves both.

We show this tradeoff can be broken with large-scale monocular pretraining. A frozen monocular depth estimator constructs training pairs from any single image by estimating metric depth, lifting the scene into a point cloud, and reprojecting under a sampled camera pose. A masked training formulation restricts losses to valid reprojected pixels; an adversarial objective covers disoccluded regions. Depth errors are absorbed by boundary masking, an L2 penalty that down-weights residuals, and training scale. At inference, the depth estimator is discarded and the model generates the novel view in a single forward pass (Figure~\ref{fig:teaser}).

Trained on 30 million in-the-wild images from ImageNet-21K~\cite{in21k}, Places~\cite{places}, Open Images~\cite{openimages}, and OSV5M~\cite{osv5m} without any multi-view supervision, \mname already rivals specialized in-domain baselines on RealEstate10K~\cite{re10k} despite never training on it, and surpasses all baselines on DL3DV~\cite{dl3dv} where all methods face the same domain shift. Its generated trajectories are also the most multi-view-consistent of any geometry-free method, ahead of baselines that estimate geometry at inference. A brief fine-tuning stage on RealEstate10K then suffices to surpass all geometry-free methods on pixel fidelity and perceptual similarity, establishing monocular pretraining as an efficient foundation for in-domain adaptation. \mname generates novel views at 116~FPS on a single H100 GPU, more than 600$\times$ faster than the fastest baseline.

Our contributions are threefold:
\begin{itemize}[topsep=0pt]
    \item \emph{We introduce a monocular pretraining paradigm that produces geometry-free models without multi-view supervision.} A masked formulation restricts losses to valid reprojected regions and an adversarial objective covers disoccluded areas, together making depth estimation a training scaffold discarded at inference.
    \item \emph{We show that monocular pretraining is an efficient foundation for in-domain adaptation.} Brief fine-tuning on a single multi-view dataset surpasses all geometry-free in-domain methods, demonstrating effective transfer from large-scale unpaired pretraining.
    \item \emph{We characterize the effect of data scale and diversity on monocular novel view synthesis.} Scale is the primary driver of performance, improving consistently from 3K to 30M training images; diverse mixing provides a complementary gain at fixed budget.
\end{itemize}

\section{Related Work}
\label{sec:related_work}
\myparagraph{Problem setting.}
Novel view synthesis spans tasks differing in input, output, and generalization scope. Per-scene optimization methods such as NeRF~\cite{nerf} and Gaussian Splatting~\cite{gaussiansplatting} fit a scene representation to tens or hundreds of posed views and generalize only within that scene. Multi-view feed-forward methods~\cite{pixelsplat,mvsplat,mvsplat360,lvsm,seva,ibrnet} generalize across scenes but require multiple source images at inference. Feed-forward reconstruction methods~\cite{flash3d,depthsplat,sharp,pixelnerf,splatterimage} predict 3D representations from a single image, yet remain largely object-centric. We address \emph{monocular scene-level novel view synthesis}: given a \emph{single} image and a target pose, synthesize the target view in one feed-forward pass with a model that generalizes across scenes and domains.

\myparagraph{Monocular novel view synthesis with multi-view supervision.}
The dominant paradigm trains feed-forward models on posed multi-view collections, which supply
geometric correspondences as supervision. SynSin~\cite{synsin} established this
paired-data approach; NViST~\cite{nvist} scaled it to MVImgNet~\cite{mvimagenet}. Geometry-free,
pose-conditioned image-to-image models (SRT~\cite{srt}, GeoGPT~\cite{geogpt},
PhotoNVS~\cite{photonvs}, and VIVID~\cite{vivid}) represent the family most related to ours,
synthesizing the target view directly in pixel space without an explicit 3D representation.
All share a structural bottleneck: they require posed, static multi-view datasets (e.g.,
RealEstate10K~\cite{re10k}, DL3DV~\cite{dl3dv}, ScanNet~\cite{scannet}), which cover only a narrow visual domain.
\mname replaces this with monocular pretraining on 30 million
single images from domains these benchmarks do not reach.

\myparagraph{Monocular novel view synthesis without multi-view supervision.}
Learning without posed pairs relies primarily on 3D-aware generative models trained on unposed image collections~\cite{HoloGAN2019,graf,piGAN2021,eg3d,GIRAFFE,get3d}. Applying these to a real input requires test-time GAN inversion~\cite{igan}, which is slow, per-image, and confined to the training category distribution; Ramirez et al.~\cite{unsupervisednv} and G3DR~\cite{g3dr} build novel view synthesis frameworks on this paradigm but remain category-specific. Cycle-consistency approaches~\cite{CycleGAN2017} do not generalize to 3D viewpoint change; Infinite Nature-Zero~\cite{infinitenaturezero} couples cycle consistency with monocular depth warping, but requires that warping at inference and covers a single domain. Single-image supervision via 3D bounding box annotations~\cite{laconic} trades one scarce signal for another. All methods in this family share the same structural bottleneck: a depth estimator is load-bearing at inference. \mname removes it through monocular pretraining on 30 million unconstrained single images, with no category prior, and reaches state-of-the-art geometry-free performance after minimal multi-view fine-tuning.

\myparagraph{Large generative priors for novel view synthesis.}
An alternative strategy adapts large pretrained generative models to novel view synthesis via fine-tuning on posed multi-view data: Zero-1-to-3~\cite{zeroonetothree} adapts Stable Diffusion~\cite{stablediffusion} on Objaverse~\cite{objaverse} for object-level view synthesis; ViewCrafter~\cite{viewcrafter} conditions DynamiCrafter~\cite{dynamicrafter} on DUSt3R~\cite{dust3r} renders for scene-level pose control; Stable Virtual Camera~\cite{seva} fine-tunes a video diffusion model with Pl\"{u}cker ray conditioning; and PE-Field~\cite{bai2025positional} adapts Flux.1~Kontext~\cite{fluxkontext} with depth-aware 3D encodings. These methods pretrain on a proxy task, image or video generation, and inherit the compute-heavy architectures of their base models, making inference expensive by design. \mname instead pretrains directly on novel view synthesis, scaled to 30 million in-the-wild images, with a compact feed-forward architecture: it runs at 116~FPS, more than 7,000$\times$ faster than ViewCrafter, while surpassing it on RealEstate10K.

\myparagraph{Single-image 3D reconstruction and scene generation.}
A parallel family maps a single image to an explicit 3D representation and renders target views from it: feed-forward Gaussian predictors (SHARP~\cite{sharp}, DiffusionGS~\cite{diffusiongs}), per-scene optimization pipelines (WonderWorld~\cite{wonderworld}, G4Splat~\cite{g4splat}), and video-diffusion scene generators (FlexWorld~\cite{flexworld}, One2Scene~\cite{one2scene}, WorldStereo~\cite{worldstereo}, HY-World~2.0~\cite{hyworld2}), several of which target a 360-degree explorable scene in their own coordinate frame rather than a target view at a requested pose. The operating regimes are complementary: these methods pay a one-time per-scene construction, from roughly a second for SHARP to several minutes for FlexWorld, after which rendering is cheap, amortized, and resolution-agnostic, whereas \mname stores no geometry and pays one forward pass per view, the first view included. Quantitative comparisons appear in the supplementary material (Section~\ref{subsec:supp_explicit3d}).

\myparagraph{Monocular depth estimation.}
Learning-based depth estimation splits into two branches.
The \emph{relative} branch (MiDaS~\cite{midas,midas3},
DPT~\cite{DPT}, Depth Anything~\cite{depthanything},
Marigold~\cite{marigold1,marigold2}, MoGe~\cite{moge1}) yields affine-invariant predictions.
The \emph{metric} branch (ZoeDepth~\cite{zoedepth},
UniDepth~\cite{unidepth1,unidepth2}, MoGe-2~\cite{moge2}) also recovers
absolute scale.
\mname exploits this: metric depth lets us build pseudo-pairs with metric translations at training time, providing geometric supervision
without manual annotation.

\myparagraph{Warping-based novel view synthesis.}
Warping-based methods unproject the source image into 3D and inpaint disoccluded regions via geometry-based pop-up models~\cite{popup,worldsheet}, layered depth inpainting~\cite{niklaus_3d,shih3d}, soft point-cloud rendering~\cite{synsin,slide}, MPI blending~\cite{single_view_mpi,re10k}, or diffusion-based pipelines such as GenWarp~\cite{genwarp} and LucidDreamer~\cite{luciddreamer}. Infinite Nature~\cite{infinitenature} extends this to perpetual view generation but relies on posed video sequences for training. MultiDiff~\cite{multidiff} conditions a video diffusion model on depth-warped inputs, tying quality to depth accuracy. Because the depth estimator is load-bearing at inference and pipelines are iterative, these methods are fragile to depth failures and computationally slow; \mname requires no depth estimate at inference.

Geometry-free methods~\cite{geogpt,photonvs,vivid} require multi-view data from the start; methods without multi-view supervision, such as Infinite Nature-Zero~\cite{infinitenaturezero}, still need depth at inference. No prior work explores monocular pretraining as an initialization for geometry-free novel view synthesis. \mname fills this gap through a masked training formulation that confines supervision to valid reprojected pixels and covers disoccluded regions adversarially, reducing depth to a training scaffold discarded at inference.

\input{figures/method_overview}
\section{Method}
\label{sec:method}

\begin{wrapfigure}{r}{0.44\columnwidth}
\vspace{-6pt}
\centering
\setlength{\tabcolsep}{1pt}
\begin{tabular}{ccc}
{\scriptsize Source} & {\scriptsize Pseudo-Target} & {\scriptsize Prediction} \\[1pt]
\includegraphics[width=0.135\columnwidth]{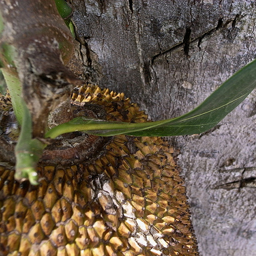} &
\includegraphics[width=0.135\columnwidth]{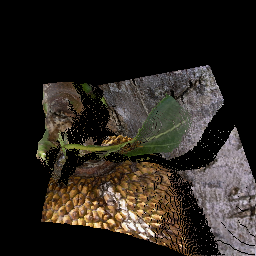} &
\includegraphics[width=0.135\columnwidth]{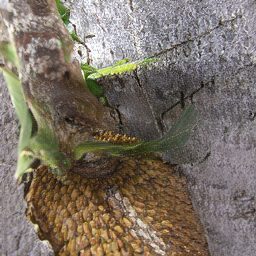} \\[1pt]
\includegraphics[width=0.135\columnwidth]{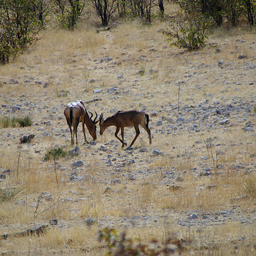} &
\includegraphics[width=0.135\columnwidth]{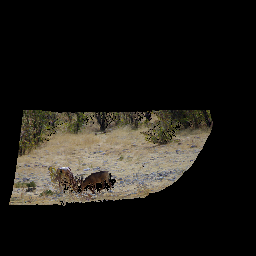} &
\includegraphics[width=0.135\columnwidth]{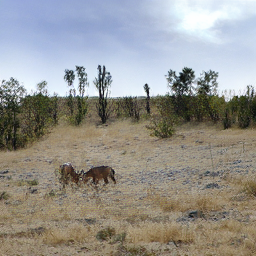} \\[1pt]
\includegraphics[width=0.135\columnwidth]{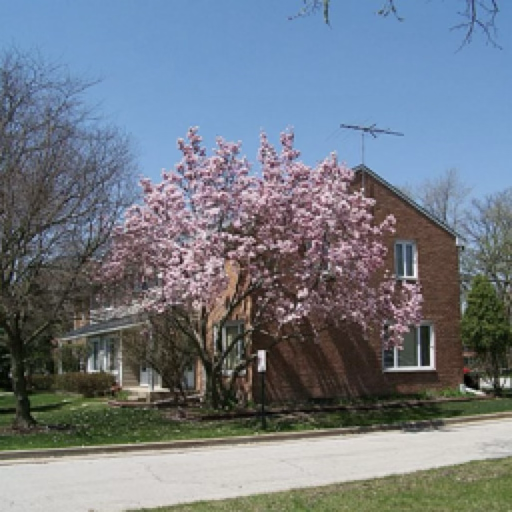} &
\includegraphics[width=0.135\columnwidth]{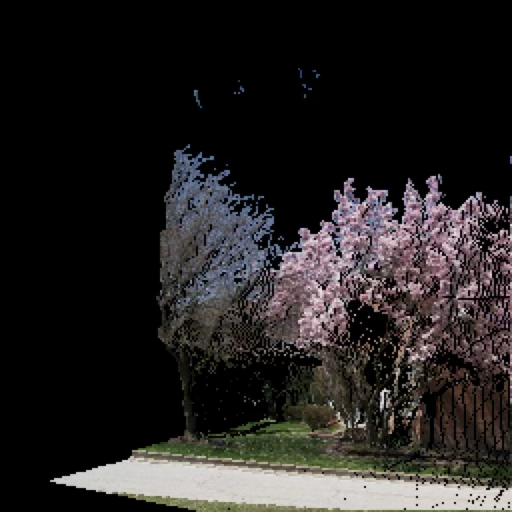} &
\includegraphics[width=0.135\columnwidth]{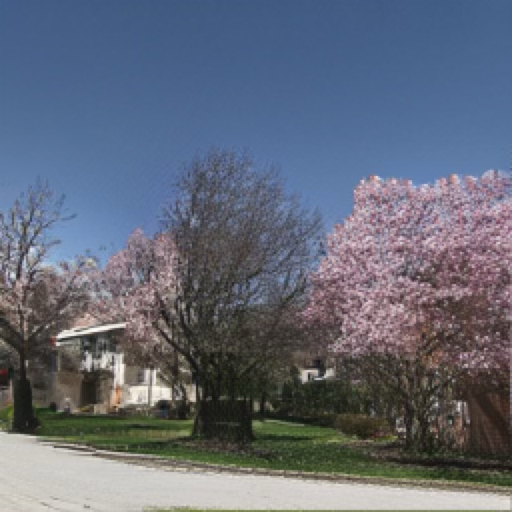}
\end{tabular}
\caption{\textbf{Pseudo-pairs and predictions.}}
\label{fig:pseudo_pair}
\vspace{-6pt}
\end{wrapfigure}

Given a source image $I_0$ and a camera transformation $T_{0 \rightarrow 1} \in SE(3)$, OVIE generates a target view $\hat{I}_1$. Monocular pretraining on large single-image collections builds on two components: on-the-fly pseudo-pair generation and partial supervision. A frozen monocular depth network lifts each source image into 3D; reprojecting the point cloud under sampled camera poses yields sparse pseudo-targets $I_1^*$ with binary validity masks $M$, which a pose-conditional image-to-image network $f_\theta$ uses as ground truth (Fig.~\ref{fig:method}, Fig.~\ref{fig:pseudo_pair}). At inference, OVIE discards the depth estimator and synthesizes the novel view in a single forward pass. The pretrained model transfers to a specific domain via brief fine-tuning on posed multi-view data.

\subsection{Annotation-Free Training Pair Construction}

\myparagraph{Depth-Based Scene Lifting.}
A pretrained monocular depth network processes the source image $I_0$, estimating a depth map $D \in \mathbb{R}^{H \times W}$ and surface normals $N \in \mathbb{R}^{H \times W \times 3}$. Together, these quantities define a point cloud $\mathcal{P} \in \mathbb{R}^{HW \times 3}$ in the source camera coordinate system via standard unprojection.

\myparagraph{Viewpoint Sampling and Reprojection.}
We sample $T_{0 \rightarrow 1} \in SE(3)$ from a mixture of six strategies assigned stochastically: identity, pure translation, pure rotation, combined rotation and translation, normal-derived (camera placed along an estimated surface normal), and frontal hemisphere (camera orbiting an anchor point). Translation magnitudes scale with median scene depth; rotations are bounded by the estimated field of view to maintain sufficient overlap. Reprojecting the transformed $\mathcal{P}$ yields a pseudo-ground-truth view $I_1^*$ and a binary mask $M \in \{0,1\}^{H \times W}$, where $M_{ij} = 0$ marks disocclusions, backface-culled regions, and out-of-frame content. Full details and sampling weights appear in the supplementary material (Section~\ref{sec:appendix_sampling}), and Section~\ref{subsec:pseudotargets} places pseudo-targets beside the views \mname generates from the same pose.

\myparagraph{Metric-Scale Supervision.}
When the depth model is metric (e.g., MoGE-2~\cite{moge2}), the resulting pairs encode metric-scale changes, a supervision signal far scarcer than standard pose annotations.

\subsection{Training Objective}
\label{sec:training_objective}

We use a multi-term objective that enforces geometric accuracy, semantic consistency, and textural realism. Because the pseudo-target $I_1^*$ contains missing content, the reconstruction and perceptual losses are restricted to observed regions via the mask $M$; the adversarial loss addresses disoccluded regions through a different mechanism, described below.

\myparagraph{Geometric Consistency.}
The main supervision is a masked reconstruction loss. Depth estimation errors propagate into the pseudo-targets, but three complementary mechanisms limit their impact: the mask $M$ eliminates gross failures at depth discontinuities and occlusion boundaries; the L2 penalty is everywhere differentiable, so the gradient stays stable even near masked boundaries; and training on 30 million images keeps any single erroneous target from dominating. The reconstruction loss is:
\begin{eqnarray}
\label{eq:recon}
\mathcal{L}_{\text{recon}} = \frac{\big\| M \odot \hat{I}_1 - M \odot I_1^* \big\|_2^2}{\left\| M \right\|_1 + \epsilon},
\end{eqnarray}
where $\epsilon > 0$ prevents division by zero.

\myparagraph{Semantic Preservation.}
We mask both the prediction and target prior to feature extraction, restricting the loss to valid regions.
Following prior work~\cite{pixelgen}, we combine LPIPS~\cite{lpips} with P-DINO, a patch-level loss derived from activations of a pretrained DINO model~\cite{dinov1, dinov2, dinov3}:
\begin{equation}
\mathcal{L}_{\text{perc}} = \lambda_{\text{LPIPS}} \mathcal{L}_{\text{LPIPS}}(M \odot \hat{I}_1, M \odot I_1^*) + \lambda_{\text{DINO}} \mathcal{L}_{\text{P-DINO}}(M \odot \hat{I}_1, M \odot I_1^*),
\end{equation}
where $\mathcal{L}_{\text{P-DINO}}(\mathbf{x}, \mathbf{y}) = 1 - \cos(\text{DINO}(\mathbf{x}), \text{DINO}(\mathbf{y}))$, and $\lambda_{\text{LPIPS}}, \lambda_{\text{DINO}} > 0$ are scalar weights balancing the two perceptual terms.

\myparagraph{Disocclusion Supervision and High-Frequency Realism.}
In disoccluded regions ($M_{ij}=0$), neither $\mathcal{L}_{\text{recon}}$ nor $\mathcal{L}_{\text{perc}}$ contributes a gradient, so the model must hallucinate plausible content. A PatchGAN adversarial objective~\cite{patchgan} covers this gap: because the incomplete $I_1^*$ cannot serve as a reliable real sample, $I_0$ acts as the real distribution instead, an approximation that holds since viewpoint sampling bounds parallax and keeps patch statistics consistent across views:
\begin{equation}
\mathcal{L}_{\text{adv}} = \mathbb{E}_{I_0}[\log D(I_0)] + \mathbb{E}_{\hat{I}_1}[\log(1 - D(\hat{I}_1))].
\end{equation}

The discriminator $D_\phi$ follows the StyleGAN-T~\cite{stylegant} design, using a pretrained representation backbone. The adaptive weight $w_{\text{adap}}$, introduced in VQ-GAN~\cite{vqgan}, balances reconstruction and adversarial terms automatically. Details appear in the supplementary material (Section~\ref{subsec:supp_arch}). 

\myparagraph{Total Objective.}
The total loss combines all three terms:
\begin{equation}
\mathcal{L}_{\text{total}} = \mathcal{L}_{\text{recon}} + \mathcal{L}_{\text{perc}} + \lambda_{\text{a}} \cdot w_{\text{adap}} \cdot \mathcal{L}_{\text{adv}},
\end{equation}
where $\lambda_{\text{a}} > 0$ controls the contribution of the adversarial term. All hyperparameters are specified in the supplementary material (Section~\ref{subsec:supp_hyperparameters}).

\myparagraph{Fine-Tuning.}
When fine-tuning on a posed multi-view dataset, the real target view $I_1$ replaces the pseudo-target: $\mathcal{L}_{\text{recon}}$ and $\mathcal{L}_{\text{perc}}$ are computed against $I_1$ without masking, and $I_1$ also serves as the real sample for the discriminator, replacing $I_0$. The objective and architecture remain unchanged.

\subsection{Model Architecture}
A convolutional encoder downsamples the source image $I_0$ into a dense feature map; a stack of pose-conditioned Transformer blocks then processes this representation; a mirrored convolutional decoder finally upsamples it back to the source resolution, producing $\hat{I}_1$.

\myparagraph{Pose Conditioning.}
We encode the relative transformation $T_{0 \rightarrow 1}$ as a 7D vector $\mathbf{p} \in \mathbb{R}^7$ (3D translation and unit quaternion), following VGGT~\cite{vggt} but omitting camera intrinsics: MoGe-2~\cite{moge2} supplies intrinsic estimates at training time, and camera geometry is read implicitly from the source image's perspective cues, with 30 million diverse training images providing robustness to intrinsic variation at inference. Dropping intrinsics also keeps the inference interface minimal. A linear layer projects $\mathbf{p}$ into a conditioning token $\mathbf{c} = W\mathbf{p} \in \mathbb{R}^{d}$, which modulates each Transformer block via Adaptive Layer Normalization (AdaLN)~\cite{dit}, detailed in Section~\ref{subsec:supp_adaln}.

\section{Experiments}
\label{sec:experiments}

\ifdefined\sotaGap\else\newlength{\sotaGap}\fi\setlength{\sotaGap}{1pt}
\ifdefined\sotaSep\else\newlength{\sotaSep}\fi\setlength{\sotaSep}{3pt}
\ifdefined\sotaImgW\else\newlength{\sotaImgW}\fi\setlength{\sotaImgW}{\dimexpr(\linewidth - 5\sotaGap - 2\sotaSep - 1pt)/6\relax}

\begin{figure*}[t]
\centering
\setlength{\tabcolsep}{0pt}%
\begin{tabular}{
  @{}
  c @{\hspace{\sotaGap}}
  c
  @{\hspace{\sotaSep}}!{\color{black!25}\vrule width 0.4pt}@{\hspace{\sotaSep}}
  c @{\hspace{\sotaGap}}
  c @{\hspace{\sotaGap}}
  c @{\hspace{\sotaGap}}
  c
  @{}
}

\small Source &
\small Target &
\small \mname (Ours) &
\small GeoGPT &
\small PhotoNVS &
\small VIVID\\[\sotaGap]

\includegraphics[width=\sotaImgW]{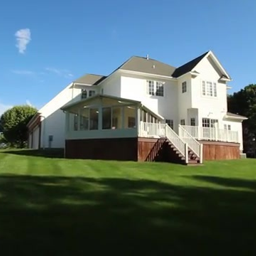} &
\includegraphics[width=\sotaImgW]{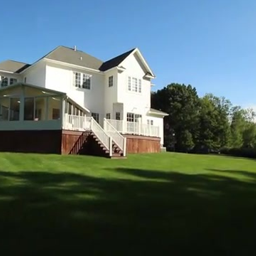} &
\includegraphics[width=\sotaImgW]{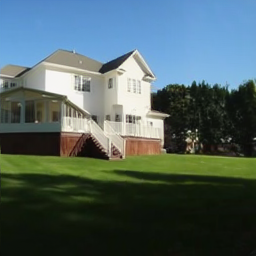} &
\includegraphics[width=\sotaImgW]{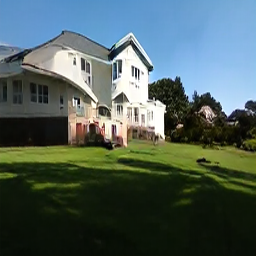} &
\includegraphics[width=\sotaImgW]{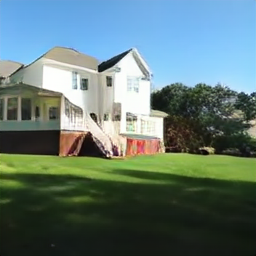} &
\includegraphics[width=\sotaImgW]{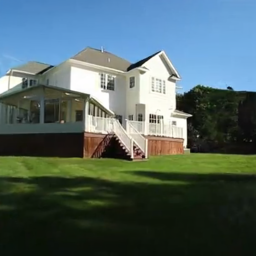} \\[\sotaGap]

\includegraphics[width=\sotaImgW]{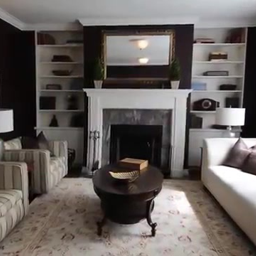} &
\includegraphics[width=\sotaImgW]{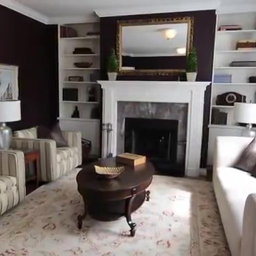} &
\includegraphics[width=\sotaImgW]{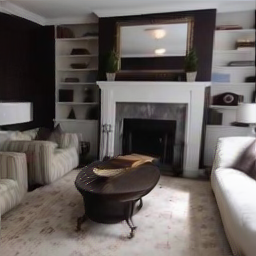} &
\includegraphics[width=\sotaImgW]{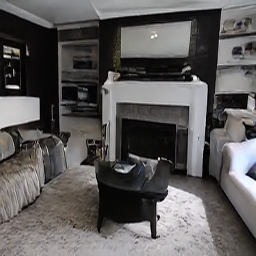} &
\includegraphics[width=\sotaImgW]{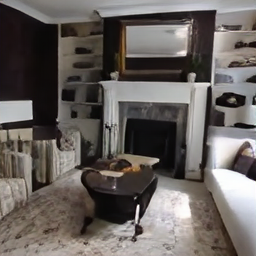} &
\includegraphics[width=\sotaImgW]{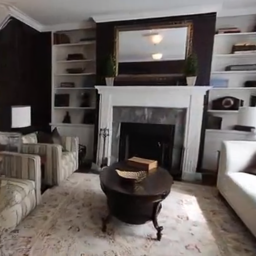} \\[\sotaGap]

\includegraphics[width=\sotaImgW]{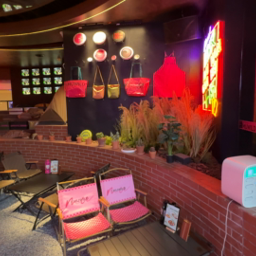} &
\includegraphics[width=\sotaImgW]{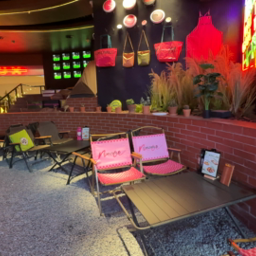} &
\includegraphics[width=\sotaImgW]{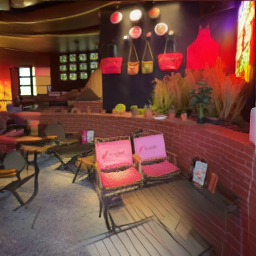} &
\includegraphics[width=\sotaImgW]{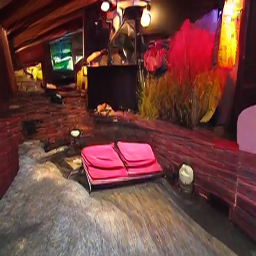} &
\includegraphics[width=\sotaImgW]{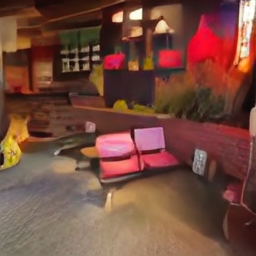} &
\includegraphics[width=\sotaImgW]{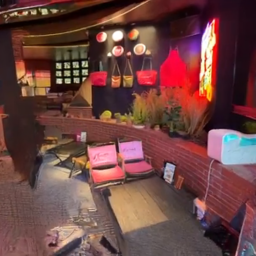} \\[\sotaGap]

\includegraphics[width=\sotaImgW]{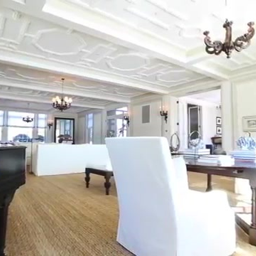} &
\includegraphics[width=\sotaImgW]{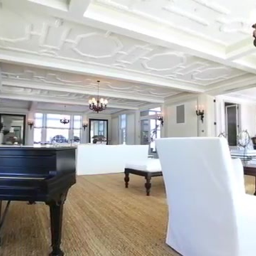} &
\includegraphics[width=\sotaImgW]{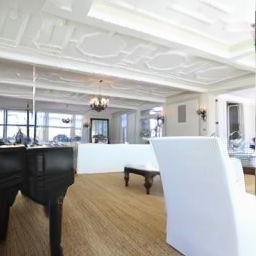} &
\includegraphics[width=\sotaImgW]{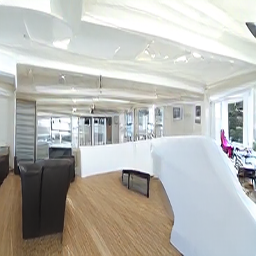} &
\includegraphics[width=\sotaImgW]{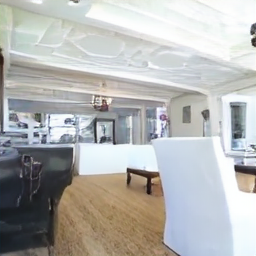} &
\includegraphics[width=\sotaImgW]{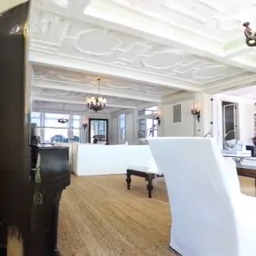}

\end{tabular}%
\caption{\textbf{Qualitative comparison with baseline geometry-free methods.}
Pretrained on single images only, \mname produces sharp, geometrically consistent novel views.}
\label{fig:qualitative}
\end{figure*}

We evaluate two model variants: \mname, trained exclusively on unpaired single images via monocular pretraining, and \mname-ft, fine-tuned on RealEstate10K from the pretrained checkpoint. Evaluation covers five axes: qualitative generalization (Figure~\ref{fig:qualitative}), quantitative comparison on RealEstate10K, where geometry-free baselines are in-domain and \mname is out-of-domain, and on DL3DV, where all geometry-free methods face the same domain shift (Table~\ref{tab:comparison_with_sota}), multi-view 3D consistency of the generated trajectories, throughput showing \mname runs at real-time speeds more than 600$\times$ faster than the fastest baseline (Figure~\ref{fig:speed}), and ablations of loss design and training data (Table~\ref{tab:loss_ablations} and Figure~\ref{fig:dataset_scale}).

\subsection{Experimental Setup}
\myparagraph{Training Data.}
\mname is trained on 30 million in-the-wild images from ImageNet-21K~\cite{in21k}, Open Images~\cite{openimages}, OpenStreetView5M (OSV5M)~\cite{osv5m}, and Places~\cite{places}, with no multi-view supervision; MoGe-2~\cite{moge2} generates pseudo-pairs on-the-fly. \mname-ft is fine-tuned on RealEstate10K~\cite{re10k}.

\myparagraph{Benchmarks and Metrics.}
We evaluate on RealEstate10K~\cite{re10k} and DL3DV~\cite{dl3dv}; \mname has never been trained on either. PSNR and SSIM measure pixel-level fidelity, LPIPS~\cite{lpips} perceptual similarity, FID distributional realism, and MEt3R~\cite{met3r} multi-view 3D consistency between pairs of generated frames. Following prior work~\cite{bai2025positional, gen3c}, we sample a starting frame from each of 750 scenes, generate 14 novel views at a stride of 3 (RealEstate10K) or 1 (DL3DV), and average metrics across scenes, with FID computed over all generated and ground-truth images. A fixed stride corresponds to different physical motions across videos, so Section~\ref{subsec:supp_envelope} reports the rotation and translation distributions each benchmark induces. Since benchmark poses are derived from SLAM and are scale-ambiguous, we perform a per-scene scale sweep for each geometry-free method, selecting the value that maximizes its own performance. ViewCrafter exposes no explicit SE(3) pose interface and recovers the target cameras by running DUSt3R over the evaluation frames, so it is evaluated at its native DUSt3R scale (Section~\ref{subsec:supp_protocol}). We re-evaluate every method under this protocol rather than quoting the numbers published in each paper, which differ in stride, pair selection, and crop conventions and are therefore not comparable to one another. Section~\ref{subsec:supp_vividprotocol} additionally evaluates \mname-ft under VIVID's own protocol as a cross-check.

\myparagraph{Baselines.}
We compare \mname against GeoGPT~\cite{geogpt}, PhotoNVS~\cite{photonvs}, and VIVID~\cite{vivid}, geometry-free image-to-image methods that share our problem formulation and train exclusively on RealEstate10K. The asymmetry is deliberate: these baselines are \emph{in-domain} on RealEstate10K while \mname is \emph{out-of-domain}, and \mname-ft is in-domain; on DL3DV every geometry-free method is out-of-domain, which makes that comparison fair. ViewCrafter~\cite{viewcrafter} is included as a geometry-based reference: it offloads geometric reasoning to DUSt3R~\cite{dust3r} and builds on a large video diffusion foundation model (${\sim}10{\times}$ more parameters), trained on both benchmarks. An additional comparison with Infinite Nature-Zero~\cite{infinitenaturezero}, a geometry-based method trained with single images, appears in the supplementary material (Section~\ref{sec:appendix_infinitenaturezero}). Single-image methods that build an explicit 3D representation and render from it (SHARP~\cite{sharp}, DiffusionGS~\cite{diffusiongs}, FlexWorld~\cite{flexworld}) are evaluated under the same protocol in Section~\ref{subsec:supp_explicit3d}.

\myparagraph{Implementation Details.}
OVIE uses a convolutional encoder, a ViT-B bottleneck~\cite{vit}, and a convolutional decoder, trained for 2M steps at batch size 512; ablations use 250K steps. The architecture is resolution-agnostic: a $512{\times}512$ variant trained under the identical recipe, with only the resolution changed, is reported in Section~\ref{subsec:supp_resolution}. Full architectural details and hyperparameters are in the supplementary material (Section~\ref{sec:appendix_implementation}), and compute requirements in Section~\ref{sec:appendix_resources}.

\subsection{Qualitative Results}

\newcommand{\tikzbanana}[1][1]{%
  \begin{tikzpicture}[scale=#1, rotate=-30]
    \fill[yellow!80!orange, draw=yellow!50!orange, line width=0.3pt]
      (0, 0)
      .. controls (0.10, 0.18) and (0.35, 0.30) .. (0.65, 0.30)
      .. controls (0.85, 0.30) and (1.05, 0.20) .. (1.15, 0.08)
      .. controls (1.18, 0.04) and (1.18, 0.00) .. (1.15,-0.02)
      .. controls (1.00, 0.08) and (0.70, 0.12) .. (0.45, 0.08)
      .. controls (0.20, 0.04) and (0.05,-0.06) .. cycle;
    \fill[yellow!95!white, opacity=0.45]
      (0.20, 0.12)
      .. controls (0.40, 0.23) and (0.65, 0.25) .. (0.85, 0.20)
      .. controls (0.70, 0.20) and (0.45, 0.18) .. (0.20, 0.12);
    \draw[yellow!45!orange, line width=0.25pt]
      (0.05, 0.04)
      .. controls (0.25, 0.18) and (0.55, 0.24) .. (0.80, 0.20)
      .. controls (0.98, 0.16) and (1.10, 0.08) .. (1.14, 0.03);
    \fill[yellow!25!brown]
      (0, 0)
      .. controls (-0.04, 0.03) and (-0.07, 0.04) .. (-0.09, 0.02)
      .. controls (-0.07,-0.01) and (-0.03,-0.02) .. cycle;
    \fill[yellow!20!brown]
      (1.15,-0.02)
      .. controls (1.18,-0.01) and (1.20,-0.03) .. (1.18,-0.05)
      .. controls (1.16,-0.04) and (1.14,-0.03) .. cycle;
  \end{tikzpicture}%
}

\newcommand{\drawOpaqueCamera}[9]{%
  \pgfmathsetmacro{\cvPx}{#1}%
  \pgfmathsetmacro{\cvPy}{#2}%
  \pgfmathsetmacro{\cvPz}{#3}%
  \cvComputeRotation{#4}{#5}{#6}{#7}%
  \pgfmathsetmacro{\cvFw}{0.12*(#9)}%
  \pgfmathsetmacro{\cvFh}{0.09*(#9)}%
  \pgfmathsetmacro{\cvFd}{0.25*(#9)}%
  \pgfmathsetmacro{\cvUh}{0.09*(#9)}%
  \pgfmathsetmacro{\cvUw}{0.08*(#9)}%
  \pgfmathsetmacro{\cvLens}{0.025*(#9)}%
  \cvTransformPoint{0}{0}{0}{cvFO}%
  \cvTransformPoint{-\cvFw}{-\cvFd}{-\cvFh}{cvFBL}%
  \cvTransformPoint{\cvFw}{-\cvFd}{-\cvFh}{cvFBR}%
  \cvTransformPoint{\cvFw}{-\cvFd}{\cvFh}{cvFTR}%
  \cvTransformPoint{-\cvFw}{-\cvFd}{\cvFh}{cvFTL}%
  \cvTransformPoint{-\cvUw}{-\cvFd}{\cvFh}{cvFUA}%
  \cvTransformPoint{\cvUw}{-\cvFd}{\cvFh}{cvFUB}%
  \cvTransformPoint{0}{-\cvFd}{\cvFh+\cvUh}{cvFUC}%
  \cvComputeVisibility
  \fill[white] (cvFO)--(cvFBL)--(cvFBR)--cycle;%
  \fill[#8, opacity=0.08] (cvFO)--(cvFBL)--(cvFBR)--cycle;%
  \fill[white] (cvFO)--(cvFBR)--(cvFTR)--cycle;%
  \fill[#8, opacity=0.08] (cvFO)--(cvFBR)--(cvFTR)--cycle;%
  \fill[white] (cvFO)--(cvFTR)--(cvFTL)--cycle;%
  \fill[#8, opacity=0.08] (cvFO)--(cvFTR)--(cvFTL)--cycle;%
  \fill[white] (cvFO)--(cvFTL)--(cvFBL)--cycle;%
  \fill[#8, opacity=0.08] (cvFO)--(cvFTL)--(cvFBL)--cycle;%
  \fill[white] (cvFBL)--(cvFBR)--(cvFTR)--(cvFTL)--cycle;%
  \fill[#8, opacity=0.15] (cvFBL)--(cvFBR)--(cvFTR)--(cvFTL)--cycle;%
  \fill[white] (cvFUA)--(cvFUB)--(cvFUC)--cycle;%
  \fill[color=#8] (cvFUA)--(cvFUB)--(cvFUC)--cycle;%
  \cvEdge{#8}{cvFO}{cvFBL}{\cvVisBOT}{\cvVisLEF}%
  \cvEdge{#8}{cvFO}{cvFBR}{\cvVisBOT}{\cvVisRIG}%
  \cvEdge{#8}{cvFO}{cvFTR}{\cvVisRIG}{\cvVisTOP}%
  \cvEdge{#8}{cvFO}{cvFTL}{\cvVisTOP}{\cvVisLEF}%
  \cvEdge{#8}{cvFBL}{cvFBR}{\cvVisBOT}{\cvVisFRO}%
  \cvEdge{#8}{cvFBR}{cvFTR}{\cvVisRIG}{\cvVisFRO}%
  \cvEdge{#8}{cvFTR}{cvFTL}{\cvVisTOP}{\cvVisFRO}%
  \cvEdge{#8}{cvFTL}{cvFBL}{\cvVisLEF}{\cvVisFRO}%
  \fill[#8] (cvFO) circle[radius=\cvLens];%
}

\newcommand{\metricImg}[2]{%
  \tikz[baseline=(img.center)]{%
    \node[inner sep=0pt, draw=#1, line width=1.5pt] (img)
      {\includegraphics[width=1.8cm]{#2}};%
  }%
}

\begin{figure}[t]
\centering
\tdplotsetmaincoords{70}{255}%
\resizebox{\linewidth}{!}{%
\setlength{\tabcolsep}{0pt}%
\begin{tabular}{
  @{}
  c
  @{\hspace{5mm}}
  !{\color{black!25}\vrule width 0.4pt}
  @{\hspace{5mm}}
  c
  @{}
}

\small\textbf{Scene 1} {\scriptsize(object scale)} &
\small\textbf{Scene 2} {\scriptsize(room scale)} \\[2mm]

\begin{tikzpicture}[tdplot_main_coords, scale=2]
  \node[inner sep=0pt] at (0, 1.5, 0) {\tikzbanana[0.8]};
  \draw[dashed, semithick, black!50]
    (0, 0.05, 0) --
    node[left=3pt, font=\scriptsize, fill=white, inner sep=1pt, pos=0.4] {50\,cm}
    (0, 1.38, 0);
  \drawOpaqueCamera{0.5}{0}{0}%
    {-0.7071}{0}{0}{0.7071}{transorange}{1.8}
  \node[font=\sffamily\fontsize{4}{4}\selectfont, transorange,
        fill=white, inner sep=0.8pt, rounded corners=0.5pt]
    at (0.5, -0.15, 0) {Gen.};
  \drawOpaqueCamera{-0.5}{0}{0}%
    {-0.7071}{0}{0}{0.7071}{srcblue}{1.8}
  \node[font=\sffamily\fontsize{4}{4}\selectfont, srcblue,
        fill=white, inner sep=0.8pt, rounded corners=0.5pt]
    at (-0.5, -0.15, 0) {Src.};
  \draw[{Stealth[length=3pt]}-{Stealth[length=3pt]},
        thick, black!70]
    (-0.5, -0.30, 0) -- (0.5, -0.30, 0);
  \node[below=2pt, font=\scriptsize, fill=white, inner sep=1pt]
    at (0.25, -0.30, 0) {20\,cm};
  \drawInfiniteGroundGrid{0.5}
\end{tikzpicture}
&
\begin{tikzpicture}[tdplot_main_coords, scale=2]
  \node[inner sep=0pt] at (0, 3, 0) {\tikzbanana[0.8]};
  \draw[dashed, semithick, black!50]
    (0, 0.05, 0) --
    node[right=3pt, font=\scriptsize, fill=white, inner sep=1pt, pos=0.4] {3\,m}
    (0, 2.88, 0);
  \drawOpaqueCamera{0.5}{0}{0}%
    {-0.7071}{0}{0}{0.7071}{transorange}{1.8}
  \node[font=\sffamily\fontsize{4}{4}\selectfont, transorange,
        fill=white, inner sep=0.8pt, rounded corners=0.5pt]
    at (0.5, -0.15, 0) {Gen.};
  \drawOpaqueCamera{-0.5}{0}{0}%
    {-0.7071}{0}{0}{0.7071}{srcblue}{1.8}
  \node[font=\sffamily\fontsize{4}{4}\selectfont, srcblue,
        fill=white, inner sep=0.8pt, rounded corners=0.5pt]
    at (-0.5, -0.15, 0) {Src.};
  \draw[{Stealth[length=3pt]}-{Stealth[length=3pt]},
        thick, black!70]
    (-0.5, -0.30, 0) -- (0.5, -0.30, 0);
  \node[below=2pt, font=\scriptsize, fill=white, inner sep=1pt]
    at (0.25, -0.30, 0) {20\,cm};
  \drawInfiniteGroundGrid{0.5}
\end{tikzpicture}
\\[2mm]

\begin{tabular}{@{}c@{\hspace{1.5mm}}c@{}}
  \metricImg{srcblue}{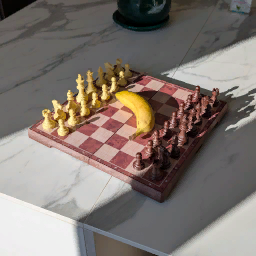} &
  \metricImg{transorange}{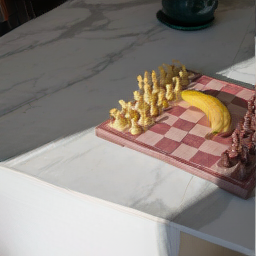} \\[1pt]
  {\scriptsize Source} & {\scriptsize Generated}
\end{tabular}
&
\begin{tabular}{@{}c@{\hspace{1.5mm}}c@{}}
  \metricImg{srcblue}{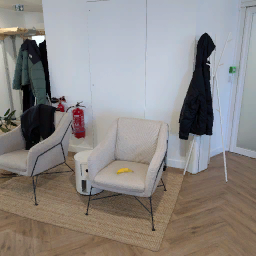} &
  \metricImg{transorange}{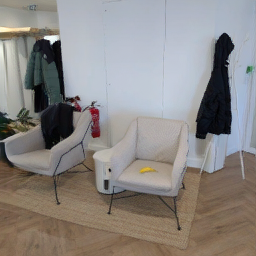} \\[1pt]
  {\scriptsize Source} & {\scriptsize Generated}
\end{tabular}

\end{tabular}%
}%
\caption{\textbf{Metric scale awareness.}
The same translation applied to a close-up scene (left) produces a larger apparent displacement than on a room-scale scene (right), consistent with metric parallax.}
\label{fig:metric}
\end{figure}

Figure~\ref{fig:teaser} shows \mname generalizing across indoor, outdoor, object-centric, and non-photographic scenes, and Figure~\ref{fig:qualitative} compares it against the baselines: despite training on none of their evaluation data, \mname matches or exceeds their quality. Figure~\ref{fig:metric} illustrates the metric scale awareness inherited from MoGe-2, where the same camera translation displaces nearby objects more than distant ones. Sections~\ref{subsec:ood} and~\ref{subsec:baselines} add further out-of-distribution inputs, including paintings, and further baseline comparisons.

\subsection{Comparison with State of the Art}

\myparagraph{RealEstate10K: competing at a disadvantage.}
\mname outperforms three of the four in-domain baselines (Table~\ref{tab:comparison_with_sota}), despite never training on RealEstate10K data; the remaining gap with VIVID reflects the domain disadvantage, as DL3DV confirms.
Fine-tuning \mname for 2\% of the pretraining budget (\mname-ft) reaches state of the art among geometry-free methods, leading all baselines on pixel fidelity and perceptual similarity, over 600$\times$ faster than any competitor. The lead comes from pretraining: the same architecture trained from scratch on RealEstate10K alone reaches 19.05 PSNR, 2.85~dB below \mname-ft.

\myparagraph{DL3DV: outperforming all baselines under domain shift.}
When all methods face the same domain shift on DL3DV, \mname outperforms all baselines on all metrics: the diversity of its large-scale training data delivers robustness that in-domain specialization cannot match. Notably, \mname outperforms \mname-ft on DL3DV on pixel fidelity, perceptual similarity, and FID, confirming that fine-tuning on a single domain trades generalization for in-domain performance; multi-view consistency is the one metric that still favours \mname-ft. The effect is symmetric: fine-tuning the same checkpoint on DL3DV raises its DL3DV PSNR to 17.07 and slightly improves its RealEstate10K scores (Section~\ref{subsec:dl3dv_finetuning}). Fine-tuning therefore selects a domain.

\begin{table}[t]
\centering
\caption{\textbf{In-distribution and out-of-domain robustness on RealEstate10K and DL3DV.}
$\uparrow$/$\downarrow$: higher/lower is better. \textbf{Bold}: best; \underline{underline}: second best.
OOD: not trained on the benchmark. MEt3R~\cite{met3r} scores multi-view 3D consistency between consecutive generated views.
$^*$Trained in-domain on DL3DV: scores greyed out and excluded from rankings for fairness.}
\label{tab:comparison_with_sota}
\setlength{\tabcolsep}{3pt}
\fitwidth{%
\begin{tabular}{@{}l cccccc cccccc@{}}
\toprule
& \multicolumn{6}{c}{\cellcolor{lavblue}RealEstate10K~\cite{re10k}} & \multicolumn{6}{c}{\cellcolor{lavblue}DL3DV~\cite{dl3dv}} \\
\cmidrule(lr){2-7} \cmidrule(lr){8-13}
Method & OOD & PSNR\,$\uparrow$ & SSIM\,$\uparrow$ & LPIPS\,$\downarrow$ & FID\,$\downarrow$ & MEt3R\,$\downarrow$
       & OOD & PSNR\,$\uparrow$ & SSIM\,$\uparrow$ & LPIPS\,$\downarrow$ & FID\,$\downarrow$ & MEt3R\,$\downarrow$ \\
\midrule
\rowcolor{white}
GeoGPT~\cite{geogpt}   & \textcolor{red}{\ding{55}} & 15.3 & 0.480 & 0.446 & 18.0 & 0.184 & \textcolor{green!60!black}{\ding{51}} & 13.1 & 0.339 & 0.560 & 35.9 & 0.300 \\
\rowcolor{ltorange}
ViewCrafter$^*$~\cite{viewcrafter} & \textcolor{red}{\ding{55}} & 16.6 & 0.559 & 0.340 & 9.98 & 0.193 & \textcolor{red}{\ding{55}} & \textcolor{gray}{15.4} & \textcolor{gray}{0.419} & \textcolor{gray}{0.425} & \textcolor{gray}{10.50} & \textcolor{gray}{0.261} \\
\rowcolor{white}
PhotoNVS~\cite{photonvs}& \textcolor{red}{\ding{55}} & 18.9  & 0.601 & 0.314 & 10.6 & 0.139 & \textcolor{green!60!black}{\ding{51}} & 13.8 & 0.349 & 0.525 & 37.6 & 0.301 \\
\rowcolor{ltorange}
VIVID~\cite{vivid}    & \textcolor{red}{\ding{55}} & \underline{20.5} & \underline{0.661} & \underline{0.241} & \textbf{4.26} & 0.076 & \textcolor{green!60!black}{\ding{51}} & \underline{14.5} & \underline{0.362} & \underline{0.471} & \underline{18.0} & 0.169 \\
\midrule
\rowcolor{white}
\mname (ours) & \textcolor{green!60!black}{\ding{51}} & 18.8  & 0.602 & 0.279 & 6.74 & \underline{0.035} & \textcolor{green!60!black}{\ding{51}} & \textbf{14.8} & \textbf{0.369} & \textbf{0.464} & \textbf{13.6} & \underline{0.078} \\
\rowcolor{ltorange}
\mname-ft (ours) & \textcolor{red}{\ding{55}} & \textbf{21.9} & \textbf{0.695} & \textbf{0.195} & \underline{5.59} & \textbf{0.029} & \textcolor{green!60!black}{\ding{51}} & 14.1 & 0.351 & 0.531 & 24.66 & \textbf{0.040} \\
\bottomrule
\end{tabular}%
}
\end{table}

\myparagraph{Multi-view consistency.}
The reconstruction and perceptual losses reach only co-visible pixels, disoccluded regions are covered by the adversarial prior alone, and no term couples different target views of the same source, so consistency along a trajectory is a property \mname acquires indirectly. MEt3R~\cite{met3r} measures it by reconstructing the shared geometry of each generated frame pair with MASt3R~\cite{mast3r} and scoring feature dissimilarity in the overlapping region, where a low value indicates that both frames depict one 3D scene. \mname leads every baseline on both benchmarks, ahead of ViewCrafter and of the geometry-dependent Infinite Nature-Zero (0.055 / 0.098), and \mname-ft improves further on both (Table~\ref{tab:comparison_with_sota}). Section~\ref{subsec:supp_met3r} extends the comparison to SHARP, to the in-domain DL3DV checkpoint, and to pairs of real video frames, and Section~\ref{subsec:supp_roundtrip} reports a round-trip test that holds near 22--23~dB through moderate motion. Section~\ref{subsec:supp_envelope} quantifies the operating envelope these results hold in, through camera displacement and co-visibility, and compares every method at matched motion.

\subsection{Real-Time Novel View Synthesis}
\begin{figure}[t]
\centering

\definecolor{grayline}{HTML}{B0B0B0}
\definecolor{graytext}{HTML}{505050}
\definecolor{geoblue}{HTML}{4C78A8}
\definecolor{vividorange}{HTML}{F28E2B}
\definecolor{oursgreen}{HTML}{59A14F}
\definecolor{oursgreenft}{HTML}{1B5E20}
\definecolor{photonvsgray}{HTML}{76B7B2}
\definecolor{vcred}{HTML}{E15759}

\begin{subfigure}[t]{0.48\textwidth}
  \centering
  \begin{tikzpicture}
    \begin{axis}[
      width=\linewidth,
      height=3.8cm,
      grid=major,
      grid style={grayline, line width=0.4pt, densely dotted, opacity=0.6},
      axis x line*=bottom,
      axis y line*=left,
      every outer x axis line/.append style={grayline, line width=0.6pt},
      every outer y axis line/.append style={grayline, line width=0.6pt},
      tick label style={font=\small, color=graytext},
      label style={font=\normalsize},
      clip=false,
      xmode=log,
      xlabel={Inference Speed (FPS)},
      ylabel={PSNR $\uparrow$},
      xmin=0.005, xmax=300,
      ymin=14, ymax=23,
      enlarge y limits=0.15,
      ytick={14,16,18,20,22},
      log ticks with fixed point,
    ]
    \addplot[only marks, mark=*, mark size=12.00pt, fill=vcred, draw=white, line width=0.7pt, fill opacity=0.85]
      coordinates {(0.0156, 16.6)} node[below, font=\footnotesize, text=vcred, yshift=-12pt,
      fill=white, fill opacity=0.7, text opacity=1, rounded corners=1pt, inner sep=2pt] {ViewCrafter};
    \addplot[only marks, mark=*, mark size=4.96pt, fill=geoblue, draw=white, line width=0.7pt, fill opacity=0.85]
      coordinates {(0.172, 15.3)} node[below, font=\footnotesize, text=geoblue, xshift=6pt, yshift=-6pt,
      fill=white, fill opacity=0.7, text opacity=1, rounded corners=1pt, inner sep=2pt] {GeoGPT};
    \addplot[only marks, mark=*, mark size=4.49pt, fill=vividorange, draw=white, line width=0.7pt, fill opacity=0.85]
      coordinates {(0.190, 20.5)} node[above, font=\footnotesize, text=vividorange, yshift=6pt,
      fill=white, fill opacity=0.7, text opacity=1, rounded corners=1pt, inner sep=2pt] {VIVID};
    \addplot[only marks, mark=*, mark size=2.52pt, fill=photonvsgray, draw=white, line width=0.7pt, fill opacity=0.85]
      coordinates {(0.024, 18.9)} node[above, font=\footnotesize, text=photonvsgray, yshift=4pt,
      fill=white, fill opacity=0.7, text opacity=1, rounded corners=1pt, inner sep=2pt] {PhotoNVS};
    \addplot[only marks, mark=*, mark size=2.62pt, fill=oursgreen, draw=white, line width=0.7pt, fill opacity=0.85]
      coordinates {(115.74, 18.8)} node[above left, font=\footnotesize\bfseries, text=oursgreen, xshift=-2pt, yshift=4pt,
      fill=white, fill opacity=0.7, text opacity=1, rounded corners=1pt, inner sep=2pt] {OVIE (ours)};
    \addplot[only marks, mark=*, mark size=2.62pt, fill=oursgreenft, draw=white, line width=0.7pt, fill opacity=0.85]
      coordinates {(115.74, 21.9)} node[above left, font=\footnotesize\bfseries, text=oursgreenft, xshift=-2pt, yshift=2pt,
      fill=white, fill opacity=0.7, text opacity=1, rounded corners=1pt, inner sep=2pt] {OVIE-ft (ours)};
    \end{axis}
  \end{tikzpicture}
  \caption{PSNR vs.\ FPS (upper-right is better)}
\end{subfigure}\hfill
\begin{subfigure}[t]{0.48\textwidth}
  \centering
  \begin{tikzpicture}
    \begin{axis}[
      width=\linewidth,
      height=3.8cm,
      grid=major,
      grid style={grayline, line width=0.4pt, densely dotted, opacity=0.6},
      axis x line*=bottom,
      axis y line*=left,
      every outer x axis line/.append style={grayline, line width=0.6pt},
      every outer y axis line/.append style={grayline, line width=0.6pt},
      tick label style={font=\small, color=graytext},
      label style={font=\normalsize},
      clip=false,
      xmode=log,
      xlabel={Inference Speed (FPS)},
      ylabel={LPIPS $\downarrow$},
      xmin=0.005, xmax=300,
      ymin=0.15, ymax=0.50,
      enlarge y limits=0.15,
      ytick={0.2,0.3,0.4,0.5},
      log ticks with fixed point,
    ]
    \addplot[only marks, mark=*, mark size=12.00pt, fill=vcred, draw=white, line width=0.7pt, fill opacity=0.85]
      coordinates {(0.0156, 0.340)} node[above, font=\footnotesize, text=vcred, yshift=12pt,
      fill=white, fill opacity=0.7, text opacity=1, rounded corners=1pt, inner sep=2pt] {ViewCrafter};
    \addplot[only marks, mark=*, mark size=4.96pt, fill=geoblue, draw=white, line width=0.7pt, fill opacity=0.85]
      coordinates {(0.172, 0.446)} node[above, font=\footnotesize, text=geoblue, yshift=6pt,
      fill=white, fill opacity=0.7, text opacity=1, rounded corners=1pt, inner sep=2pt] {GeoGPT};
    \addplot[only marks, mark=*, mark size=4.49pt, fill=vividorange, draw=white, line width=0.7pt, fill opacity=0.85]
      coordinates {(0.190, 0.241)} node[below, font=\footnotesize, text=vividorange, yshift=-6pt,
      fill=white, fill opacity=0.7, text opacity=1, rounded corners=1pt, inner sep=2pt] {VIVID};
    \addplot[only marks, mark=*, mark size=2.52pt, fill=photonvsgray, draw=white, line width=0.7pt, fill opacity=0.85]
      coordinates {(0.024, 0.314)} node[right, font=\footnotesize, text=photonvsgray, xshift=4pt,
      fill=white, fill opacity=0.7, text opacity=1, rounded corners=1pt, inner sep=2pt] {PhotoNVS};
    \addplot[only marks, mark=*, mark size=2.62pt, fill=oursgreen, draw=white, line width=0.7pt, fill opacity=0.85]
      coordinates {(115.74, 0.279)} node[below left, font=\footnotesize\bfseries, text=oursgreen, xshift=-2pt, yshift=-2pt,
      fill=white, fill opacity=0.7, text opacity=1, rounded corners=1pt, inner sep=2pt] {OVIE (ours)};
    \addplot[only marks, mark=*, mark size=2.62pt, fill=oursgreenft, draw=white, line width=0.7pt, fill opacity=0.85]
      coordinates {(115.74, 0.195)} node[below left, font=\footnotesize\bfseries, text=oursgreenft, xshift=-2pt, yshift=-4pt,
      fill=white, fill opacity=0.7, text opacity=1, rounded corners=1pt, inner sep=2pt] {OVIE-ft (ours)};

    \coordinate (bleg) at (rel axis cs: 0.95, 0.94);
    \end{axis}
    \begin{scope}[shift={($(bleg)+(4pt,-28pt)$)}]
      \fill[graytext, opacity=0.10] (0pt, 12.00pt) circle (12.00pt);  %
      \fill[graytext, opacity=0.22] (0pt,  4.90pt) circle ( 4.90pt);  %
      \fill[graytext, opacity=0.38] (0pt,  3.47pt) circle ( 3.47pt);  %
      \fill[graytext, opacity=0.58] (0pt,  2.19pt) circle ( 2.19pt);  %
      \draw[graytext!60, line width=0.3pt] (-2.19pt,  2.19pt) -- (-15pt,  2pt)
        node[left, font=\tiny, color=graytext, inner sep=1pt] {100M};
      \draw[graytext!60, line width=0.3pt] (-3.47pt,  3.47pt) -- (-15pt,  8pt)
        node[left, font=\tiny, color=graytext, inner sep=1pt] {250M};
      \draw[graytext!60, line width=0.3pt] (-4.90pt,  4.90pt) -- (-15pt, 14pt)
        node[left, font=\tiny, color=graytext, inner sep=1pt] {500M};
      \draw[graytext!60, line width=0.3pt] (-12.00pt, 12.00pt) -- (-15pt, 20pt)
        node[left, font=\tiny, color=graytext, inner sep=1pt] {3B};
    \end{scope}
  \end{tikzpicture}
  \caption{LPIPS vs.\ FPS (lower-right is better)}
\end{subfigure}

\caption{\textbf{Quality vs.\ Inference speed tradeoff on RealEstate10K.} Bubble radius indicates parameter count. OVIE-ft achieves competitive quality at drastically higher FPS than baselines.}
\label{fig:speed}
\end{figure}
Figure~\ref{fig:speed} plots throughput against generation quality (PSNR and LPIPS) on a single H100 GPU on RealEstate10K, with SSIM and FID in Section~\ref{subsec:speed}. \mname achieves 116~FPS (8.6~ms), against 0.19~FPS for VIVID (50 diffusion steps), 0.17~FPS for GeoGPT (autoregressive), 0.024~FPS for PhotoNVS (2000 steps), and 0.016~FPS for ViewCrafter, which also runs DUSt3R reconstruction before 50 steps of video diffusion: more than 600$\times$ faster than the fastest baseline. \mname-ft matches this throughput while surpassing all baselines on RealEstate10K. \mname stores no 3D representation, so the first view costs a single forward pass and every later view costs the same; single-image methods that build an explicit 3D scene pay a one-time per-scene construction first (Section~\ref{subsec:supp_explicit3d}). With keyboard and mouse inputs mapped to extrinsic changes, this throughput makes real-time scene navigation possible (Sections~\ref{subsec:sidebyside} and~\ref{subsec:realtime}).

\subsection{Ablation Studies}
\label{subsec:ablation}

\myparagraph{Loss terms.}
Table~\ref{tab:loss_ablations} shows each term's distinct role. Keeping only $\mathcal{L}_{\text{recon}}$ maximizes PSNR and SSIM while driving FID to 62.1 / 111.0, showing why pixel metrics alone are insufficient. P-DINO and LPIPS contribute complementary perceptual supervision: removing both raises FID to 18.7 / 27.0. The adversarial loss matters most out of domain (FID $+$34.2 on DL3DV against $+$6.3 on RealEstate10K). Section~\ref{subsec:supp_lossablation} shows what each term contributes visually: the perceptual terms carry fine texture, and the adversarial term is what removes seams from the disoccluded region.

\begin{table}[h!]
\centering
\caption{\textbf{Loss ablation studies on RealEstate10K and DL3DV}. Each group varies one
design axis while keeping all others at the default configuration
(\textbf{bold}).}
\label{tab:loss_ablations}
\small
\newcommand{\lon}{\ding{51}}
\newcommand{\loff}{}
\setlength{\tabcolsep}{3pt}
\fitwidth{%
\begin{tabular}{@{}ccc cccc cccc@{}}
\toprule
& & & \multicolumn{4}{c}{\cellcolor{lavblue}RealEstate10K~\cite{re10k}} & \multicolumn{4}{c}{\cellcolor{lavblue}DL3DV~\cite{dl3dv}} \\
\cmidrule(lr){4-7} \cmidrule(lr){8-11}
{\scriptsize P-DINO} & {\scriptsize LPIPS} & {\scriptsize GAN} & PSNR$\uparrow$ & SSIM$\uparrow$ & LPIPS$\downarrow$ & FID$\downarrow$ & PSNR$\uparrow$ & SSIM$\uparrow$ & LPIPS$\downarrow$ & FID$\downarrow$ \\
\midrule
\multicolumn{11}{l}{\textit{Loss component ablation}} \\
\rowcolor{white}
\lon  & \lon  & \lon  & 18.9 & 0.596 & \textbf{0.284} & \textbf{7.12} & 15.0 & 0.373 & \textbf{0.468} & \textbf{14.3} \\
\rowcolor{ltorange}
\loff & \lon  & \lon  & 19.0\up{+0.1} & 0.599\up{+.003} & \underline{0.288}\dn{+.004} & \underline{8.34}\dn{+1.22} & 15.1\up{+0.1} & \underline{0.377}\up{+.004} & \underline{0.472}\dn{+.004} & 15.7\dn{+1.4} \\
\rowcolor{white}
\lon  & \loff & \lon  & 18.7\dn{$-$0.2} & 0.592\dn{$-$.004} & 0.297\dn{+.013} & 8.43\dn{+1.31} & 14.9\dn{$-$0.1} & 0.371\dn{$-$.002} & 0.478\dn{+.010} & \underline{15.3}\dn{+1.0} \\
\rowcolor{ltorange}
\loff & \loff & \lon  & 18.7\dn{$-$0.2} & 0.584\dn{$-$.012} & 0.367\dn{+.083} & 18.7\dn{+11.6} & 14.9\dn{$-$0.1} & 0.368\dn{$-$.005} & 0.540\dn{+.072} & 27.0\dn{+12.7} \\
\rowcolor{white}
\lon  & \lon  & \loff & \underline{19.2}\up{+0.3} & \underline{0.598}\up{+.002} & 0.301\dn{+.017} & 13.4\dn{+6.28} & \underline{15.4}\up{+0.4} & 0.375\up{+.002} & 0.496\dn{+.028} & 48.5\dn{+34.2} \\
\rowcolor{ltorange}
\loff & \loff & \loff & \textbf{19.6}\up{+0.7} & \textbf{0.627}\up{+.031} & 0.416\dn{+.132} & 62.1\dn{+55.0} & \textbf{15.7}\up{+0.7} & \textbf{0.441}\up{+.068} & 0.627\dn{+.159} & 111.0\dn{+96.7} \\
\midrule
\multicolumn{11}{l}{\textit{Reconstruction loss}} \\
\rowcolor{white}
\multicolumn{3}{@{}l}{\quad \textbf{L2}}        & \textbf{18.9} & \textbf{0.596} & \textbf{0.284} & \textbf{7.12} & \textbf{15.0} & \textbf{0.373} & \textbf{0.468} & 14.3 \\
\rowcolor{ltorange}
\multicolumn{3}{@{}l}{\quad L1}                  & 18.5\dn{$-$0.4} & 0.594\dn{$-$.002} & 0.297\dn{+.013} & 8.57\dn{+1.45} & 14.5\dn{$-$0.5} & 0.367\dn{$-$.006} & 0.477\dn{+.009} & 14.3\up{$-$0.0} \\
\rowcolor{white}
\multicolumn{3}{@{}l}{\quad Charbonnier}         & 18.5\dn{$-$0.4} & 0.592\dn{$-$.004} & 0.296\dn{+.012} & 8.35\dn{+1.23} & 14.5\dn{$-$0.5} & 0.367\dn{$-$.006} & 0.476\dn{+.008} & \textbf{14.1}\up{$-$0.2} \\
\bottomrule
\end{tabular}%
}
\end{table}

\myparagraph{Reconstruction loss.}
L2 outperforms L1 and Charbonnier~\cite{charbonnier} across both benchmarks, despite the intuition that robust losses better handle erroneous pseudo-targets.

\myparagraph{Depth estimator and training poses.}
Replacing metric MoGe-2 with the smaller, relative, mask-free Depth Anything~3~\cite{depthanything3} costs 0.28~dB on RealEstate10K and 0.42~dB on DL3DV (Section~\ref{subsec:supp_depth}), so the depth model only fixes the scale convention the network inherits and any strong monocular estimator is interchangeable. The co-visibility distribution of the training-pose sampler trades near-view fidelity against large-motion robustness: the released sampler sits between a low-overlap variant that leads below co-visibility 0.4 and a high-overlap variant that leads above it (Section~\ref{subsec:supp_overlap}).

\subsection{Data Scaling and Diversity}

\myparagraph{Effect of data scale.}
Performance improves rapidly in the first decade of data (3K to 30K images) and keeps improving across every subsequent decade (Figure~\ref{fig:dataset_scale}), well beyond the 10K scenes of dedicated multi-view datasets. SSIM and LPIPS follow the same trend (Section~\ref{subsec:datascaling}).

\begin{figure}[t]
\centering

\definecolor{grayline}{HTML}{B0B0B0}
\definecolor{graytext}{HTML}{505050}
\definecolor{sciblue}{HTML}{4C78A8}
\definecolor{scired}{HTML}{E15759}

\pgfplotsset{
    scalingplot/.style={
        width=\linewidth,
        height=0.55\linewidth,
        grid=major,
        grid style={grayline, line width=0.4pt, densely dotted, opacity=0.6},
        axis x line*=bottom,
        axis y line*=left,
        every outer x axis line/.append style={grayline, line width=0.6pt},
        every outer y axis line/.append style={grayline, line width=0.6pt},
        tick label style={font=\small, color=graytext},
        label style={font=\small},
        xmin=1500, xmax=60000000,
        xtick={3000, 30000, 300000, 3000000, 30000000},
        xticklabels={3K, 30K, 300K, 3M, 30M},
        xlabel={Dataset size},
    }
}

\begin{subfigure}{0.49\textwidth}
\begin{tikzpicture}
  \begin{axis}[
    scalingplot,
    xmode=log,
    ylabel={PSNR $\uparrow$},
    y label style={color=sciblue},
    ymin=17.70, ymax=19.10,
    ytick={17.80, 18.20, 18.60, 19.00},
    yticklabel style={/pgf/number format/.cd, fixed, fixed zerofill, precision=2},
  ]
  \addplot[color=sciblue, mark=*, mark size=2pt, line width=1.5pt,
    mark options={fill=sciblue, draw=white, line width=0.5pt}]
    coordinates {
        (3000,     17.90353491)
        (30000,    18.5887)
        (300000,   18.6680)
        (3000000,  18.7928674901326)
        (30000000, 18.8293834317525)
    };
  \end{axis}
\end{tikzpicture}
\end{subfigure}\hfill
\begin{subfigure}{0.49\textwidth}
\begin{tikzpicture}
  \begin{axis}[
    scalingplot,
    xmode=log,
    ylabel={FID $\downarrow$},
    y label style={color=scired},
    ymin=6.30, ymax=8.80,
    ytick={6.50, 7.20, 7.90, 8.60},
    yticklabel style={/pgf/number format/.cd, fixed, fixed zerofill, precision=2},
  ]
  \addplot[color=scired, mark=*, mark size=2pt, line width=1.5pt,
    mark options={fill=scired, draw=white, line width=0.5pt}]
    coordinates {
        (3000,     8.568372188)
        (30000,    7.2903)
        (300000,   7.2529)
        (3000000,  7.01018865419359)
        (30000000, 6.73781517538327)
    };
  \end{axis}
\end{tikzpicture}
\end{subfigure}

\caption{\textbf{Scaling with dataset size.} PSNR and FID improve as data volume increases.}
\label{fig:dataset_scale}
\end{figure}

\myparagraph{Effect of data diversity.}
To isolate diversity from scale, we train four single-source models and a mixed model on 2M images each. The mixed model matches or exceeds the best single source on both benchmarks (FID 7.08 / 14.2 vs.\ up to 8.59 / 20.3), confirming that diverse mixing yields a modest but reliable gain (Table~\ref{tab:data_coverage}).

\begin{table}[!ht]
\centering
\caption{\textbf{Comparison of data coverage on model performance} on RealEstate10K and DL3DV.
\textbf{Bold}: best; \underline{underline}: second best. Differences are relative to the Mix baseline.}
\label{tab:data_coverage}
\setlength{\tabcolsep}{3pt}
\fitwidth{%
\begin{tabular}{@{}l l cccc cccc@{}}
\toprule
& & \multicolumn{4}{c}{\cellcolor{lavblue}RealEstate10K~\cite{re10k}} & \multicolumn{4}{c}{\cellcolor{lavblue}DL3DV~\cite{dl3dv}} \\
\cmidrule(lr){3-6} \cmidrule(lr){7-10}
Dataset & Domain & PSNR$\uparrow$ & SSIM$\uparrow$ & LPIPS$\downarrow$ & FID$\downarrow$ & PSNR$\uparrow$ & SSIM$\uparrow$ & LPIPS$\downarrow$ & FID$\downarrow$ \\
\midrule
\rowcolor{white}
Mix        & Mixed       & \textbf{18.8} & \underline{0.595} & \textbf{0.284} & \textbf{7.08} & \textbf{15.0} & \textbf{0.372} & \textbf{0.467} & \textbf{14.2} \\
\rowcolor{ltorange}
OSV5M~\cite{osv5m}       & Street View & \underline{18.2}\dn{$-$0.6} & 0.566\dn{$-$.029} & 0.318\dn{+.034} & 8.59\dn{+1.51} & \textbf{15.0} & 0.369\dn{$-$.003} & 0.481\dn{+.014} & 18.9\dn{+4.7} \\
\rowcolor{white}
ImageNet21K~\cite{in21k} & Objects     & \textbf{18.8} & 0.593\dn{$-$.002} & 0.289\dn{+.005} & 7.32\dn{+0.24} & \underline{14.9}\dn{$-$0.1} & \underline{0.370}\dn{$-$.003} & 0.472\dn{+.005} & 15.9\dn{+1.7} \\
\rowcolor{ltorange}
Places~\cite{places}      & Scenes      & \textbf{18.8} & \textbf{0.596}\up{+.001} & \underline{0.286}\dn{+.002} & 7.41\dn{+0.33} & 14.8\dn{$-$0.2} & 0.367\dn{$-$.006} & 0.477\dn{+.010} & 20.3\dn{+6.1} \\
\rowcolor{white}
OpenImages~\cite{openimages}  & General     & \textbf{18.8} & 0.593\dn{$-$.002} & 0.287\dn{+.003} & \underline{7.20}\dn{+0.12} & \underline{14.9}\dn{$-$0.1} & \underline{0.370}\dn{$-$.003} & \underline{0.469}\dn{+.002} & \underline{14.8}\dn{+0.6} \\
\bottomrule
\end{tabular}%
}
\end{table}

\section{Conclusion}
\label{sec:conclusion}

We introduce monocular pretraining for novel view synthesis: a frozen depth estimator constructs pseudo-pairs from any single image, masked losses restrict supervision to valid regions, and an adversarial objective covers disoccluded areas, reducing depth to a training scaffold discarded at inference. Trained on 30 million uncurated internet images without multi-view supervision, \mname rivals specialized in-domain baselines under domain shift at 116~FPS, and brief fine-tuning suffices to surpass all geometry-free methods. Internet-scale single-image collections are therefore a viable foundation for the field, and domains with no multi-view counterpart, such as historical photographs and paintings, become training sources.

\myparagraph{Limitations.}
Performance is bounded by the depth estimator: failures on reflective surfaces, transparent materials, or low light propagate as missing regions that masking cannot recover. Disoccluded content is hallucinated and can differ from the true scene; cross-view consistency does not make it correct. \mname covers moderate viewpoint change, degrading smoothly as co-visibility drops, and makes no claim to 360-degree or rear-view synthesis. Output resolution is fixed at training time, so unlike methods that render from an explicit 3D representation, \mname cannot produce arbitrary resolutions at inference.

\begin{ack}
We thank Robin Courant for proof-reading, and Eloi Alonso, Mathieu Aubry, Antoine Guédon, Anthony Hu, Loïc Landrieu, Vincent Lepetit, Vincent Micheli, Manu Orsini, Amélie Royer, and Václav Volhejn for interesting discussions. We also thank the anonymous reviewers and the area chair for feedback that improved the paper.
\end{ack}

\bibliographystyle{abbrvnat}
\bibliography{sections/references}

\appendix

\section{Summary of Supplementary Material}
\label{sec:supp_summary}

This supplementary document provides additional details and results that complement the main paper. It is organized as follows:
\begin{itemize}
    \item \textbf{\Cref{sec:appendix_quantitative} -- Additional Quantitative Results.} We report supplementary SSIM and LPIPS curves for the data-scaling (\ref{subsec:datascaling}) and SSIM and FID curves for the throughput analysis (\ref{subsec:speed}), and the effect of domain-specific finetuning on DL3DV~\cite{dl3dv} (\ref{subsec:dl3dv_finetuning}). We then detail the evaluation protocol, covering the per-scene scale sweep and the per-method resolutions (\ref{subsec:supp_protocol}), and measure geometric consistency with MEt3R~\cite{met3r} (\ref{subsec:supp_met3r}) and with a round-trip test (\ref{subsec:supp_roundtrip}). We quantify the operating envelope through camera displacement and co-visibility, including a comparison against all baselines under matched motion (\ref{subsec:supp_envelope}). Two controls follow: a depth-estimator ablation (\ref{subsec:supp_depth}) and an ablation of the training-pose overlap (\ref{subsec:supp_overlap}). We then report \mname-512, the $512{\times}512$ model, with quantitative and qualitative results (\ref{subsec:supp_resolution}), compare against single-image methods that build an explicit 3D representation (\ref{subsec:supp_explicit3d}), and evaluate \mname under VIVID's original protocol (\ref{subsec:supp_vividprotocol}).
    \item \textbf{\Cref{sec:appendix_sampling} -- Camera Sampling Details.} We describe the distribution from which relative camera transformations are sampled during training, including the parameterization (\ref{subsec:detailed_sampling_methods}, \ref{subsec:lookat}), the geometry-aware reprojection (\ref{subsec:reprojection}) and pseudo-views sampling hyperparameters (\ref{subsec:sampling_methods_hyperparameters}).
    \item \textbf{\Cref{sec:appendix_implementation} -- Additional Implementation Details.} We provide extended details on architecture (\ref{subsec:supp_arch}), pose-conditioning (\ref{subsec:supp_adaln}), and training details and hyperparameters (\ref{subsec:supp_hyperparameters}).
    \item \textbf{\Cref{sec:appendix_resources} -- Training Resources and Broader Impact.} We report compute requirements and discuss the broader societal and environmental implications of the work.
    \item \textbf{\Cref{sec:appendix_qualitative} -- Additional Qualitative Results.} We illustrate OVIE's robust generalization by synthesizing additional novel views from diverse out-of-distribution images, including non-realistic source inputs such as paintings (\ref{subsec:ood}). We compare OVIE-generated views against their corresponding pseudo-ground-truth supervision targets (\ref{subsec:pseudotargets}). We present more qualitative comparisons on RealEstate10K~\cite{re10k}, contrasting OVIE's outputs with the source, the ground-truth novel view, and baseline methods (\ref{subsec:baselines}). To highlight generation consistency and responsiveness, we provide animated side-by-side navigations (\ref{subsec:sidebyside}) and real-time interactive screen recordings, driven by mouse and keyboard inputs, all generated continuously from a single initial frame (\ref{subsec:realtime}).
    \item \textbf{\Cref{sec:appendix_infinitenaturezero} -- Additional Comparison to InfiniteNature-Zero~\cite{infinitenaturezero}.} We compare OVIE to InfiniteNature-Zero. While omitted from the baselines because its inference requires geometry-based depth estimation and sky segmentation, it shares OVIE's approach of training entirely on single-view images.
\end{itemize}

\section{Additional Quantitative Results}
\label{sec:appendix_quantitative}

\subsection{Effect of Dataset Size on SSIM and LPIPS}
\label{subsec:datascaling}
Figure~\ref{fig:supp_ssim_lpips_dataset_size} shows SSIM and LPIPS as functions of dataset size, confirming the trends observed for PSNR and FID in Figure~\ref{fig:dataset_scale} of the main paper: both metrics improve consistently with more training data.

\begin{figure}[t]
\centering

\definecolor{grayline}{HTML}{B0B0B0}
\definecolor{graytext}{HTML}{505050}
\definecolor{scigreen}{HTML}{59A14F}
\definecolor{sciorange}{HTML}{F28E2B}

\pgfplotsset{
    scalingplot/.style={
        width=\linewidth,
        height=0.75\linewidth,
        grid=major,
        grid style={grayline, line width=0.4pt, densely dotted, opacity=0.6},
        axis x line*=bottom,
        axis y line*=left,
        every outer x axis line/.append style={grayline, line width=0.6pt},
        every outer y axis line/.append style={grayline, line width=0.6pt},
        tick label style={font=\small, color=graytext},
        label style={font=\small},
        xmin=1500, xmax=60000000,
        xtick={3000, 30000, 300000, 3000000, 30000000},
        xticklabels={3K, 30K, 300K, 3M, 30M},
        xlabel={Dataset size},
    }
}

\begin{subfigure}{0.48\textwidth}
\begin{tikzpicture}
  \begin{axis}[
    scalingplot,
    xmode=log,
    ylabel={SSIM $\uparrow$},
    y label style={color=scigreen},
    ymin=0.555, ymax=0.610,
    ytick={0.560, 0.575, 0.590, 0.605},
    yticklabel style={/pgf/number format/.cd, fixed, fixed zerofill, precision=3},
  ]
  \addplot[color=scigreen, mark=*, mark size=2pt, line width=1.5pt,
    mark options={fill=scigreen, draw=white, line width=0.5pt}]
    coordinates {
        (3000,     0.5611357346)
        (30000,    0.587771310184683) 
        (300000,   0.588615399730702) 
        (3000000,  0.599569534528468) 
        (30000000, 0.602569149279168)
    };
  \end{axis}
\end{tikzpicture}
\end{subfigure}\hfill
\begin{subfigure}{0.48\textwidth}
\begin{tikzpicture}
  \begin{axis}[
    scalingplot,
    xmode=log,
    ylabel={LPIPS $\downarrow$},
    y label style={color=sciorange},
    ymin=0.270, ymax=0.325,
    ytick={0.275, 0.290, 0.305, 0.320},
    yticklabel style={/pgf/number format/.cd, fixed, fixed zerofill, precision=3},
  ]
  \addplot[color=sciorange, mark=*, mark size=2pt, line width=1.5pt,
    mark options={fill=sciorange, draw=white, line width=0.5pt}]
    coordinates {
        (3000,     0.3186951026)
        (30000,    0.2928878586) 
        (300000,   0.2905903265) 
        (3000000,  0.283744703233241) 
        (30000000, 0.2786194687)
    };
  \end{axis}
\end{tikzpicture}
\end{subfigure}

\caption{\textbf{Scaling with dataset size -- SSIM and LPIPS.} Complementary to Figure~\ref{fig:dataset_scale} in the main paper, SSIM and LPIPS on RealEstate10K~\cite{re10k} follow the same monotonic improvement as data volume increases.}
\label{fig:supp_ssim_lpips_dataset_size}
\end{figure}

\begin{figure}[t]
\centering

\definecolor{grayline}{HTML}{B0B0B0}
\definecolor{graytext}{HTML}{505050}
\definecolor{geoblue}{HTML}{4C78A8}
\definecolor{vividorange}{HTML}{F28E2B}
\definecolor{oursgreen}{HTML}{59A14F}
\definecolor{oursgreenft}{HTML}{1B5E20}
\definecolor{photonvsgray}{HTML}{76B7B2}
\definecolor{vcred}{HTML}{E15759}

\begin{subfigure}[t]{0.48\textwidth}
  \centering
  \begin{tikzpicture}
    \begin{axis}[
      width=6.5cm,
      height=5.5cm,
      grid=major,
      grid style={grayline, line width=0.4pt, densely dotted, opacity=0.6},
      axis x line*=bottom,
      axis y line*=left,
      every outer x axis line/.append style={grayline, line width=0.6pt},
      every outer y axis line/.append style={grayline, line width=0.6pt},
      tick label style={font=\small, color=graytext},
      label style={font=\normalsize},
      clip=false,
      xmode=log,
      xlabel={Inference Speed (FPS)},
      ylabel={SSIM $\uparrow$},
      xmin=0.01, xmax=300,
      ymin=0.44, ymax=0.72,
      ytick={0.48, 0.54, 0.60, 0.66, 0.72},
      yticklabel style={/pgf/number format/.cd, fixed, fixed zerofill, precision=2},
      log ticks with fixed point,
    ]
    \addplot[only marks, mark=*, mark size=12.00pt, fill=vcred, draw=white, line width=0.7pt, fill opacity=0.85]
      coordinates {(0.016, 0.559)} node[above, font=\footnotesize, text=vcred, yshift=10pt,
      fill=white, fill opacity=0.7, text opacity=1, rounded corners=1pt, inner sep=2pt] {ViewCrafter};
    \addplot[only marks, mark=*, mark size=4.96pt, fill=geoblue, draw=white, line width=0.7pt, fill opacity=0.85]
      coordinates {(0.172, 0.480)} node[below, font=\footnotesize, text=geoblue, yshift=-4pt,
      fill=white, fill opacity=0.7, text opacity=1, rounded corners=1pt, inner sep=2pt] {GeoGPT};
    \addplot[only marks, mark=*, mark size=4.49pt, fill=vividorange, draw=white, line width=0.7pt, fill opacity=0.85]
      coordinates {(0.190, 0.661)} node[above, font=\footnotesize, text=vividorange, yshift=4pt,
      fill=white, fill opacity=0.7, text opacity=1, rounded corners=1pt, inner sep=2pt] {VIVID};
    \addplot[only marks, mark=*, mark size=2.52pt, fill=photonvsgray, draw=white, line width=0.7pt, fill opacity=0.85]
      coordinates {(0.024, 0.601)} node[right, font=\footnotesize, text=photonvsgray, xshift=4pt,
      fill=white, fill opacity=0.7, text opacity=1, rounded corners=1pt, inner sep=2pt] {PhotoNVS};
    \addplot[only marks, mark=*, mark size=2.62pt, fill=oursgreen, draw=white, line width=0.7pt, fill opacity=0.85]
      coordinates {(115.74, 0.602)} node[below left, font=\footnotesize\bfseries, text=oursgreen, xshift=-2pt, yshift=-2pt,
      fill=white, fill opacity=0.7, text opacity=1, rounded corners=1pt, inner sep=2pt] {OVIE};
    \addplot[only marks, mark=*, mark size=2.62pt, fill=oursgreenft, draw=white, line width=0.7pt, fill opacity=0.85]
      coordinates {(115.74, 0.695)} node[above left, font=\footnotesize\bfseries, text=oursgreenft, xshift=-2pt, yshift=2pt,
      fill=white, fill opacity=0.7, text opacity=1, rounded corners=1pt, inner sep=2pt] {OVIE-ft};
    \end{axis}
  \end{tikzpicture}
  \caption{SSIM vs.\ FPS (upper-right is better)}
\end{subfigure}
\hfill
\begin{subfigure}[t]{0.48\textwidth}
  \centering
  \begin{tikzpicture}
    \begin{axis}[
      width=6.5cm,
      height=5.5cm,
      grid=major,
      grid style={grayline, line width=0.4pt, densely dotted, opacity=0.6},
      axis x line*=bottom,
      axis y line*=left,
      every outer x axis line/.append style={grayline, line width=0.6pt},
      every outer y axis line/.append style={grayline, line width=0.6pt},
      tick label style={font=\small, color=graytext},
      label style={font=\normalsize},
      clip=false,
      xmode=log,
      xlabel={Inference Speed (FPS)},
      ylabel={FID $\downarrow$},
      xmin=0.01, xmax=300,
      ymin=2, ymax=22,
      ytick={4, 8, 12, 16, 20},
      log ticks with fixed point,
    ]
    \addplot[only marks, mark=*, mark size=12.00pt, fill=vcred, draw=white, line width=0.7pt, fill opacity=0.85]
      coordinates {(0.016, 9.98)} node[right, font=\footnotesize, text=vcred, xshift=10pt,
      fill=white, fill opacity=0.7, text opacity=1, rounded corners=1pt, inner sep=2pt] {ViewCrafter};
    \addplot[only marks, mark=*, mark size=4.96pt, fill=geoblue, draw=white, line width=0.7pt, fill opacity=0.85]
      coordinates {(0.172, 18.0)} node[above, font=\footnotesize, text=geoblue, yshift=4pt,
      fill=white, fill opacity=0.7, text opacity=1, rounded corners=1pt, inner sep=2pt] {GeoGPT};
    \addplot[only marks, mark=*, mark size=4.49pt, fill=vividorange, draw=white, line width=0.7pt, fill opacity=0.85]
      coordinates {(0.190, 4.26)} node[below, font=\footnotesize, text=vividorange, yshift=-4pt,
      fill=white, fill opacity=0.7, text opacity=1, rounded corners=1pt, inner sep=2pt] {VIVID};
    \addplot[only marks, mark=*, mark size=2.52pt, fill=photonvsgray, draw=white, line width=0.7pt, fill opacity=0.85]
      coordinates {(0.024, 10.6)} node[above, font=\footnotesize, text=photonvsgray, yshift=4pt,
      fill=white, fill opacity=0.7, text opacity=1, rounded corners=1pt, inner sep=2pt] {PhotoNVS};
    \addplot[only marks, mark=*, mark size=2.62pt, fill=oursgreen, draw=white, line width=0.7pt, fill opacity=0.85]
      coordinates {(115.74, 6.74)} node[above left, font=\footnotesize\bfseries, text=oursgreen, xshift=-2pt, yshift=2pt,
      fill=white, fill opacity=0.7, text opacity=1, rounded corners=1pt, inner sep=2pt] {OVIE};
    \addplot[only marks, mark=*, mark size=2.62pt, fill=oursgreenft, draw=white, line width=0.7pt, fill opacity=0.85]
      coordinates {(115.74, 5.59)} node[below left, font=\footnotesize\bfseries, text=oursgreenft, xshift=-2pt, yshift=-2pt,
      fill=white, fill opacity=0.7, text opacity=1, rounded corners=1pt, inner sep=2pt] {OVIE-ft};
    \end{axis}
  \end{tikzpicture}
  \caption{FID vs.\ FPS (lower-right is better)}
\end{subfigure}

\caption{\textbf{Quality vs.\ Inference tradeoff on RealEstate10K -- SSIM and FID.} Complementary to Figure~\ref{fig:speed} in the main paper. Bubble radius indicates parameter count. The same trend holds: \mname is faster and better-performing than baseline methods.}
\label{fig:supp_throughput}
\end{figure}

\subsection{Throughput analysis}
\label{subsec:speed}
Figure~\ref{fig:supp_throughput} demonstrates that \mname is 600$\times$ faster than the fastest baseline while also achieving improved SSIM and FID scores on RealEstate10K~\cite{re10k}. This complements the PSNR and LPIPS results discussed in Figure~\ref{fig:speed} of the main paper.

\subsection{Domain-specific finetuning on DL3DV}
\label{subsec:dl3dv_finetuning}
In the main paper we report OVIE-ft, the base model finetuned on RealEstate10K~\cite{re10k}. For completeness, we apply the identical protocol to finetune the base \mname model on DL3DV~\cite{dl3dv}. Finetuning markedly improves in-domain reconstruction on DL3DV relative to the base model, raising PSNR from 14.8 to 17.07, SSIM from 0.369 to 0.433, and lowering LPIPS from 0.464 to 0.368; FID, however, rises from 13.6 to 17.24, suggesting that domain-specific finetuning favors faithful per-view reconstruction at a slight cost to global distributional realism. More notably, this DL3DV-finetuned variant retains strong performance on RealEstate10K even though that benchmark is now out-of-distribution: it matches or slightly surpasses the base model on every reconstruction metric (PSNR 19.34 vs.\ 18.8, SSIM 0.608 vs.\ 0.602, LPIPS 0.260 vs.\ 0.279), with only a marginal increase in FID (7.14 vs.\ 6.74). This indicates that finetuning sharpens in-domain fidelity without eroding, and in this case slightly improving, cross-domain generalization.

\subsection{Evaluation protocol details}
\label{subsec:supp_protocol}

\paragraph{Per-scene scale sweep.}
Benchmark poses come from SLAM and carry an unknown global scale, so a translation in benchmark units has no fixed physical meaning. For each geometry-free method (\mname, VIVID~\cite{vivid}, PhotoNVS~\cite{photonvs}, GeoGPT~\cite{geogpt}) we sweep a per-scene factor on the input translation and keep the value maximizing that method's own PSNR. ViewCrafter~\cite{viewcrafter} accepts no SE(3) pose: its single-image controls are parametric (elevation, azimuth, radius, pan) and it recovers the target cameras with DUSt3R~\cite{dust3r} in DUSt3R's own frame and scale, so it runs unswept, as a greyed geometry-based reference.

\paragraph{Resolution.}
Methods differ in native input and output resolution (Table~\ref{tab:resolution_protocol}). Every prediction and the corresponding ground truth are brought to $256{\times}256$ identically (ADM~\cite{adm} center-crop; GeoGPT is matched at its native $208{\times}368$ aspect ratio, then cropped), so every reported number is computed at a common resolution.

\begin{table}[h]
\centering
\caption{\textbf{Input and output resolution of each evaluated method.} All metrics are computed at $256{\times}256$.}
\label{tab:resolution_protocol}
\setlength{\tabcolsep}{5pt}
\fitwidth{%
\begin{tabular}{@{}l l l l@{}}
\toprule
Method & Input & Native output & Scored at \\
\midrule
\mname                    & Single $256{\times}256$ image                                              & $256{\times}256$                    & $256{\times}256$ \\
SHARP~\cite{sharp}        & Single $256{\times}256$ image (center-crop, below SHARP's native $1536$)   & 3DGS render, arbitrary resolution   & $256{\times}256$ \\
FlexWorld~\cite{flexworld}& Single image at 16:9; video diffusion stage at $480{\times}720$           & 3DGS render, $288{\times}512$       & $256{\times}256$ center-crop \\
ViewCrafter~\cite{viewcrafter} & Single image and its DUSt3R~\cite{dust3r} reconstruction              & $576{\times}1024$                   & $256{\times}256$ center-crop \\
VIVID~\cite{vivid}        & Single image                                                              & $256{\times}256$                    & $256{\times}256$ \\
PhotoNVS~\cite{photonvs}  & Single image                                                              & $256{\times}256$                    & $256{\times}256$ \\
GeoGPT~\cite{geogpt}      & Single image                                                              & $208{\times}368$                    & $256{\times}256$ \\
\bottomrule
\end{tabular}}
\end{table}

\subsection{Multi-view consistency}
\label{subsec:supp_met3r}
MEt3R~\cite{met3r} reconstructs the shared 3D geometry of each generated frame pair with MASt3R~\cite{mast3r} and scores feature dissimilarity in the overlapping region. Table~\ref{tab:supp_met3r} extends the MEt3R columns of Table~\ref{tab:comparison_with_sota} with SHARP, Infinite Nature-Zero, the in-domain DL3DV checkpoint, and pairs of real target frames. Every \mname variant leads every geometry-free baseline on both benchmarks. \mname-ft is the most consistent method on both benchmarks, ahead of the geometry-dependent Infinite Nature-Zero and of SHARP~\cite{sharp}, which renders every view from a single shared explicit 3D representation; base \mname ties SHARP on RealEstate10K and trails it on DL3DV. The real-pair reference carries the properties of real video: moving content, exposure shifts, and cuts at clip boundaries all raise MEt3R. \mname-ft scores below that reference on both benchmarks, 0.029 against 0.042 and 0.040 against 0.070.

\begin{table}[h]
\centering
\caption{\textbf{Multi-view 3D consistency on RealEstate10K and DL3DV.}
MEt3R~\cite{met3r}, lower is better, over 750 scenes per benchmark.
\textbf{Bold}: best; \underline{underline}: second best.
Geometry-based methods estimate depth or a point cloud at inference; they are greyed out and excluded from the ranking for fairness.
$^*$Trained in-domain on DL3DV: its DL3DV score is greyed out and excluded from the ranking, as in Table~\ref{tab:comparison_with_sota}.}
\label{tab:supp_met3r}
\setlength{\tabcolsep}{6pt}
\begin{tabular}{@{}l cc@{}}
\toprule
Method & \cellcolor{lavblue}RealEstate10K~\cite{re10k} & \cellcolor{lavblue}DL3DV~\cite{dl3dv} \\
\midrule
\multicolumn{3}{l}{\textit{Geometry-free}} \\
\rowcolor{white}
\quad GeoGPT~\cite{geogpt}    & 0.184 & 0.300 \\
\rowcolor{ltorange}
\quad PhotoNVS~\cite{photonvs}& 0.139 & 0.301 \\
\rowcolor{white}
\quad VIVID~\cite{vivid}      & 0.076 & 0.169 \\
\midrule
\multicolumn{3}{l}{\textit{Geometry-based references}} \\
\rowcolor{ltorange}
\quad ViewCrafter~\cite{viewcrafter}          & \textcolor{gray}{0.193} & \textcolor{gray}{0.261} \\
\rowcolor{white}
\quad Infinite Nature-Zero~\cite{infinitenaturezero} & \textcolor{gray}{0.055} & \textcolor{gray}{0.098} \\
\rowcolor{ltorange}
\quad SHARP~\cite{sharp}                      & \textcolor{gray}{0.035} & \textcolor{gray}{0.046} \\
\midrule
\rowcolor{white}
\mname (ours)                                & 0.035 & \underline{0.078} \\
\rowcolor{ltorange}
\mname-ft, RealEstate10K (ours)              & \underline{0.029} & \textbf{0.040} \\
\rowcolor{white}
\mname, DL3DV-finetuned$^*$ (ours)           & \textbf{0.028} & \textcolor{gray}{0.059} \\
\midrule
\rowcolor{ltorange}
Real target frame pairs (reference) & \textcolor{gray}{0.042} & \textcolor{gray}{0.070} \\
\bottomrule
\end{tabular}
\end{table}

\subsection{Generated trajectories}
\label{subsec:supp_trajectory}
Figure~\ref{fig:supp_trajectory} shows \mname's generations along the camera path of five evaluation sequences. Each view is produced independently from the first frame of the sequence and its target pose, so any drift between views would be visible as content changing from one column to the next. Content instead stays stable along each path, including where the motion disoccludes new regions, which is the qualitative counterpart to the MEt3R scores above.

\ifdefined\trImgW\else\newlength{\trImgW}\fi\setlength{\trImgW}{2.3cm}
\ifdefined\trGap\else\newlength{\trGap}\fi\setlength{\trGap}{0.8pt}
\newcommand{\trImg}[1]{\includegraphics[width=\trImgW]{#1}}

\begin{figure}[t]
\centering
\setlength{\tabcolsep}{0pt}%
\begin{tabular}{@{}c@{\hspace{\trGap}} c@{\hspace{\trGap}} c@{\hspace{\trGap}} c@{\hspace{\trGap}} c@{\hspace{\trGap}} c@{}}
\scriptsize view 1 & \scriptsize view 4 & \scriptsize view 7 &
\scriptsize view 10 & \scriptsize view 13 & \scriptsize view 15 \\[\trGap]
\trImg{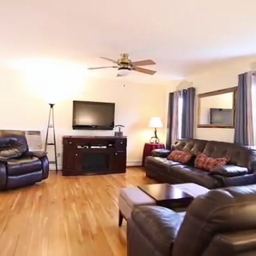} & \trImg{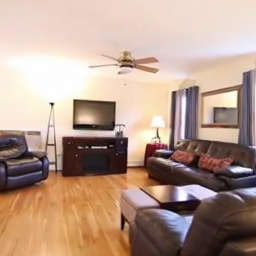} & \trImg{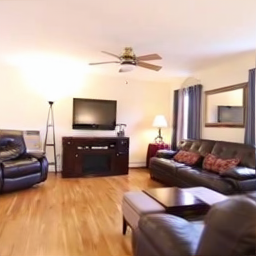} &
\trImg{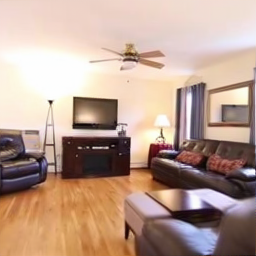} & \trImg{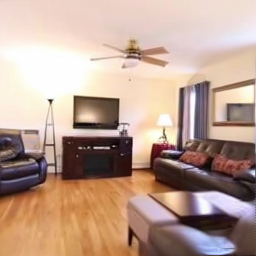} & \trImg{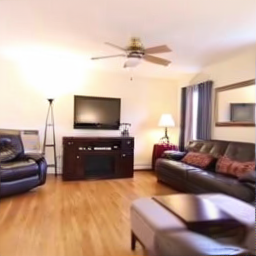} \\[\trGap]
\trImg{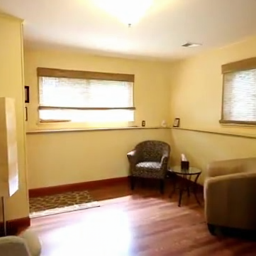} & \trImg{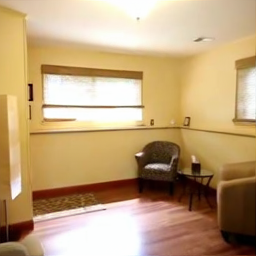} & \trImg{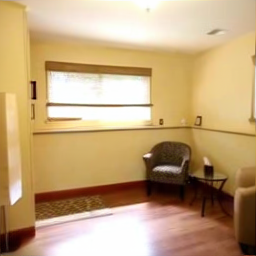} &
\trImg{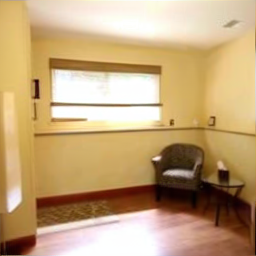} & \trImg{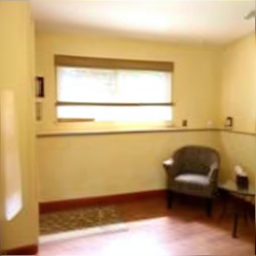} & \trImg{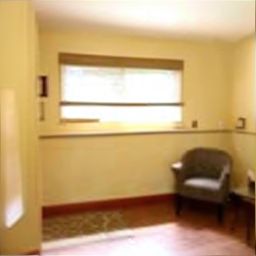} \\[\trGap]
\trImg{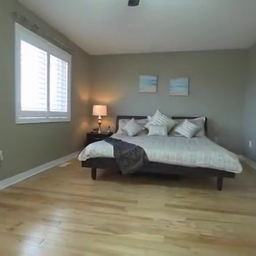} & \trImg{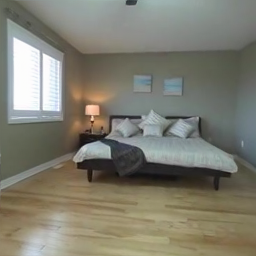} & \trImg{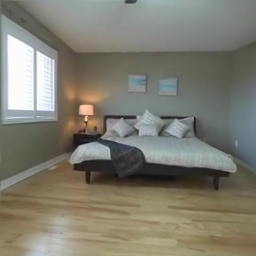} &
\trImg{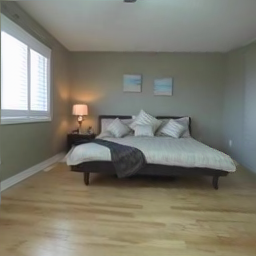} & \trImg{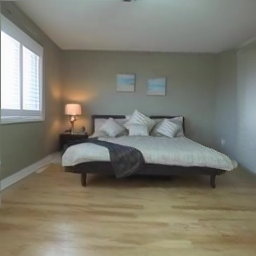} & \trImg{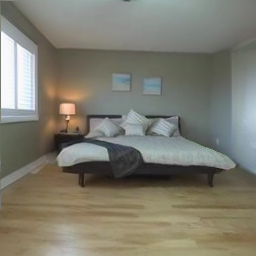} \\[\trGap]
\trImg{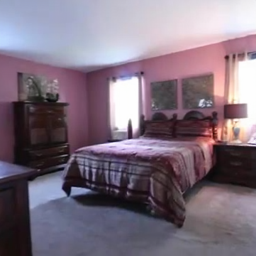} & \trImg{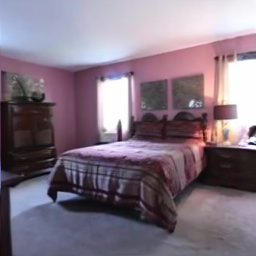} & \trImg{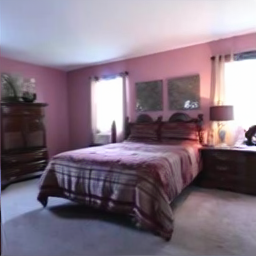} &
\trImg{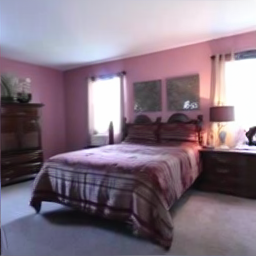} & \trImg{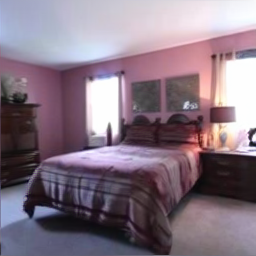} & \trImg{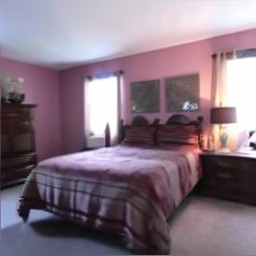} \\[\trGap]
\trImg{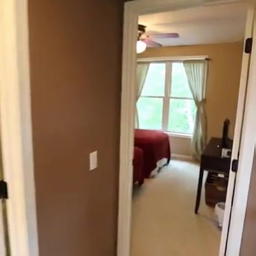} & \trImg{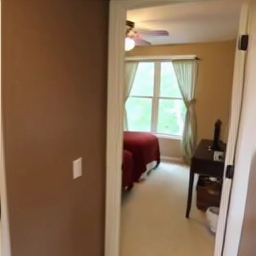} & \trImg{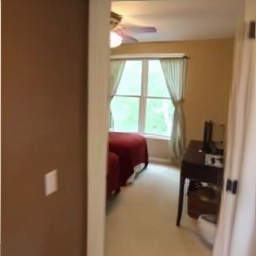} &
\trImg{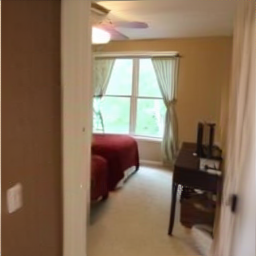} & \trImg{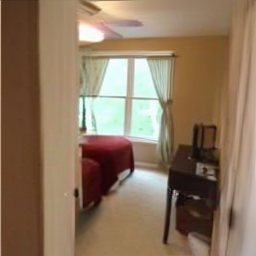} & \trImg{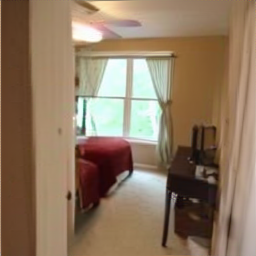} \\[\trGap]
\end{tabular}
\caption{\textbf{Generated trajectories on RealEstate10K.} Five evaluation sequences, each shown at
six points along the camera path. Every view is produced independently from the \emph{first} frame
of the sequence and its target pose, with no access to any intermediate ground-truth frame and no
recurrence between views. Content stays stable along each path, including in regions disoccluded by
the motion, which is the qualitative counterpart to the MEt3R scores in \Cref{subsec:supp_met3r}.
These are the released model's evaluation outputs, so they are the same generations the reported
metrics are computed on.}
\label{fig:supp_trajectory}
\end{figure}

\subsection{Round-trip consistency}
\label{subsec:supp_roundtrip}
Applying a pose change and then its exact reverse tests consistency a second way, comparing the twice-generated image to the source on the pixels carried through both passes (Table~\ref{tab:roundtrip}). Motion is set by the source-to-target frame offset on RealEstate10K~\cite{re10k}. Offset 0 still passes the image through the generator twice, so it is a no-motion reference and not a lossless bound. The round trip holds near 22--23~dB through moderate motion and decays gracefully as the carried-through region shrinks, averaging 20.8~dB over the motion-bearing offsets.

\begin{table}[h]
\centering
\caption{\textbf{Round-trip consistency on RealEstate10K.} A pose change followed by its reverse, scored on the pixels visible in both passes. The co-visible fraction is the share of the source still visible after the motion.}
\label{tab:roundtrip}
\setlength{\tabcolsep}{8pt}
\begin{tabular}{@{}l ccccc@{}}
\toprule
Motion offset (frames) & 0 & 3 & 15 & 45 & 90 \\
\midrule
\rowcolor{white}
Round-trip PSNR\,$\uparrow$ & 22.2 & 22.9 & 22.2 & 20.0 & 18.1 \\
\rowcolor{ltorange}
Co-visible fraction         & 0.99 & 0.94 & 0.80 & 0.58 & 0.41 \\
\bottomrule
\end{tabular}
\end{table}

\subsection{Operating envelope: displacement and co-visibility}
\label{subsec:supp_envelope}
\mname targets moderate viewpoint change. We quantify that range three ways: rotation in degrees, translation as a percentage of median scene depth, which makes the baseline scale-invariant, and co-visibility, the fraction of the target reconstructible from the source, \ie one minus disocclusion. Because both benchmarks are scale-ambiguous, translation and co-visibility use the per-scene scale selected at evaluation (\Cref{subsec:supp_protocol}).

\paragraph{Displacement distributions.}
Figure~\ref{fig:supp_displacement} reports the distributions over all evaluation pairs. The training mix spans a wider range than either benchmark: median rotation $7.3^\circ$ against $2.1^\circ$ and $5.6^\circ$, translation 90th percentile 68\% of scene depth, median co-visibility 0.43 against 0.80 and 0.72. At the last generated frame the median displacement grows to $5.6^\circ$ and 11\% of scene depth on RealEstate10K and to $21.5^\circ$ and 13\% on DL3DV, with median co-visibility falling to 0.64 and 0.50. DL3DV at stride 1 therefore exercises larger motion than RealEstate10K at stride 3.

\begin{figure}[!ht]
\centering

\pgfplotsset{
    dispbox/.style={
        width=\linewidth,
        height=0.62\linewidth,
        boxplot/draw direction=y,
        grid=major,
        grid style={grayline, line width=0.4pt, densely dotted, opacity=0.6},
        axis x line*=bottom,
        axis y line*=left,
        every outer x axis line/.append style={grayline, line width=0.6pt},
        every outer y axis line/.append style={grayline, line width=0.6pt},
        tick label style={font=\small, color=graytext},
        label style={font=\small},
        xtick={1,2,3},
        xticklabels={Training mix, RealEstate10K, DL3DV},
        xticklabel style={font=\scriptsize, color=graytext, rotate=20, anchor=east,
                          yshift=-4pt},
        ymin=0,
    }
}

\begin{subfigure}{0.49\textwidth}
\begin{tikzpicture}
  \begin{axis}[dispbox, ylabel={Rotation ($^\circ$)}, ymax=31, ytick={0,10,20,30}]
    \addplot[boxplot prepared={lower whisker=0.0, lower quartile=0.0, median=7.3,
        upper quartile=16.7, upper whisker=28.7}, color=sciblue, fill=sciblue!22,
        solid, line width=0.8pt] coordinates {};
    \addplot[boxplot prepared={lower whisker=0.1, lower quartile=0.5, median=2.1,
        upper quartile=5.9, upper whisker=10.6}, color=scired, fill=scired!22,
        solid, line width=0.8pt] coordinates {};
    \addplot[boxplot prepared={lower whisker=0.8, lower quartile=2.1, median=5.6,
        upper quartile=15.2, upper whisker=27.1}, color=scigreen, fill=scigreen!22,
        solid, line width=0.8pt] coordinates {};
  \end{axis}
\end{tikzpicture}
\end{subfigure}\hfill
\begin{subfigure}{0.49\textwidth}
\begin{tikzpicture}
  \begin{axis}[dispbox, ylabel={Translation (\% of scene depth)}, ymax=74, ytick={0,20,40,60}]
    \addplot[boxplot prepared={lower whisker=0.0, lower quartile=0.0, median=22.0,
        upper quartile=41.9, upper whisker=68.4}, color=sciblue, fill=sciblue!22,
        solid, line width=0.8pt] coordinates {};
    \addplot[boxplot prepared={lower whisker=0.4, lower quartile=1.4, median=4.3,
        upper quartile=10.3, upper whisker=18.1}, color=scired, fill=scired!22,
        solid, line width=0.8pt] coordinates {};
    \addplot[boxplot prepared={lower whisker=0.0, lower quartile=0.5, median=5.2,
        upper quartile=15.3, upper whisker=31.9}, color=scigreen, fill=scigreen!22,
        solid, line width=0.8pt] coordinates {};
  \end{axis}
\end{tikzpicture}
\end{subfigure}

\caption{\textbf{Camera displacement distributions} for the training mix and for both benchmarks,
over all pairs. Boxes span the first and third quartiles around the median; whiskers mark the
10th and 90th percentiles. The training mix spans a wider range of both rotation and translation
than either benchmark, so evaluation stays inside the training envelope.}
\label{fig:supp_displacement}
\end{figure}

\paragraph{Degradation beyond the envelope.}
Binning RealEstate10K pairs by co-visibility gives the degradation curve, and since stride acts on quality only through the overlap it induces, the same binning captures the effect of stride. Full-frame PSNR falls monotonically with co-visibility (correlation 0.66) without collapsing (Figure~\ref{fig:supp_covisibility}). On co-visible pixels alone, which separate reconstruction of observed content from hallucination of disoccluded content, \mname beats its warp teacher in every bin below 0.9 and loses in the near-copy regime above it, so pooled over all pairs the two tie at 19.5 against 19.7~dB.

\paragraph{Baselines under matched motion.}
Binning by co-visibility also compares methods at matched motion instead of pooling easy and hard pairs together (Figure~\ref{fig:supp_covis_matched}). \mname-ft leads in every bin, and its margin widens as motion grows: at the largest motions PhotoNVS collapses to 7.57~dB while \mname-ft holds 13.18~dB. Zero-shot base \mname also holds up against the in-domain baselines under large motion, for instance 13.85 against VIVID's 13.47 in the 0.1--0.2 bin.

\begin{figure}[!ht]
\centering

\pgfplotsset{
    covisplot/.style={
        width=\linewidth,
        height=0.72\linewidth,
        grid=major,
        grid style={grayline, line width=0.4pt, densely dotted, opacity=0.6},
        axis x line*=bottom,
        axis y line*=left,
        every outer x axis line/.append style={grayline, line width=0.6pt},
        every outer y axis line/.append style={grayline, line width=0.6pt},
        tick label style={font=\small, color=graytext},
        label style={font=\small},
        legend style={font=\scriptsize, draw=grayline, fill=white, fill opacity=0.85,
                      text opacity=1, inner sep=2pt, row sep=-1pt},
        xmin=0, xmax=1,
        xtick={0,0.2,0.4,0.6,0.8,1.0},
        xlabel={Co-visibility},
        ylabel={PSNR $\uparrow$},
    },
    covisline/.style={solid, mark=*, mark size=1.8pt, line width=1.3pt,
        mark options={draw=white, line width=0.4pt}},
}

\begin{subfigure}{0.49\textwidth}
\begin{tikzpicture}
  \begin{axis}[covisplot, ymin=10, ymax=25, ytick={10,14,18,22},
               legend pos=north west]
    \addplot[covisline, color=scigray, dashed, mark options={fill=scigray, draw=white}]
      coordinates {(0.05,10.8) (0.35,13.2) (0.55,15.6) (0.75,17.6) (0.95,22.4)};
    \addlegendentry{\mname, full frame}
    \addplot[covisline, color=sciblue, mark options={fill=sciblue, draw=white}]
      coordinates {(0.05,13.1) (0.35,14.6) (0.55,16.6) (0.75,18.4) (0.95,22.6)};
    \addlegendentry{\mname, co-visible}
    \addplot[covisline, color=scired, mark=square*, mark options={fill=scired, draw=white}]
      coordinates {(0.05,12.1) (0.35,13.8) (0.55,16.1) (0.75,18.0) (0.95,23.6)};
    \addlegendentry{Warp teacher, co-visible}
  \end{axis}
\end{tikzpicture}
\caption{\mname against its geometric warp teacher.}
\label{fig:supp_covis_warp}
\end{subfigure}\hfill
\begin{subfigure}{0.49\textwidth}
\begin{tikzpicture}
  \begin{axis}[covisplot, ymin=7, ymax=27, ytick={8,12,16,20,24},
               legend pos=north west]
    \addplot[covisline, color=sciblue, mark options={fill=sciblue, draw=white}]
      coordinates {(0.05,13.18) (0.15,15.42) (0.35,16.99) (0.55,19.07) (0.75,20.67) (0.95,25.65)};
    \addlegendentry{\mname-ft}
    \addplot[covisline, color=sciteal, mark=triangle*, mark size=2.2pt,
             mark options={fill=sciteal, draw=white}]
      coordinates {(0.05,11.42) (0.15,13.85) (0.35,14.52) (0.55,16.41) (0.75,17.92) (0.95,21.82)};
    \addlegendentry{\mname}
    \addplot[covisline, color=scired, mark=square*, mark options={fill=scired, draw=white}]
      coordinates {(0.05,12.98) (0.15,13.47) (0.35,15.18) (0.55,17.23) (0.75,18.98) (0.95,25.28)};
    \addlegendentry{VIVID}
    \addplot[covisline, color=sciorange, mark=diamond*, mark size=2.2pt,
             mark options={fill=sciorange, draw=white}]
      coordinates {(0.05,7.57) (0.15,11.68) (0.35,14.66) (0.55,16.55) (0.75,17.95) (0.95,21.49)};
    \addlegendentry{PhotoNVS}
    \addplot[covisline, color=scipurple, mark=pentagon*, mark size=2.2pt,
             mark options={fill=scipurple, draw=white}]
      coordinates {(0.05,11.88) (0.15,11.79) (0.35,12.57) (0.55,13.62) (0.75,14.48) (0.95,17.47)};
    \addlegendentry{GeoGPT}
  \end{axis}
\end{tikzpicture}
\caption{All methods at matched motion, full frame.}
\label{fig:supp_covis_matched}
\end{subfigure}

\caption{\textbf{Quality as a function of co-visibility} on RealEstate10K, where 1.0 denotes full
source-target overlap and lower values mean larger motion and more disocclusion.
\subref{fig:supp_covis_warp}~Restricting the measurement to co-visible pixels separates
reconstruction of observed content from hallucination of disoccluded content: \mname beats the
warp it is trained to imitate below 0.9 and the warp wins in the near-copy regime, which holds
most pairs, so pooled over all pairs the two essentially tie (19.5 against 19.7~dB).
\subref{fig:supp_covis_matched}~Binning by co-visibility compares methods at matched motion:
\mname-ft leads in every bin and widens its margin as motion grows, while PhotoNVS collapses at
the largest motions. Pooled over all pairs the methods reach 21.87 (\mname-ft), 20.51 (VIVID),
18.83 (\mname), 18.78 (PhotoNVS) and 15.25~dB (GeoGPT).}
\label{fig:supp_covisibility}
\end{figure}

\subsection{Sensitivity to the depth estimator}
\label{subsec:supp_depth}
Training from scratch on RealEstate10K under \mname's recipe, we change the depth estimator alone, replacing metric MoGe-2~\cite{moge2} with Depth Anything~3~\cite{depthanything3} (Table~\ref{tab:depth_backbone}). Depth Anything~3 predicts relative depth, supplies no validity mask, and has under half the parameters (135M against 331M), yet view quality moves by 0.28~dB on RealEstate10K and 0.42~dB on DL3DV, with SSIM and LPIPS within 0.02. The masked loss discards depth errors in disoccluded and invalid regions, and averaging over millions of images dilutes the rest. Errors on reflective surfaces, transparent materials, and thin structures remain a limitation.

\begin{table}[h]
\centering
\caption{\textbf{Depth-estimator ablation.} Models trained from scratch on RealEstate10K under an identical recipe, changing only the frozen depth network. Differences are relative to MoGe-2.}
\label{tab:depth_backbone}
\setlength{\tabcolsep}{4pt}
\fitwidth{%
\begin{tabular}{@{}l l ccc ccc@{}}
\toprule
& & \multicolumn{3}{c}{\cellcolor{lavblue}RealEstate10K~\cite{re10k}} & \multicolumn{3}{c}{\cellcolor{lavblue}DL3DV~\cite{dl3dv}} \\
\cmidrule(lr){3-5} \cmidrule(lr){6-8}
Depth estimator & Params & PSNR\,$\uparrow$ & SSIM\,$\uparrow$ & LPIPS\,$\downarrow$ & PSNR\,$\uparrow$ & SSIM\,$\uparrow$ & LPIPS\,$\downarrow$ \\
\midrule
\rowcolor{white}
MoGe-2~\cite{moge2} (metric, with mask) & 331M & \textbf{19.05} & \textbf{0.605} & \textbf{0.276} & \textbf{15.00} & \textbf{0.367} & \textbf{0.476} \\
\rowcolor{ltorange}
Depth Anything~3~\cite{depthanything3} (relative, no mask) & 135M & 18.77\dn{$-$0.28} & 0.592\dn{$-$.013} & 0.287\dn{+.011} & 14.58\dn{$-$0.42} & 0.353\dn{$-$.014} & 0.497\dn{+.021} \\
\bottomrule
\end{tabular}}
\end{table}

\subsection{Training-pose overlap}
\label{subsec:supp_overlap}
Lowering the training co-visibility places more of each example under the adversarial objective, since disoccluded pixels are covered by that term alone. We retrain \mname under the identical recipe, changing only the co-visibility distribution of the training-pose sampler: a high-overlap variant at median 0.59 and a low-overlap variant at 0.21, around the 0.43 of the released sampler (Table~\ref{tab:overlap_sampler}).

Aggregate differences are small because both benchmarks are dominated by high-overlap pairs. The trade-off appears once pairs are binned by co-visibility (Figure~\ref{fig:supp_overlap}): the low-overlap sampler leads below co-visibility 0.4 and is softer near copies, and the high-overlap sampler behaves the other way round. Low overlap also carries the worst FID, at 8.5 and 19.8, the realism cost of training on larger disocclusions. The choice affects training alone: at inference all three variants are single-pass and geometry-free.

\begin{table}[h]
\centering
\caption{\textbf{Training-pose overlap ablation.} Three samplers differing only in the co-visibility distribution of the sampled training poses, at ablation scale. The released-sampler row is the Mix run of Table~\ref{tab:data_coverage}. \textbf{Bold}: best; \underline{underline}: second best.}
\label{tab:overlap_sampler}
\setlength{\tabcolsep}{4pt}
\fitwidth{%
\begin{tabular}{@{}l cccc cccc@{}}
\toprule
& \multicolumn{4}{c}{\cellcolor{lavblue}RealEstate10K~\cite{re10k}} & \multicolumn{4}{c}{\cellcolor{lavblue}DL3DV~\cite{dl3dv}} \\
\cmidrule(lr){2-5} \cmidrule(lr){6-9}
Training sampler & PSNR\,$\uparrow$ & SSIM\,$\uparrow$ & LPIPS\,$\downarrow$ & FID\,$\downarrow$ & PSNR\,$\uparrow$ & SSIM\,$\uparrow$ & LPIPS\,$\downarrow$ & FID\,$\downarrow$ \\
\midrule
\rowcolor{white}
High overlap (median co-visibility 0.59) & \textbf{19.0} & \textbf{0.600} & \textbf{0.280} & \textbf{6.7} & \underline{14.9} & \underline{0.366} & \textbf{0.466} & \textbf{13.7} \\
\rowcolor{ltorange}
Released sampler (0.43)                  & \underline{18.8} & \underline{0.595} & \underline{0.284} & \underline{7.1} & \textbf{15.0} & \textbf{0.372} & \underline{0.467} & \underline{14.2} \\
\rowcolor{white}
Low overlap (0.21)                       & 18.4 & 0.572 & 0.310 & 8.5 & 14.8 & \underline{0.366} & 0.489 & 19.8 \\
\bottomrule
\end{tabular}}
\end{table}

\begin{figure}[!ht]
\centering
\begin{subfigure}{0.62\textwidth}
\begin{tikzpicture}
  \begin{axis}[
    width=\linewidth,
    height=0.58\linewidth,
    grid=major,
    grid style={grayline, line width=0.4pt, densely dotted, opacity=0.6},
    axis x line*=bottom,
    axis y line*=left,
    every outer x axis line/.append style={grayline, line width=0.6pt},
    every outer y axis line/.append style={grayline, line width=0.6pt},
    tick label style={font=\small, color=graytext},
    label style={font=\small},
    legend style={font=\scriptsize, draw=grayline, fill=white, fill opacity=0.85,
                  text opacity=1, inner sep=2pt, row sep=-1pt},
    legend pos=north west,
    xmin=0, xmax=1,
    xtick={0,0.2,0.4,0.6,0.8,1.0},
    xlabel={Co-visibility},
    ylabel={Co-visible PSNR $\uparrow$},
    ymin=11.5, ymax=25,
    ytick={12,15,18,21,24},
  ]
    \addplot[mark=square*, mark size=1.8pt, line width=1.3pt, color=scired, dashed,
             mark options={fill=scired, draw=white, line width=0.4pt}]
      coordinates {(0.05,12.1) (0.35,13.8) (0.55,16.1) (0.75,18.0) (0.95,23.6)};
    \addlegendentry{Warp teacher}
    \addplot[mark=*, mark size=1.8pt, line width=1.3pt, color=sciblue,
             mark options={fill=sciblue, draw=white, line width=0.4pt}]
      coordinates {(0.05,12.6) (0.35,14.2) (0.55,16.5) (0.75,18.3) (0.95,23.1)};
    \addlegendentry{High overlap (0.59)}
    \addplot[mark=triangle*, mark size=2.2pt, line width=1.3pt, color=scigreen,
             mark options={fill=scigreen, draw=white, line width=0.4pt}]
      coordinates {(0.05,13.1) (0.35,14.5) (0.55,16.2) (0.75,17.9) (0.95,22.1)};
    \addlegendentry{Low overlap (0.21)}
  \end{axis}
\end{tikzpicture}
\end{subfigure}

\caption{\textbf{Co-visible PSNR of the training-pose overlap variants} on RealEstate10K, binned
by co-visibility, against the geometric warp teacher. The low-overlap sampler is the most robust
at large viewpoint change and leads below co-visibility 0.4, while remaining softer near copies;
the high-overlap sampler behaves the other way round. The two curves cross, which is the
trade-off the released sampler sits between.}
\label{fig:supp_overlap}
\end{figure}

\subsection{Qualitative loss ablation}
\label{subsec:supp_lossablation}
Figure~\ref{fig:supp_loss_ablation} compares four loss configurations on three RealEstate10K test scenes, all trained for 250K steps under an identical recipe and differing only in their loss weights (rows 1, 4, 5 and 6 of Table~\ref{tab:loss_ablations}). Each row uses the benchmark's own relative pose, so the true target view is shown alongside. Removing the perceptual terms costs texture fidelity while the adversarial term keeps the output sharp; removing the adversarial term instead leaves streaks and soft patches along the disoccluded edge, which the reconstruction and perceptual losses never supervise; removing all three blurs the entire frame, the behaviour behind the FID of 62.1 and 111.0 in Table~\ref{tab:loss_ablations}.

\ifdefined\laImgW\else\newlength{\laImgW}\fi\setlength{\laImgW}{1.8cm}
\ifdefined\laGap\else\newlength{\laGap}\fi\setlength{\laGap}{1pt}
\ifdefined\laSep\else\newlength{\laSep}\fi\setlength{\laSep}{3pt}
\newcommand{\laImg}[1]{\includegraphics[width=\laImgW]{#1}}

\begin{figure}[t]
\centering
\resizebox{\linewidth}{!}{%
\setlength{\tabcolsep}{0pt}%
\begin{tabular}{
  @{}
  c @{\hspace{\laGap}}
  c
  @{\hspace{\laSep}}!{\color{black!25}\vrule width 0.4pt}@{\hspace{\laSep}}
  c @{\hspace{\laGap}}
  c @{\hspace{\laGap}}
  c @{\hspace{\laGap}}
  c
  @{}
}
\small Source &
\small Target &
\small \mname &
\small no perceptual &
\small no GAN &
\small recon.\ only \\[\laGap]
\laImg{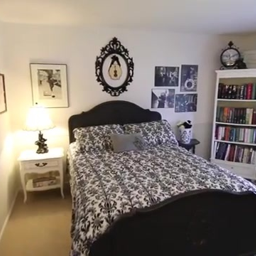} & \laImg{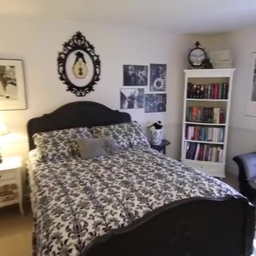} & \laImg{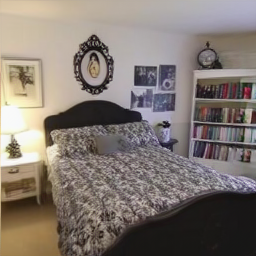} & \laImg{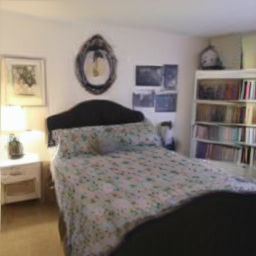} & \laImg{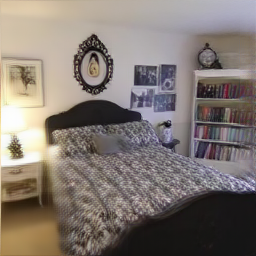} & \laImg{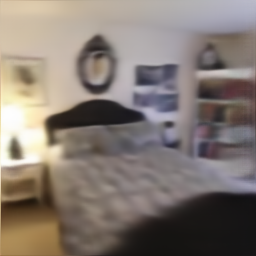} \\[\laGap]
\laImg{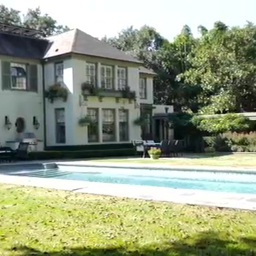} & \laImg{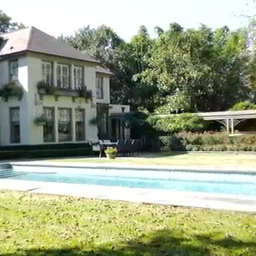} & \laImg{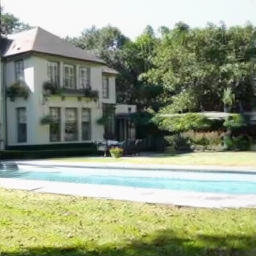} & \laImg{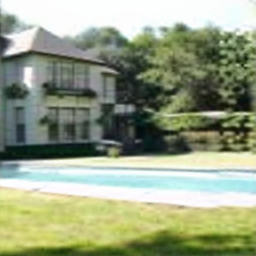} & \laImg{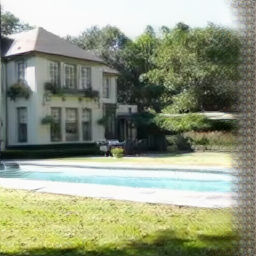} & \laImg{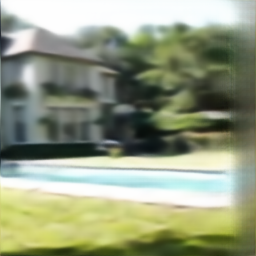} \\[\laGap]
\laImg{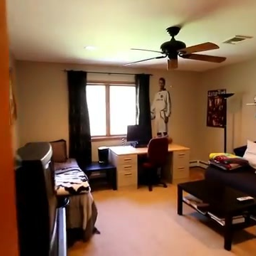} & \laImg{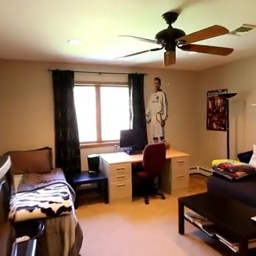} & \laImg{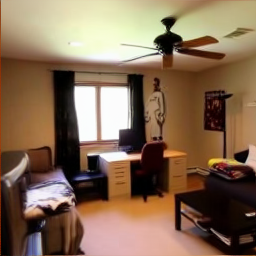} & \laImg{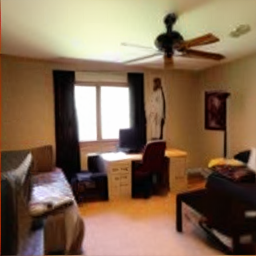} & \laImg{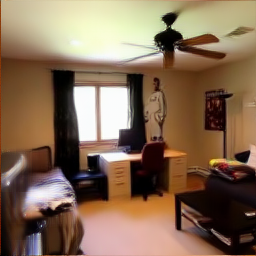} & \laImg{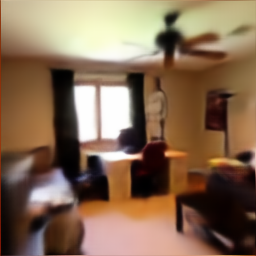} \\
\end{tabular}%
}
\caption{\textbf{Qualitative loss ablation on three RealEstate10K test scenes.} All four models
train for 250K steps under an identical recipe and differ only in their loss weights, matching
rows 1, 4, 5 and 6 of Table~\ref{tab:loss_ablations}. Each row uses one source frame and the
benchmark's own relative pose to the target frame 24 frames later, held fixed across the four
models, so \emph{Target} is the true novel view. Removing the perceptual terms costs texture
fidelity; removing the adversarial term leaves streaks and soft patches along the disoccluded
edge; removing all three blurs the whole frame.}
\label{fig:supp_loss_ablation}
\end{figure}

\subsection{Higher-resolution model: \mname-512}
\label{subsec:supp_resolution}
The architecture is resolution-agnostic by construction: the encoder and decoder are fully convolutional, and the ViT bottleneck operates on the downsampled latent with positional encodings regenerated for the new grid. We therefore also train \mname-512 at $512{\times}512$, on the same in-the-wild mix with MoGe-2 pseudo-pairs and under the identical recipe, for 250K steps at global batch size 256, changing only the resolution. We release both models.

Table~\ref{tab:ovie512} compares the two. At the like-for-like $256{\times}256$ evaluation, \mname-512 matches or improves on \mname on PSNR, SSIM, LPIPS, and MEt3R on both benchmarks, at a cost of roughly one FID point, which we attribute to the training mix: part of it sits below $512$ and is upscaled, carrying little true high-frequency detail. Inference remains a single feed-forward pass with no per-scene construction, at 41.6~ms per view after compilation on an H100 against 8.6~ms at $256$, tracking the four-fold growth in pixels and ViT tokens.

\begin{table}[h]
\centering
\caption{\textbf{\mname at $512{\times}512$}, zero-shot on both benchmarks.
\texttt{eval@256} is the like-for-like comparison, with every output brought to $256{\times}256$ before scoring; \texttt{eval@512} scores \mname-512 at its native resolution and is given for reference only.
\textbf{Bold}: better of the two models at \texttt{eval@256}.}
\label{tab:ovie512}
\setlength{\tabcolsep}{4pt}
\fitwidth{%
\begin{tabular}{@{}l ccccc ccccc@{}}
\toprule
& \multicolumn{5}{c}{\cellcolor{lavblue}RealEstate10K~\cite{re10k}} & \multicolumn{5}{c}{\cellcolor{lavblue}DL3DV~\cite{dl3dv}} \\
\cmidrule(lr){2-6} \cmidrule(lr){7-11}
Model & PSNR\,$\uparrow$ & SSIM\,$\uparrow$ & LPIPS\,$\downarrow$ & FID\,$\downarrow$ & MEt3R\,$\downarrow$
      & PSNR\,$\uparrow$ & SSIM\,$\uparrow$ & LPIPS\,$\downarrow$ & FID\,$\downarrow$ & MEt3R\,$\downarrow$ \\
\midrule
\rowcolor{white}
\mname, \texttt{eval@256}     & 18.8 & 0.602 & \textbf{0.279} & \textbf{6.74} & 0.035 & 14.8 & 0.369 & \textbf{0.464} & \textbf{13.6} & 0.078 \\
\rowcolor{ltorange}
\mname-512, \texttt{eval@256} & \textbf{19.1} & \textbf{0.611} & 0.283 & 7.62 & \textbf{0.034} & \textbf{15.2} & \textbf{0.389} & \textbf{0.464} & 14.8 & \textbf{0.077} \\
\midrule
\rowcolor{white}
\mname-512, \texttt{eval@512} & 18.9 & 0.658 & 0.344 & -- & -- & 15.1 & 0.475 & 0.509 & -- & -- \\
\bottomrule
\end{tabular}}
\end{table}

\Cref{fig:supp_res512_1,fig:supp_res512_2} show \mname-512 outputs at native resolution on held-out OpenImages~\cite{openimages} photographs that are already $512{\times}512$, under target poses drawn from the training sampler.

\ifdefined\resImgW\else\newlength{\resImgW}\fi\setlength{\resImgW}{0.325\linewidth}
\ifdefined\resGap\else\newlength{\resGap}\fi\setlength{\resGap}{2pt}

\begin{figure*}[t]
\centering
\setlength{\tabcolsep}{0pt}%
\setlength{\lineskip}{0pt}%
\begin{tabular}{
  @{}
  c @{\hspace{\resGap}}
  c @{\hspace{\resGap}}
  c
  @{}
}

\makebox[\resImgW][c]{\small Source} &
\makebox[\resImgW][c]{\small Pseudo-Target} &
\makebox[\resImgW][c]{\small Generated View} \\[4pt]

\includegraphics[width=\resImgW]{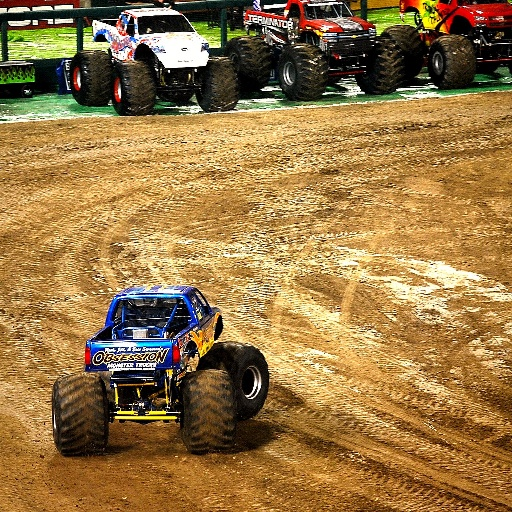} &
\includegraphics[width=\resImgW]{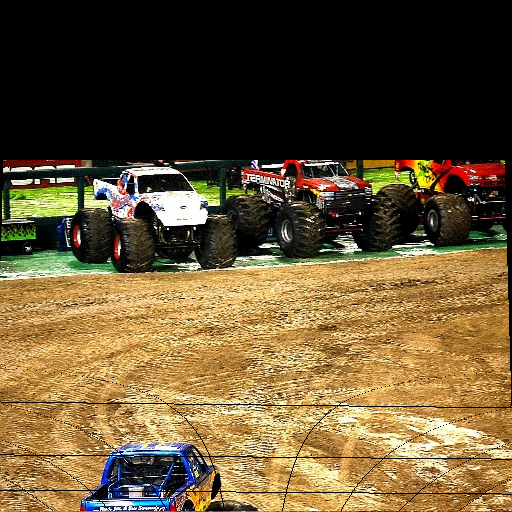} &
\includegraphics[width=\resImgW]{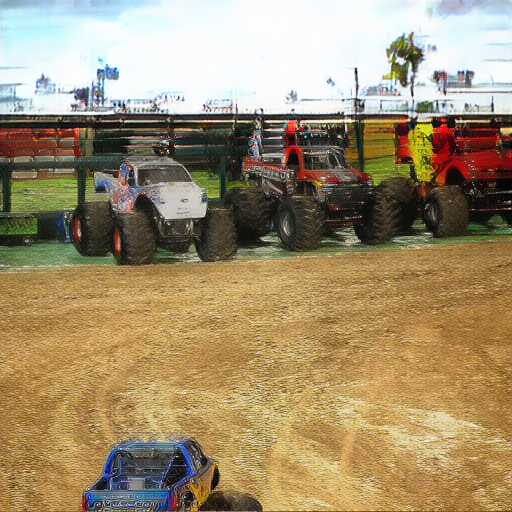} \\[\dimexpr\resGap-\dp\strutbox\relax]

\includegraphics[width=\resImgW]{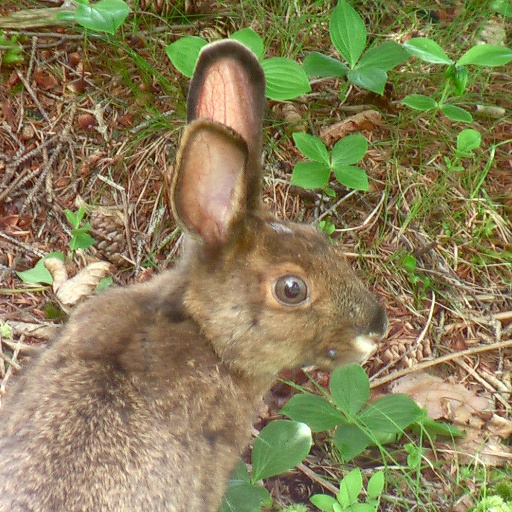} &
\includegraphics[width=\resImgW]{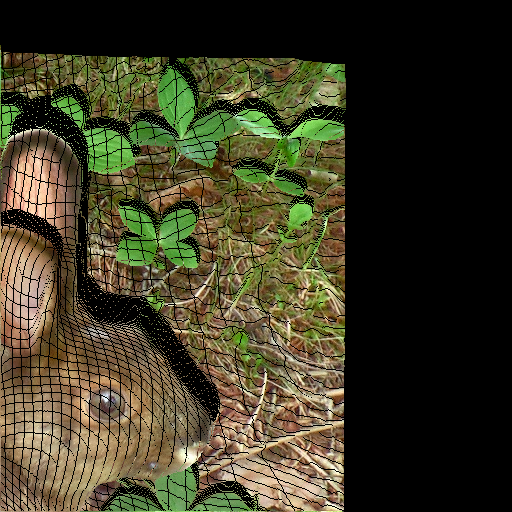} &
\includegraphics[width=\resImgW]{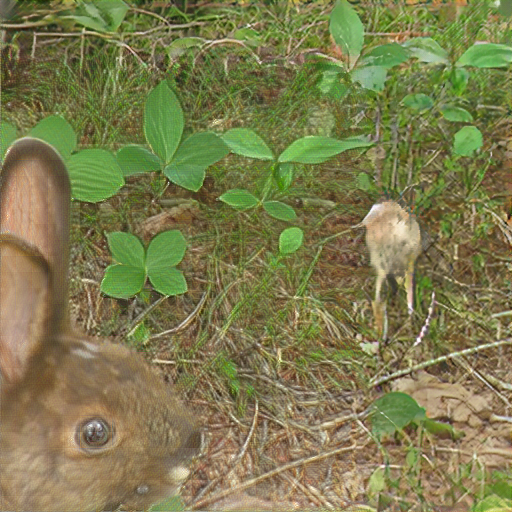} \\[\dimexpr\resGap-\dp\strutbox\relax]

\includegraphics[width=\resImgW]{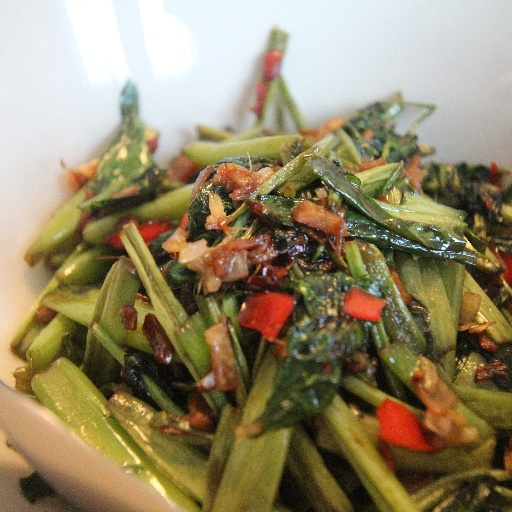} &
\includegraphics[width=\resImgW]{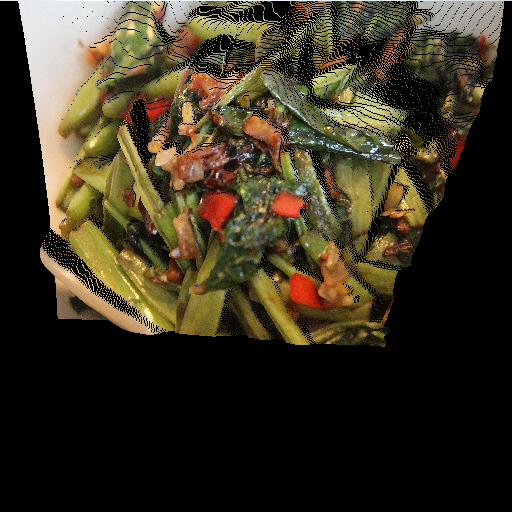} &
\includegraphics[width=\resImgW]{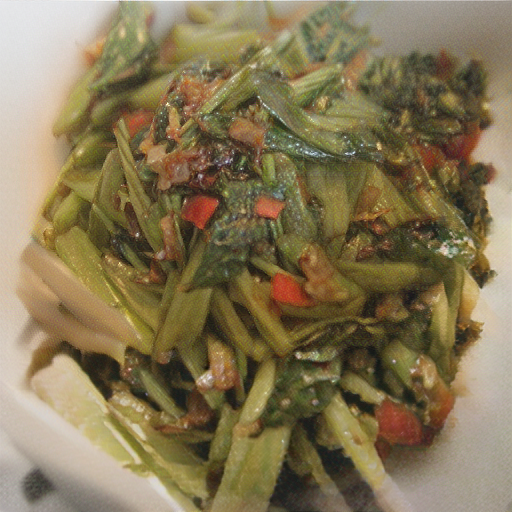}

\end{tabular}
\caption{\textbf{\mname-512 novel views at $512{\times}512$.} Source images are held-out
OpenImages~\cite{openimages} validation photographs at native $512{\times}512$, so no upscaling is
involved. Target poses are drawn from the training sampler of
\Cref{subsec:sampling_methods_hyperparameters}, and the pseudo-target column shows the MoGe-2
point cloud reprojected under that pose, exactly as during training. The generated views recover
detail the reprojection cannot provide, such as fur, foliage, and crowd texture, and fill the
regions the motion disoccludes. Co-visibility for these rows is 0.67, 0.47 and 0.43.}
\label{fig:supp_res512_1}
\end{figure*}

\ifdefined\resImgW\else\newlength{\resImgW}\fi\setlength{\resImgW}{0.325\linewidth}
\ifdefined\resGap\else\newlength{\resGap}\fi\setlength{\resGap}{2pt}

\begin{figure*}[t]
\centering
\setlength{\tabcolsep}{0pt}%
\setlength{\lineskip}{0pt}%
\begin{tabular}{
  @{}
  c @{\hspace{\resGap}}
  c @{\hspace{\resGap}}
  c
  @{}
}

\makebox[\resImgW][c]{\small Source} &
\makebox[\resImgW][c]{\small Pseudo-Target} &
\makebox[\resImgW][c]{\small Generated View} \\[4pt]

\includegraphics[width=\resImgW]{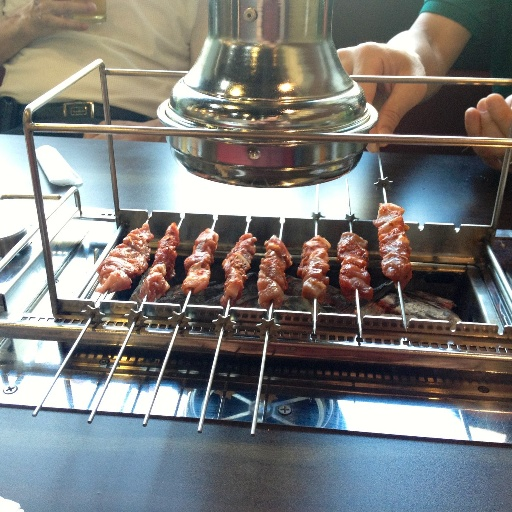} &
\includegraphics[width=\resImgW]{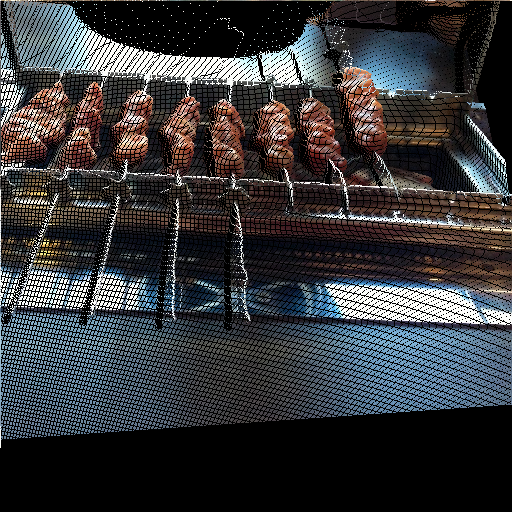} &
\includegraphics[width=\resImgW]{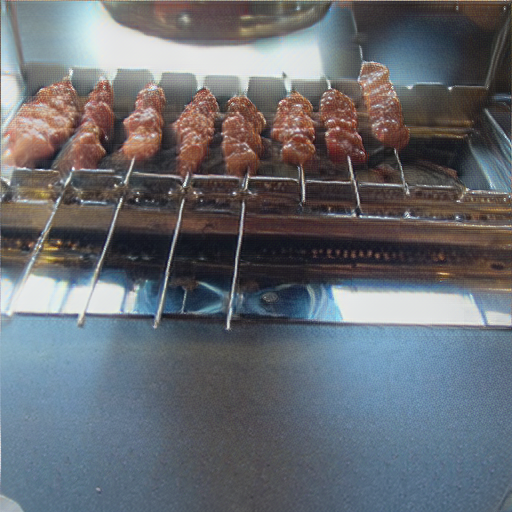} \\[\dimexpr\resGap-\dp\strutbox\relax]

\includegraphics[width=\resImgW]{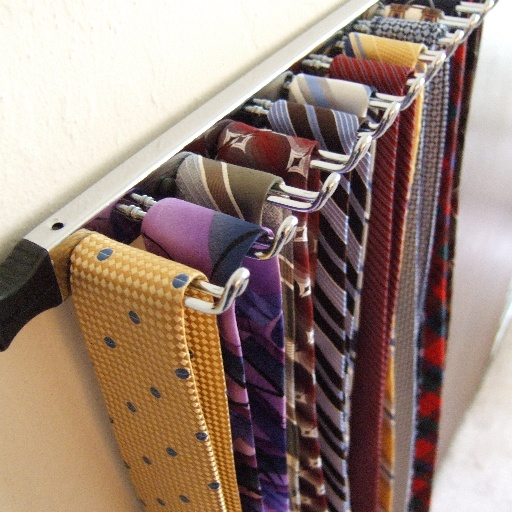} &
\includegraphics[width=\resImgW]{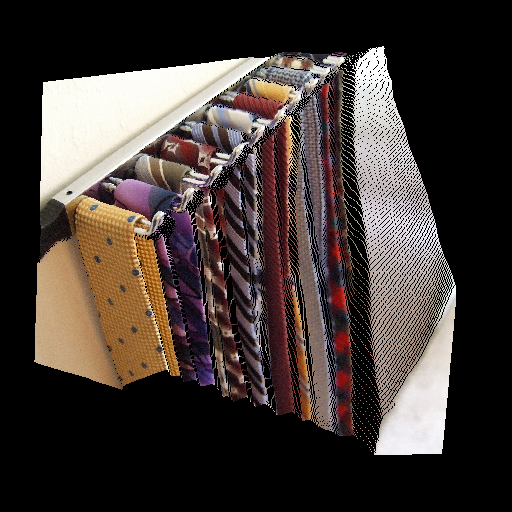} &
\includegraphics[width=\resImgW]{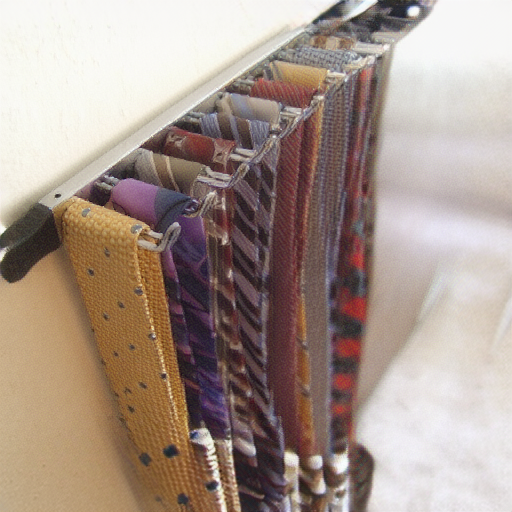} \\[\dimexpr\resGap-\dp\strutbox\relax]

\includegraphics[width=\resImgW]{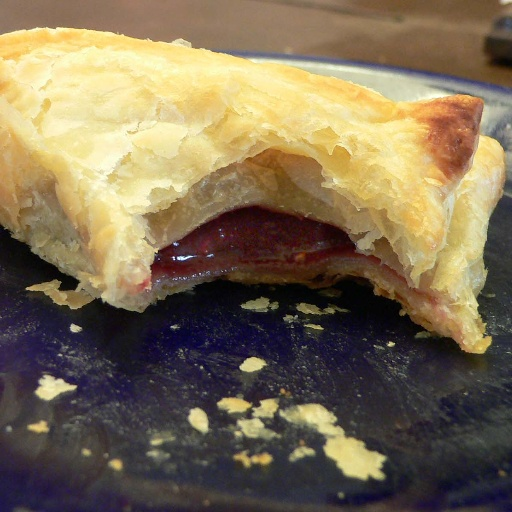} &
\includegraphics[width=\resImgW]{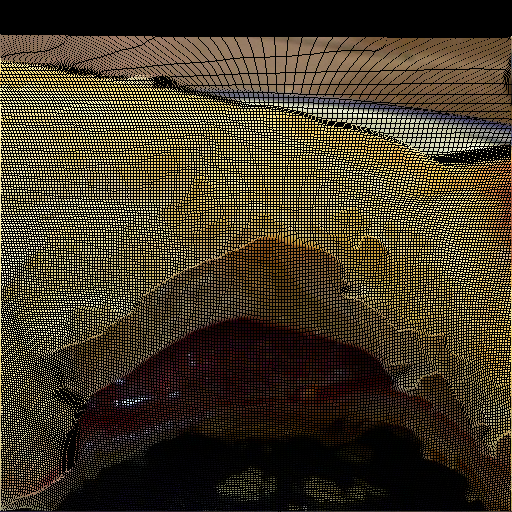} &
\includegraphics[width=\resImgW]{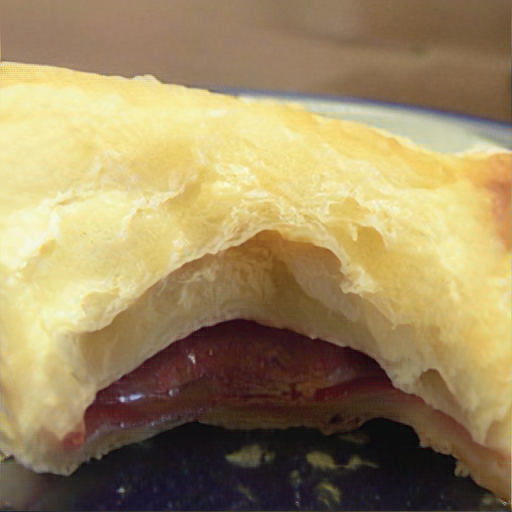}

\end{tabular}
\caption{\textbf{\mname-512 novel views at $512{\times}512$, continued.} Same protocol as
\Cref{fig:supp_res512_1}. These rows show specular metal, a dense repeated structure, and layered
flaky texture. Co-visibility is 0.52, 0.43 and 0.42.}
\label{fig:supp_res512_2}
\end{figure*}

\subsection{Comparison to single-image 3D reconstruction and scene generation}
\label{subsec:supp_explicit3d}
A neighbouring family maps a single image to an explicit 3D representation and renders target views from it. We evaluate the three members of that family that run under our protocol, 14 targets at stride 3 with the per-scene scale sweep, grouped by regime (Table~\ref{tab:explicit3d}).

Zero-shot, base \mname beats SHARP on every metric and matches FlexWorld on PSNR and SSIM while leading it on LPIPS and FID; FlexWorld's pixel-metric edge comes from softer per-scene-optimized renders, which blur rewards on PSNR and SSIM and which LPIPS and FID penalize. In-domain, \mname-ft leads the public DiffusionGS re-implementation on PSNR, LPIPS, and FID, with DiffusionGS ahead on SSIM. Cost in this family is dominated by scene construction, which \mname does not perform; FlexWorld runs 11.3B parameters at inference across a video diffusion model, a text encoder, and DUSt3R, against \mname's 143M.

\begin{table}[h]
\centering
\caption{\textbf{Comparison to single-image methods that build an explicit 3D representation}, on RealEstate10K under our protocol.
Time to first view includes any per-scene 3D construction; subsequent views are amortized for the explicit-3D methods and cost one forward pass for \mname.
The DiffusionGS row uses the public re-implementation, since the original model is unreleased and its authors report that it differs from the paper.
\textbf{Bold}: best per regime.}
\label{tab:explicit3d}
\setlength{\tabcolsep}{5pt}
\fitwidth{%
\begin{tabular}{@{}l l cccc l@{}}
\toprule
Method & Params & PSNR\,$\uparrow$ & SSIM\,$\uparrow$ & LPIPS\,$\downarrow$ & FID\,$\downarrow$ & Time to first view \\
\midrule
\multicolumn{7}{l}{\textit{Zero-shot on RealEstate10K}} \\
\rowcolor{white}
\quad SHARP~\cite{sharp}         & 702M   & 18.51 & 0.589 & 0.299 & 10.18 & ${\sim}1$\,s \\
\rowcolor{ltorange}
\quad FlexWorld~\cite{flexworld} & 11.3B & \textbf{18.96} & \textbf{0.625} & 0.290 & 19.32 & ${\sim}225$\,s \\
\rowcolor{white}
\quad \mname (ours)              & 143M   & 18.8 & 0.602 & \textbf{0.279} & \textbf{6.74} & $<10$\,ms \\
\midrule
\multicolumn{7}{l}{\textit{In-domain on RealEstate10K}} \\
\rowcolor{ltorange}
\quad DiffusionGS~\cite{diffusiongs} & --  & 21.34 & \textbf{0.731} & 0.217 & 8.99 & ${\sim}0.8$\,s \\
\rowcolor{white}
\quad \mname-ft (ours)           & 143M   & \textbf{21.9} & 0.695 & \textbf{0.195} & \textbf{5.59} & $<10$\,ms \\
\bottomrule
\end{tabular}}
\end{table}

\subsection{Evaluation under VIVID's protocol}
\label{subsec:supp_vividprotocol}
Table~\ref{tab:comparison_with_sota} re-evaluates every method under one protocol, since frame stride, pair selection, metric-scale handling, and crop conventions vary from paper to paper. As a cross-check, we also evaluate \mname-ft under the protocol of VIVID~\cite{vivid}, our closest baseline: one source-target pair per scene, the target 30 to 120 frames from the source, the ground-truth structure-from-motion pose used directly, at $256{\times}256$ (Table~\ref{tab:vivid_protocol}).

\mname-ft matches VIVID's published numbers at both ranges, within roughly 0.1~dB, while generating each view in a single geometry-free forward pass against VIVID's 32-step pixel-space diffusion sampler, roughly 30 times fewer network evaluations per view. VIVID trains with scale preprocessing matched to this evaluation; \mname does not. The single-pair, large-gap protocol is a larger-motion regime than our stride-3 setup, which is why VIVID reads 20.5 in Table~\ref{tab:comparison_with_sota} and 17.36 in its own paper.

\begin{table}[h]
\centering
\caption{\textbf{RealEstate10K under VIVID's original protocol} (PSNR, higher is better). Source-copy returns the source image unchanged and bounds the difficulty of each range.}
\label{tab:vivid_protocol}
\setlength{\tabcolsep}{8pt}
\begin{tabular}{@{}l ccc@{}}
\toprule
Range & \mname-ft (ours) & VIVID~\cite{vivid} (published) & Source-copy \\
\midrule
\rowcolor{white}
Mid-range (gap 30--60)   & 17.24 & \textbf{17.36} & 13.12 \\
\rowcolor{ltorange}
Long-range (gap 60--120) & \textbf{15.30} & 15.21 & 11.91 \\
\bottomrule
\end{tabular}
\end{table}

\section{Camera Sampling Details}
\label{sec:appendix_sampling}

During training, we generate novel views of the input image to serve as pseudo-ground-truth targets by reprojecting the scene's 3D point cloud into newly sampled camera viewpoints. To achieve this, a monocular depth estimator first predicts both the absolute depth and the camera's horizontal field of view ($\Theta_\text{h}$), together enabling reconstruction of the point cloud in a true metric, real-world scale. Because this geometry possesses accurate physical dimensions, the distributions used to sample new camera poses can be defined directly in actual scene units (\textit{e.g.}, meters) rather than an arbitrary coordinate space. This ensures that camera displacements and distances scale consistently with the specific geometry of each scene. Next, a routing module stochastically assigns each batch element to one of six sampling methods based on fixed prior weights. All methods ultimately produce a world-to-camera extrinsic matrix $[\mathbf{R} \mid \mathbf{t}] \in \mathbb{R}^{3 \times 4}$.

Throughout this section, we denote continuous uniform and normal distributions as $\mathcal{U}$ and $\mathcal{N}$, respectively, and let $\hat{y} = [0, 1, 0]^\top$ and $\hat{z} = [0, 0, 1]^\top$ represent the canonical up and forward directional unit vectors.

\subsection{Sampling methods.}
\label{subsec:detailed_sampling_methods}
The six strategies span a range of transformation types, from trivial (identity) to geometry-grounded (normal-derived, frontal hemisphere), ensuring the model is trained on diverse yet plausible viewpoint changes.

\begin{itemize}

\item \textbf{Identity.} No transformation is applied. The extrinsic is set to $[\mathbf{I} \mid \mathbf{0}]$.

\item \textbf{Pure translation.} The camera is shifted relative to the scene without any rotation, with the shift magnitude tied to the spatial extent of the point cloud. The rotation is fixed to $\mathbf{R} = \mathbf{I}$. The translation is sampled as $\mathbf{t} \sim \mathcal{U}[-\alpha_\text{t} \boldsymbol{\sigma}, +\alpha_\text{t} \boldsymbol{\sigma}]$, where $\alpha_\text{t} \in \mathbb{R}^+$ is a scaling hyperparameter and $\boldsymbol{\sigma} \in \mathbb{R}^3$ is the per-axis standard deviation of the point cloud. To prevent points from passing behind the camera, the z-component of the translation is clamped to $t_z \leq \min_i z_i$, where $z_i$ is the depth (z-coordinate) of the $i$-th point.

\item \textbf{Pure rotation.} The camera rotates in place, with the maximum rotation angle bounded by the field of view. The translation is fixed to $\mathbf{t} = \mathbf{0}$. A forward direction is sampled by rotating the canonical forward axis $\hat{z}$ by polar angle $\theta \sim \mathcal{U}[0,\, \alpha_\text{r} \Theta_\text{h}]$ (where $\alpha_\text{r} \in \mathbb{R}^+$ is a rotation scaling factor and $\Theta_\text{h}$ is the horizontal field of view estimated by the monocular depth estimator) and azimuth $\phi \sim \mathcal{U}[0, 2\pi)$. This direction is then orthonormalized against the canonical up vector $\hat{y}$ to form $\mathbf{R}$.

\item \textbf{Combined rotation and translation.} The camera is both shifted and rotated, combining the two previous strategies. A translation $\mathbf{t}_\text{t}$ and rotation $\mathbf{R}_\text{r}$ are sampled independently as above and composed as $\mathbf{R}_{\text{hybrid}} = \mathbf{R}_\text{r}$, $\mathbf{t}_{\text{hybrid}} = \mathbf{R}_\text{r} \mathbf{t}_\text{t}$.

\item \textbf{Normal-derived.} The camera is placed above a randomly selected surface point, looking at it from along its normal direction, as estimated by the monocular depth estimator. An anchor point $\mathbf{p} \in \mathbb{R}^3$ is sampled with probability $\propto \|\mathbf{p}\|^{-1}$, restricted to points whose surface normal $\hat{\mathbf{n}} \in \mathbb{R}^3$ satisfies $|n_y| < \tau$, where $n_y$ is the y-component of the normal and $\tau$ is a filtering threshold. The camera is placed at $\mathbf{c} = \mathbf{p} + s\, \hat{\mathbf{n}}$, where the distance multiplier $s$ is drawn from a log-uniform distribution $s \sim \log\mathcal{U}(d_{\min} \|\mathbf{p}\|,\, d_{\max} \|\mathbf{p}\|)$, with $d_{\min}$ and $d_{\max}$ representing the minimum and maximum distance bounds. Log-uniform sampling is used here to ensure that exponentially large distances are not overrepresented. Finally, $\mathbf{R}$ is set by a look-at from $\mathbf{c}$ to $\mathbf{p}$. Elements for which no valid normal survives filtering fall back to identity.

\item \textbf{Frontal hemisphere.} The camera orbits around a randomly selected scene point, staying roughly frontal with a limited angular deviation. An anchor $\mathbf{p}$ is sampled with probability $\propto \|\mathbf{p}\|^{-1}$ and jittered as $\tilde{\mathbf{p}} = \mathbf{p} + \boldsymbol{\epsilon}$, $\boldsymbol{\epsilon} \sim \mathcal{N}(\mathbf{0},\, (\sigma_{\text{anchor}} \|\mathbf{p}\|)^2 \mathbf{I})$, where $\sigma_{\text{anchor}}$ is a hyperparameter controlling the variance of the jitter. The reference direction $\hat{\mathbf{r}} = -\tilde{\mathbf{p}} / \|\tilde{\mathbf{p}}\|$ is perturbed by azimuth and elevation each drawn from $\mathcal{U}[-\delta, \delta]$, where $\delta$ bounds the maximum angular deviation, to obtain a new viewing direction $\hat{\mathbf{d}}$. The camera is placed at $\mathbf{c} = \tilde{\mathbf{p}} + z \hat{\mathbf{d}}$, where $z = \|\mathbf{p}\| \cdot s$ and $s \sim \log\text{-}\mathcal{U}(d_{\min}, d_{\max})$. As previously mentioned, the log sampling of the distance multiplier $s$ ensures that large distances are not overrepresented. Finally, $\mathbf{R}$ is set by a look-at from $\mathbf{c}$ to $\tilde{\mathbf{p}}$.

\end{itemize}

\subsection{Look-at construction.} Given camera 
\label{subsec:lookat}position $\mathbf{c}$ and target $\mathbf{p}$, we compute the forward vector $\hat{f} = (\mathbf{p} - \mathbf{c})/\|\mathbf{p} - \mathbf{c}\|$, the right vector $\hat{r} = (\hat{y} \times \hat{f})/\|\hat{y} \times \hat{f}\|$, the true up vector $\hat{u} = \hat{f} \times \hat{r}$, and set $\mathbf{R} = [\hat{r} \mid \hat{u} \mid \hat{f}]$.

\subsection{Geometry-aware reprojection.}
\label{subsec:reprojection}
Given a target viewpoint, source colors are reprojected by mapping each 3D point into the new camera's image plane. Formally, each source point $\mathbf{p}_i \in \mathbb{R}^3$, derived from the monocular depth estimator, with normal $\hat{\mathbf{n}}_i$ is projected to 2D pixel coordinates $\mathbf{q}_i$. This is expressed in homogeneous coordinates as $\mathbf{q}_i \sim \mathbf{K}\,(\mathbf{R}\mathbf{p}_i + \mathbf{t})$, where $\mathbf{K} \in \mathbb{R}^{3 \times 3}$ is the known camera intrinsic matrix.

To handle occlusions, we apply a strategy akin to backface culling in computer graphics, discarding points that face away from the camera, \textit{i.e.}, those satisfying $\hat{\mathbf{n}}_i^\top (\mathbf{c} - \mathbf{p}_i) \leq 0$, where $\mathbf{c} = -\mathbf{R}^\top \mathbf{t}$ is the target camera center in world coordinates. When multiple valid points project onto the exact same discrete pixel, a z-buffer resolves the collision by assigning the pixel the color of the point with the minimum projected depth. If no points project onto a given pixel, it remains black.

Finally, a visibility mask is computed to indicate these valid, populated pixels; this mask is later applied during the computation of perceptual losses as discussed in the main paper.
\subsection{Hyperparameters.} 
\label{subsec:sampling_methods_hyperparameters}
All sampling hyperparameters and model settings are summarized in \Cref{tab:sampling_hyperparams}. 

\begin{table}[h]
\centering
\caption{Camera sampling hyperparameters and model settings used during training.}
\label{tab:sampling_hyperparams}
\begin{tabular}{lll}
\toprule
\textbf{Parameter} & \textbf{Symbol} & \textbf{Value} \\
\midrule
\multicolumn{3}{l}{\textit{Sampling probabilities}} \\
\quad Identity                             & --                     & 0.15 \\
\quad Pure translation                     & --                     & 0.10 \\
\quad Pure rotation                        & --                     & 0.10 \\
\quad Combined rotation \& translation     & --                     & 0.35 \\
\quad Normal-derived                       & --                     & 0.05 \\
\quad Frontal hemisphere                   & --                     & 0.25 \\
\midrule
\multicolumn{3}{l}{\textit{Translation \& rotation}} \\
\quad Translation scaling factor           & $\alpha_\text{t}$             & 1.0  \\
\quad Rotation scaling factor              & $\alpha_\text{r}$             & 1.0  \\
\midrule
\multicolumn{3}{l}{\textit{Normal-derived \& frontal hemisphere}} \\
\quad Distance range                       & $[d_{\min}, d_{\max}]$ & $[0.75,\ 1.5]$ \\
\quad Max perturbation angle               & $\delta$               & $25^\circ$ \\
\quad Anchor jitter scale                  & $\sigma_{\text{anchor}}$ & 0.02 \\
\midrule
\multicolumn{3}{l}{\textit{Model settings}} \\
\quad Depth estimator                      & --                     & moge-2-vitl-normal~\cite{moge2} \\
\bottomrule
\end{tabular}
\end{table}

\section{Implementation Details}
\label{sec:appendix_implementation}

\subsection{Architecture}
\label{subsec:supp_arch}
OVIE consists of a convolutional encoder ($8\times$ spatial downsampling), a ViT-B bottleneck, and a symmetric convolutional decoder. Input images ($256{\times}256$) are compressed to a $32{\times}32{\times}512$ feature map via three ResNet stages (GroupNorm, SiLU), then patchified into 1024 tokens ($1{\times}1$ patches) and linearly projected to $D{=}768$ dimensions. A 12-layer ViT-B ($D{=}768$, 12 heads, RMSNorm, SwiGLU) processes the tokens, after which they are unpatchified and decoded symmetrically, concluding with a $1{\times}1$ convolution and Sigmoid activation.

\subsection{Camera Conditioning via AdaLN}
\label{subsec:supp_adaln}

Camera extrinsics $\mathbf{p} \in \mathbb{R}^7$ (translation and a quaternion for rotation) are projected to a conditioning embedding $\mathbf{c} \in \mathbb{R}^D$ via a single linear layer. This embedding modulates the ViT-B bottleneck via adaLN-Zero~\cite{dit}: for each transformer block, a two-layer MLP regresses, from $\mathbf{c}$, dimension-wise scale $\gamma$, shift $\beta$, and residual gate $\alpha$ for both the MSA and SwiGLU sub-layers:
\begin{align}
\mathbf{x}' &= \mathbf{x} + \alpha_{\text{msa}} \odot \,\text{MSA}\!\bigl(\gamma_{\text{msa}} \odot \text{RMSNorm}(\mathbf{x}) + \beta_{\text{msa}}\bigr), \\
\mathbf{x}'' &= \mathbf{x}' + \alpha_{\text{mlp}} \odot \,\text{SwiGLU}\!\bigl(\gamma_{\text{mlp}} \odot \text{RMSNorm}(\mathbf{x}') + \beta_{\text{mlp}}\bigr).
\end{align}
Following adaLN-Zero, the MLP's final linear layer is zero-initialized so the conditioning path contributes nothing at the start of training.

\subsection{Optimization and Training}
OVIE is trained for 2{,}000{,}000 steps with a global batch size of 512 using the AdamW optimizer ($\beta_1=0.9$, $\beta_2=0.999$, weight decay of 0.05). We apply gradient clipping with a maximum norm of 1.0. The learning rate follows a cosine decay schedule, annealing from a peak of $2 \times 10^{-4}$ down to a minimum of $2 \times 10^{-5}$, which is preceded by a linear warmup phase over the first $0.625\%$ of training ($\sim$12.5k steps). Finally, we maintain an exponential moving average (EMA) of the generator weights with a decay rate of 0.999 for inference.

\paragraph{Loss.}
The training objective combines L2 reconstruction loss, LPIPS ($\lambda_{\text{LPIPS}}{=}1.0$), and a P-DINO perceptual loss ($\lambda_\text{P-DINO}{=}0.5$) extracted from a pretrained DINOv3-ViT-B/16 model \cite{dinov3}.

\paragraph{Adversarial Training.}
A PatchGAN discriminator~\cite{patchgan} sharpens high-frequency detail. Following Representation Autoencoders (RAE)~\cite{rae}, we adopt a frozen DINO-S/8 backbone (inputs resized to $224{\times}224$) over the standard DINO-S/16, which reduces adversarial patch artifacts. The discriminator uses standard batch normalization, a convolutional head (kernel size 9) with Spectral Normalization, a hinge loss for discriminator updates, and a non-saturating loss for the generator. To balance the scale of the reconstruction and adversarial gradients, we employ a dynamic adaptive weighting scheme originally introduced in VQGAN~\cite{vqgan}. At each training step, the adaptive weight $\lambda$ is computed as:
$$ \lambda = \text{clamp}\left( \frac{\|\nabla_{\mathbf{W}_L} \mathcal{L}_{\text{recon}}\|_2}{\|\nabla_{\mathbf{W}_L} \mathcal{L}_{\text{GAN}}\|_2 + \epsilon},\, 0,\, \lambda_{\text{max}} \right), $$
where $\nabla_{\mathbf{W}_L}$ denotes the gradient with respect to the weights of the last convolutional layer of the decoder, $\mathcal{L}_{\text{recon}}$ is the L2 reconstruction loss, $\mathcal{L}_{\text{GAN}}$ is the generator's adversarial loss, and $\epsilon = 10^{-6}$ ensures numerical stability. The weight is clamped to a maximum limit of $\lambda_{\text{max}} = 10{,}000$ to prevent gradient explosion. The final adversarial penalty added to the total training objective is scaled by $\lambda_{\text{adv}} \lambda \mathcal{L}_{\text{GAN}}$, where $\lambda_{\text{adv}} = 0.75$ is a fixed scalar. Finally, discriminator updates and adversarial penalties are delayed until 37.5\% and 40\% of total steps, respectively, to prevent early collapse.

\paragraph{Hyperparameters.}
\label{subsec:supp_hyperparameters}
All hyperparameters are reported in Table~\ref{tab:full_hyperparams}. For all ablation studies and supplementary experiments, we use the identical hyperparameter configuration but reduce the total training duration to 250{,}000 steps, keeping the absolute number of warmup steps constant.

\begin{table}[h]
\centering
\caption{Architecture, optimization, and loss hyperparameters.}
\label{tab:full_hyperparams}
\renewcommand{\arraystretch}{0.95} %
\fitwidth{%
\begin{tabular}{ll@{\hspace{3em}}ll}
\toprule
\textbf{Hyperparameter} & \textbf{Value} & \textbf{Hyperparameter} & \textbf{Value} \\
\midrule
\multicolumn{2}{l}{\textit{Generator Architecture}} & \multicolumn{2}{l}{\textit{Losses}} \\
Resolution                        & $256 \times 256$ & Reconstruction loss               & L2 (MSE) \\
Base channels                     & 128 & LPIPS weight $\lambda_\text{LPIPS}$       & 1.0 \\
Channel multipliers               & $[1,\,2,\,4]$ & P-DINO model                      & \texttt{DINOv3-ViT-B/16} \\
Downsampling factor               & $8\times$ & P-DINO weight $\lambda_\text{DINO}$       & 0.5 \\
ViT bottleneck                    & & Adversarial weight $\lambda_\text{adv}$   & 0.75 \\
\quad \# Layers                   & 12 & Max adaptive weight limit         & 10{,}000 \\
\quad \# Heads           & 12 & & \\
\quad Hidden dimension                     & 768 & \multicolumn{2}{l}{\textit{Discriminator \& Adversarial Training}} \\
Normalization                     & RMSNorm & Backbone                          & Frozen DINO-S/8, $224{\times}224$ \\
Activation                        & SwiGLU & Head                              & Conv $9{\times}9$, BN, SN \\
Camera embedder                   & Linear$(\mathbb{R}^7 \to \mathbb{R}^D)$ & Augmentation probability          & 1.0 \\
\multicolumn{2}{l}{\textit{Generator Optimisation}} & Discriminator loss                & Hinge \\
Batch size                        & 512 & Generator loss                    & Non-saturating (Vanilla) \\
Training steps                    & 2{,}000{,}000 & Optimizer                         & AdamW \\
Optimizer                         & AdamW & Optimizer betas $(\beta_1, \beta_2)$ & $(0.9,\, 0.95)$ \\
Optimizer betas $(\beta_1, \beta_2)$ & $(0.9,\, 0.999)$ & Peak learning rate                & $2{\times}10^{-4}$ \\
Peak learning rate                & $2{\times}10^{-4}$ & Minimum learning rate             & $2{\times}10^{-5}$ \\
Minimum learning rate             & $2{\times}10^{-5}$ & Learning rate scheduler           & Cosine with warmup \\
Learning rate scheduler           & Cosine with warmup & Warmup ratio                      & $5\%$ \\
Warmup ratio                      & $0.625\%$ (${\sim}12.5$k steps) & Weight decay                      & 0.0 \\
Weight decay                      & 0.05 & $D_\phi$ update start             & $37.5\%$ (${\sim}750$k steps) \\
Gradient clip (max norm)          & 1.0 & $\mathcal{L}_\text{adv}$ start    & $40.0\%$ (${\sim}800$k steps) \\
EMA decay                         & 0.999 & & \\
\bottomrule
\end{tabular}}
\end{table}

\section{Training Resources and Broader Impact}
\label{sec:appendix_resources}

\paragraph{Training resources.}
All models are trained on 32 H100 GPUs. Ablation models complete in approximately 29 hours (250K steps); the full model completes in approximately 9.5 days (2M steps).

\paragraph{Broader impact.}
Training on unpaired single images opens novel view synthesis to domains where multi-view capture is impossible, including cultural heritage and archival photography. The ability to synthesize views from a single photograph also carries misuse risk for disinformation.

On the environmental side, the one-time training cost (9.5 days on 32 H100 GPUs) is comparable to similar-scale generative models. At inference, however, \mname operates at 116~FPS in a single forward pass, more than 600$\times$ faster than the fastest baseline. This makes its per-image carbon footprint orders of magnitude smaller than diffusion-based competitors, whose iterative sampling (50--2000 steps) dominates inference cost at deployment scale.

\section{Additional Qualitative Results}
\label{sec:appendix_qualitative}

\subsection{Out-of-distribution novel views}
\label{subsec:ood}

Figure \ref{fig:supp_ood} illustrates novel views synthesized from out-of-domain, non-realistic source images (\eg, paintings). Notably, training on such artistic domains would be unfeasible using standard monocular novel-view synthesis methods reliant on multi-view datasets.

\subsection{Comparison between training pseudo-targets and generated views}
\label{subsec:pseudotargets}
Figures \ref{fig:source_pc_gen_1}--\ref{fig:source_pc_gen_4} show examples of pseudo-targets used for supervision during training, along with the views generated from the source image and the target camera pose. Fine detail in observed regions survives the viewpoint change: the last row of Figure~\ref{fig:source_pc_gen_4} contains shop signage whose text stays legible in the generated view, even though the pseudo-target covers only part of the frame.

The grid-like patterns in the pseudo-target images (middle) stem from the point cloud's regular spatial structure. Because each 3D point is a back-projected source pixel, the points inherit the original image's grid layout. When rendered from a novel viewpoint, the spacing between these points becomes visible as a grid that varies with depth and angle.

\subsection{Comparison to baseline methods}
\label{subsec:baselines}
Figure \ref{fig:sup_comparison_1}--\ref{fig:sup_comparison_3} show qualitative comparisons between GeoGPT~\cite{geogpt}, PhotoNVS~\cite{photonvs}, VIVID~\cite{vivid}, and OVIE on the RealEstate10K~\cite{re10k} dataset.

\subsection{Side-by-side navigation clips}
\label{subsec:sidebyside}
The project page, \url{https://kyutai.org/blog/2026-04-14-ovie/}, hosts animations demonstrating OVIE on continuous trajectories. These animations use sequences from the RealEstate10K~\cite{re10k} dataset. Each generated sequence is synthesized by conditioning solely on the first frame, without using any subsequent ground-truth frames. Unlike diffusion or autoregressive models, OVIE is a deterministic mapping: fixed inputs always produce the same output, with no stochastic sampling. Even in disoccluded regions, the source image serves as a fixed conditioning anchor: rather than drawing from stochastic noise as diffusion models do, OVIE maps the same input to the same hallucinated content at each forward pass. Temporal coherence over a trajectory then follows from the neural network's smooth response to continuously varying pose inputs: nearby viewpoints produce nearby outputs, including in unobserved regions.

\subsection{Real-time interactive navigation clips}
\label{subsec:realtime}
The project page also hosts screen recordings of OVIE's real-time interactive navigation. To achieve this, we map standard mouse and keyboard actions to small, incremental changes in the camera's position and rotation, similar to the control mechanics found in first-person video games. Based on these inputs, we continuously update the camera extrinsics to generate a new image on the fly, conditioned strictly on the initial source image.

\ifdefined\qualImgW\else\newlength{\qualImgW}\fi\setlength{\qualImgW}{3.5cm}
\ifdefined\qualGap\else\newlength{\qualGap}\fi\setlength{\qualGap}{2pt}
\ifdefined\qualSep\else\newlength{\qualSep}\fi\setlength{\qualSep}{6pt}

\begin{figure*}[t]
\centering
\resizebox{\textwidth}{!}{%
\setlength{\tabcolsep}{0pt}%
\begin{tabular}{
  @{}
  c @{\hspace{\qualSep}} %
  c @{\hspace{\qualSep}}
  c @{\hspace{\qualSep}}
  c 
  @{}
}

$\vcenter{\hbox{\rotatebox{90}{\small Source}}}$ & 
$\vcenter{\hbox{\includegraphics[width=\qualImgW]{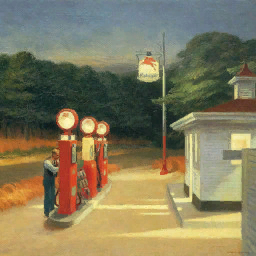}}}$ &
$\vcenter{\hbox{\includegraphics[width=\qualImgW]{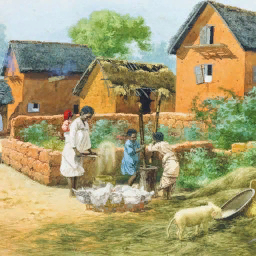}}}$ &
$\vcenter{\hbox{\includegraphics[width=\qualImgW]{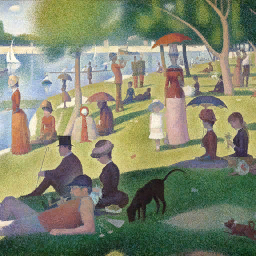}}}$ \\[\qualGap]

$\vcenter{\hbox{\rotatebox{90}{\small Generated-View}}}$ & 
$\vcenter{\hbox{\includegraphics[width=\qualImgW]{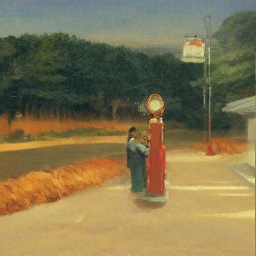}}}$ &
$\vcenter{\hbox{\includegraphics[width=\qualImgW]{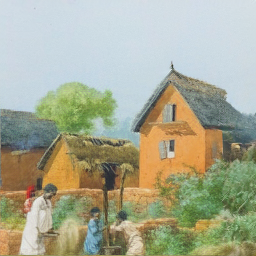}}}$ &
$\vcenter{\hbox{\includegraphics[width=\qualImgW]{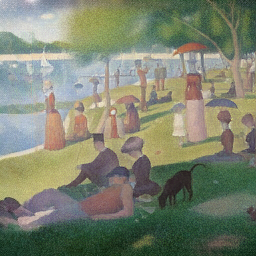}}}$ \\[\qualGap]

\noalign{\smallskip}\hline\noalign{\smallskip}

$\vcenter{\hbox{\rotatebox{90}{\small Source}}}$ & 
$\vcenter{\hbox{\includegraphics[width=\qualImgW]{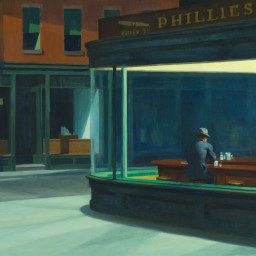}}}$ &
$\vcenter{\hbox{\includegraphics[width=\qualImgW]{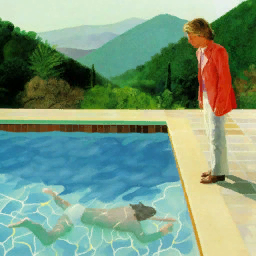}}}$ &
$\vcenter{\hbox{\includegraphics[width=\qualImgW]{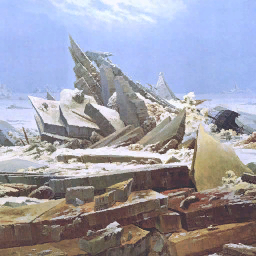}}}$ \\[\qualGap]

$\vcenter{\hbox{\rotatebox{90}{\small Generated-View}}}$ & 
$\vcenter{\hbox{\includegraphics[width=\qualImgW]{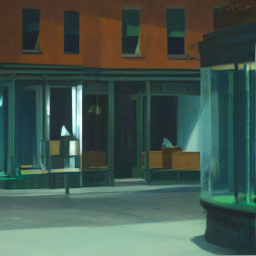}}}$ &
$\vcenter{\hbox{\includegraphics[width=\qualImgW]{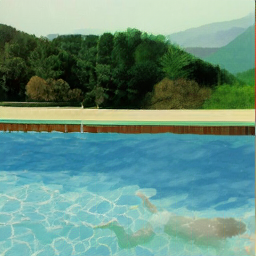}}}$ &
$\vcenter{\hbox{\includegraphics[width=\qualImgW]{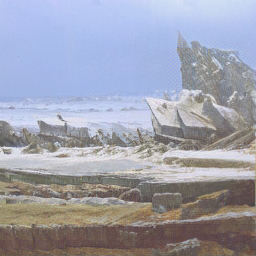}}}$ \\

\end{tabular}%
}%
\caption{\textbf{Qualitative results on out-of-distribution images.} Each pair shows the input source image followed by the generated novel view. Source views are, from left to right and top to bottom: \textit{Gas} by Edward Hopper, \textit{Untitled} by Ralambo, \textit{A Sunday on La Grande Jatte} by Georges Seurat, \textit{Nighthawks} by Edward Hopper, \textit{Portrait of an Artist (Pool with Two Figures)} by David Hockney, and \textit{The Sea of Ice} by Caspar David Friedrich.}
\label{fig:supp_ood}
\end{figure*}

\ifdefined\sotaImgW\else\newlength{\sotaImgW}\fi\setlength{\sotaImgW}{0.325\linewidth}
\ifdefined\sotaGap\else\newlength{\sotaGap}\fi\setlength{\sotaGap}{2pt}

\begin{figure*}[t]
\centering
\setlength{\tabcolsep}{0pt}%
\setlength{\lineskip}{0pt}%
\begin{tabular}{
  @{}
  c @{\hspace{\sotaGap}}
  c @{\hspace{\sotaGap}}
  c
  @{}
}

\makebox[\sotaImgW][c]{\small Source} &
\makebox[\sotaImgW][c]{\small Pseudo-Target} &
\makebox[\sotaImgW][c]{\small Generated View} \\[4pt]

\includegraphics[width=\sotaImgW]{images/supplementary/source_pointcloud_generated/05/source} &
\includegraphics[width=\sotaImgW]{images/supplementary/source_pointcloud_generated/05/target} &
\includegraphics[width=\sotaImgW]{images/supplementary/source_pointcloud_generated/05/gen} \\[\dimexpr\sotaGap-\dp\strutbox\relax]

\includegraphics[width=\sotaImgW]{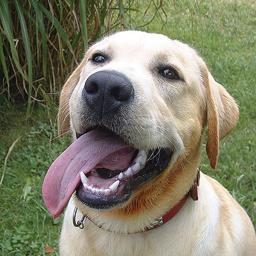} &
\includegraphics[width=\sotaImgW]{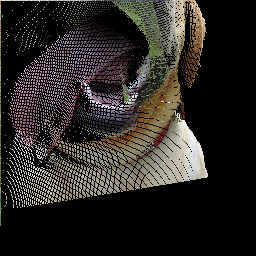} &
\includegraphics[width=\sotaImgW]{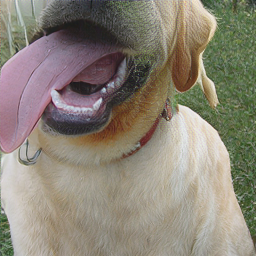} \\[\dimexpr\sotaGap-\dp\strutbox\relax]

\includegraphics[width=\sotaImgW]{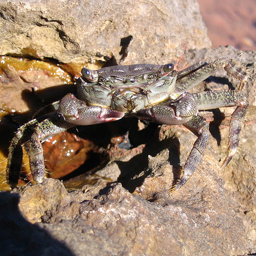} &
\includegraphics[width=\sotaImgW]{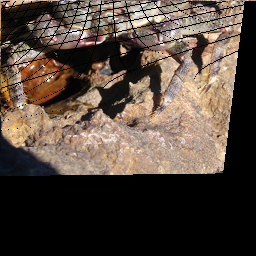} &
\includegraphics[width=\sotaImgW]{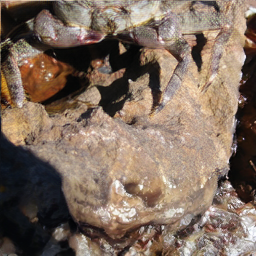} \\[\dimexpr\sotaGap-\dp\strutbox\relax]

\includegraphics[width=\sotaImgW]{images/supplementary/source_pointcloud_generated/08/source} &
\includegraphics[width=\sotaImgW]{images/supplementary/source_pointcloud_generated/08/target} &
\includegraphics[width=\sotaImgW]{images/supplementary/source_pointcloud_generated/08/gen}

\end{tabular}
\caption{\textbf{Comparison of source inputs, training pseudo-targets, and generated views.} During training, OVIE is supervised on pseudo-targets (middle) created by depth-lifting the source image (left) to a sampled pose. At inference, it generates novel views (right) from a source image and target pose. Here, the generated views are rendered at the same poses as their corresponding pseudo-targets.}
\label{fig:source_pc_gen_1}
\end{figure*}

\ifdefined\sotaImgW\else\newlength{\sotaImgW}\fi\setlength{\sotaImgW}{0.325\linewidth}
\ifdefined\sotaGap\else\newlength{\sotaGap}\fi\setlength{\sotaGap}{2pt}

\begin{figure*}[t]
\centering
\setlength{\tabcolsep}{0pt}%
\setlength{\lineskip}{0pt}%
\begin{tabular}{
  @{}
  c @{\hspace{\sotaGap}}
  c @{\hspace{\sotaGap}}
  c
  @{}
}

\makebox[\sotaImgW][c]{\small Source} &
\makebox[\sotaImgW][c]{\small Pseudo-Target} &
\makebox[\sotaImgW][c]{\small Generated View} \\[4pt]

\includegraphics[width=\sotaImgW]{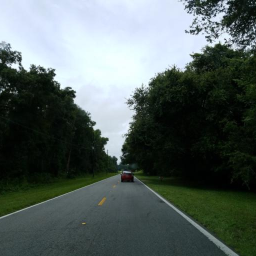} &
\includegraphics[width=\sotaImgW]{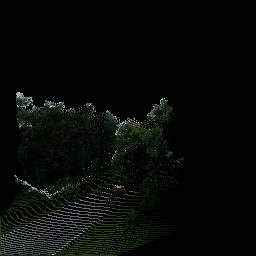} &
\includegraphics[width=\sotaImgW]{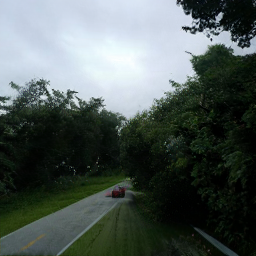} \\[\dimexpr\sotaGap-\dp\strutbox\relax]

\includegraphics[width=\sotaImgW]{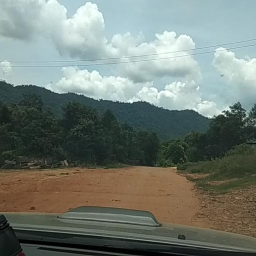} &
\includegraphics[width=\sotaImgW]{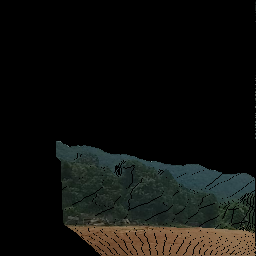} &
\includegraphics[width=\sotaImgW]{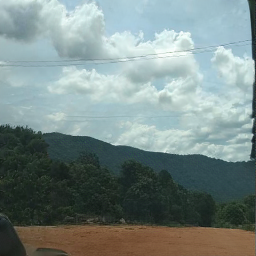} \\[\dimexpr\sotaGap-\dp\strutbox\relax]

\includegraphics[width=\sotaImgW]{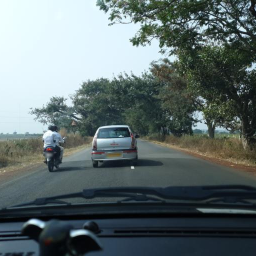} &
\includegraphics[width=\sotaImgW]{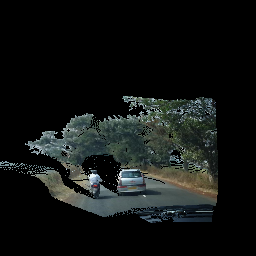} &
\includegraphics[width=\sotaImgW]{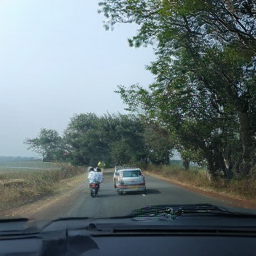} \\[\dimexpr\sotaGap-\dp\strutbox\relax]

\includegraphics[width=\sotaImgW]{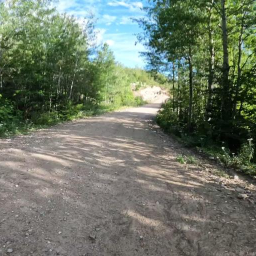} &
\includegraphics[width=\sotaImgW]{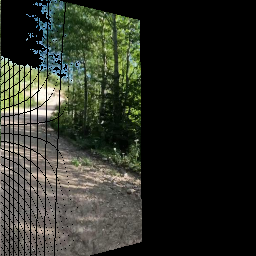} &
\includegraphics[width=\sotaImgW]{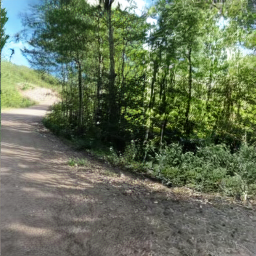}

\end{tabular}
\caption{\textbf{Comparison of source inputs, training pseudo-targets, and generated views.} During training, OVIE is supervised on pseudo-targets (middle) created by depth-lifting the source image (left) to a sampled pose. At inference, it generates novel views (right) from a source image and target pose. Here, the generated views are rendered at the same poses as their corresponding pseudo-targets.}
\label{fig:source_pc_gen_2}
\end{figure*}

\ifdefined\sotaImgW\else\newlength{\sotaImgW}\fi\setlength{\sotaImgW}{0.325\linewidth}
\ifdefined\sotaGap\else\newlength{\sotaGap}\fi\setlength{\sotaGap}{2pt}

\begin{figure*}[t]
\centering
\setlength{\tabcolsep}{0pt}%
\setlength{\lineskip}{0pt}%
\begin{tabular}{
  @{}
  c @{\hspace{\sotaGap}}
  c @{\hspace{\sotaGap}}
  c
  @{}
}

\makebox[\sotaImgW][c]{\small Source} &
\makebox[\sotaImgW][c]{\small Pseudo-Target} &
\makebox[\sotaImgW][c]{\small Generated View} \\[4pt]

\includegraphics[width=\sotaImgW]{images/supplementary/source_pointcloud_generated/21/source} &
\includegraphics[width=\sotaImgW]{images/supplementary/source_pointcloud_generated/21/target} &
\includegraphics[width=\sotaImgW]{images/supplementary/source_pointcloud_generated/21/gen} \\[\dimexpr\sotaGap-\dp\strutbox\relax]

\includegraphics[width=\sotaImgW]{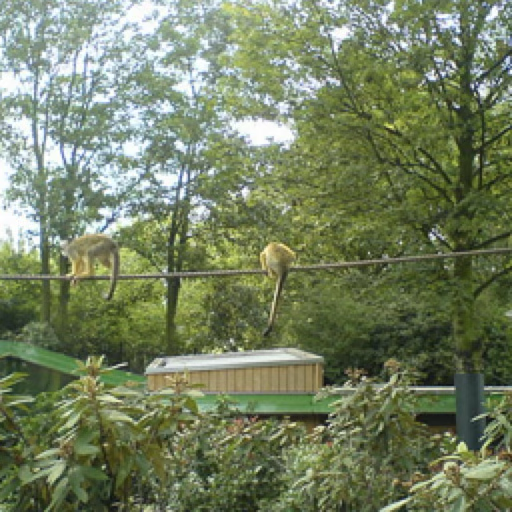} &
\includegraphics[width=\sotaImgW]{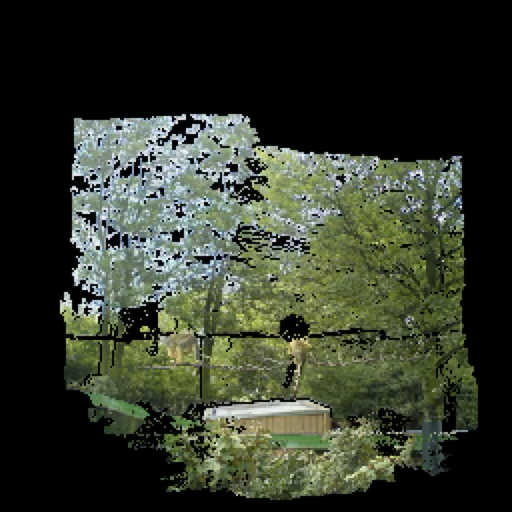} &
\includegraphics[width=\sotaImgW]{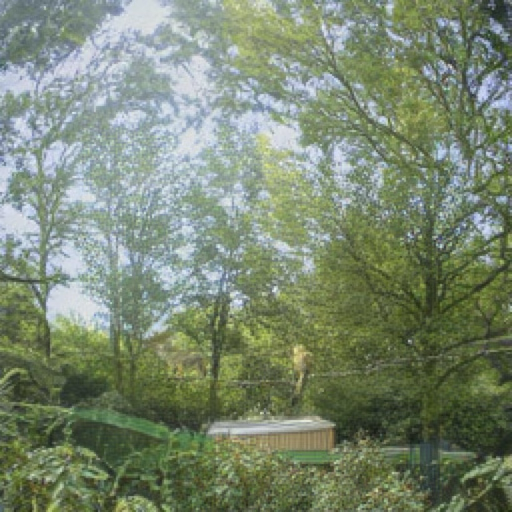} \\[\dimexpr\sotaGap-\dp\strutbox\relax]

\includegraphics[width=\sotaImgW]{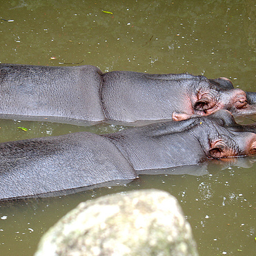} &
\includegraphics[width=\sotaImgW]{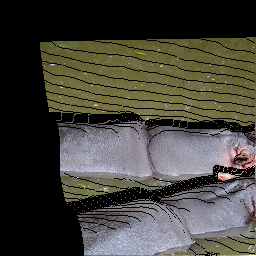} &
\includegraphics[width=\sotaImgW]{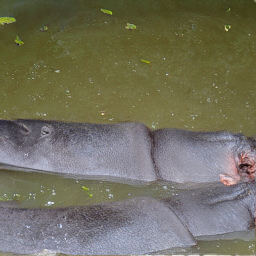} \\[\dimexpr\sotaGap-\dp\strutbox\relax]

\includegraphics[width=\sotaImgW]{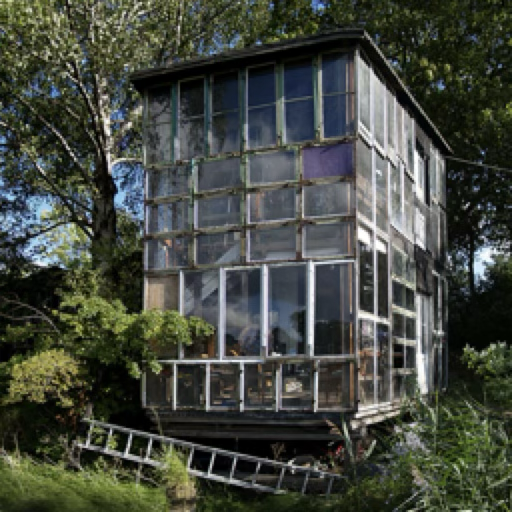} &
\includegraphics[width=\sotaImgW]{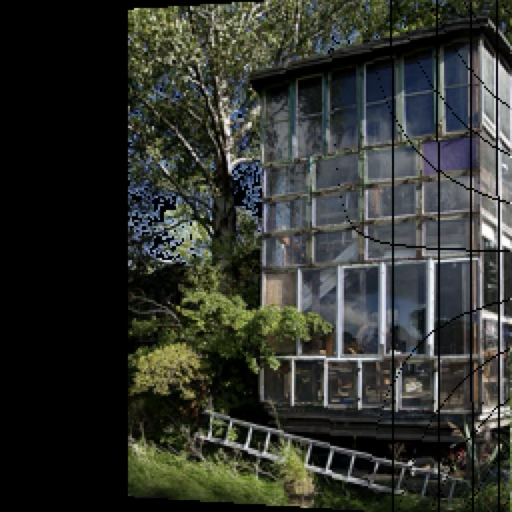} &
\includegraphics[width=\sotaImgW]{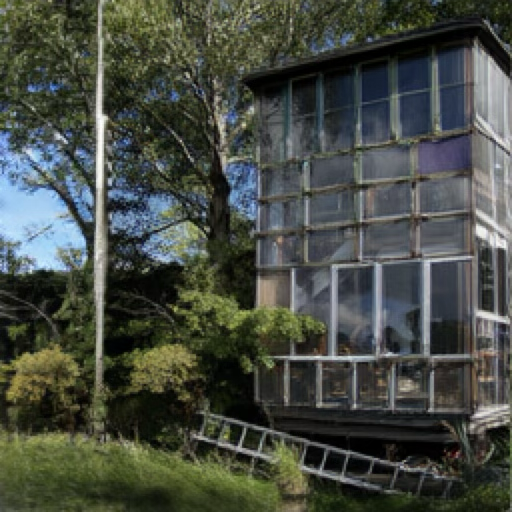}

\end{tabular}
\caption{\textbf{Comparison of source inputs, training pseudo-targets, and generated views.} During training, OVIE is supervised on pseudo-targets (middle) created by depth-lifting the source image (left) to a sampled pose. At inference, it generates novel views (right) from a source image and target pose. Here, the generated views are rendered at the same poses as their corresponding pseudo-targets.}
\label{fig:source_pc_gen_3}
\end{figure*}

\ifdefined\sotaImgW\else\newlength{\sotaImgW}\fi\setlength{\sotaImgW}{0.325\linewidth}
\ifdefined\sotaGap\else\newlength{\sotaGap}\fi\setlength{\sotaGap}{2pt}

\begin{figure*}[t]
\centering
\setlength{\tabcolsep}{0pt}%
\setlength{\lineskip}{0pt}%
\begin{tabular}{
  @{}
  c @{\hspace{\sotaGap}}
  c @{\hspace{\sotaGap}}
  c
  @{}
}

\makebox[\sotaImgW][c]{\small Source} &
\makebox[\sotaImgW][c]{\small Pseudo-Target} &
\makebox[\sotaImgW][c]{\small Generated View} \\[4pt]

\includegraphics[width=\sotaImgW]{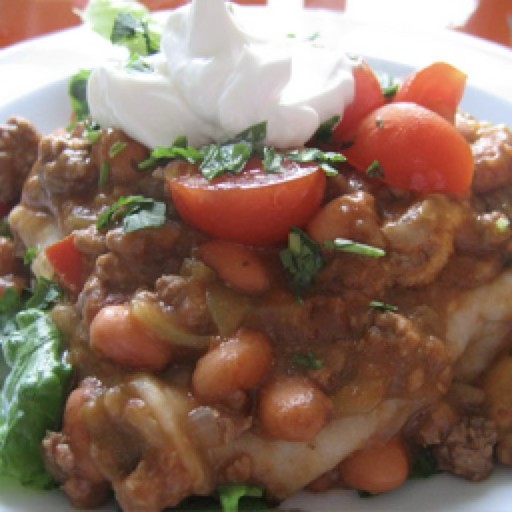} &
\includegraphics[width=\sotaImgW]{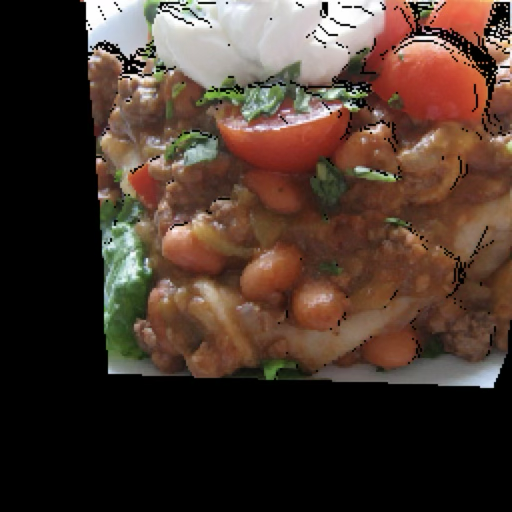} &
\includegraphics[width=\sotaImgW]{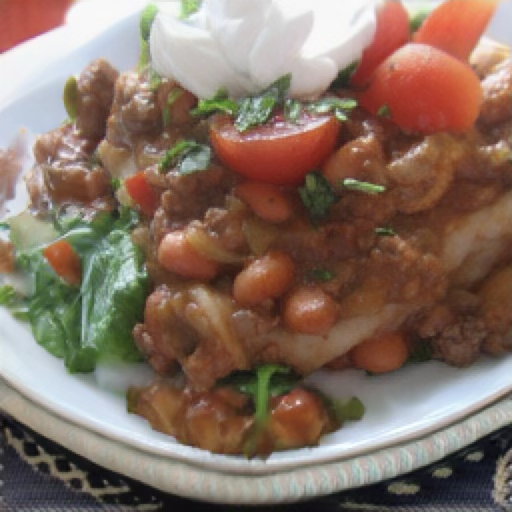} \\[\dimexpr\sotaGap-\dp\strutbox\relax]

\includegraphics[width=\sotaImgW]{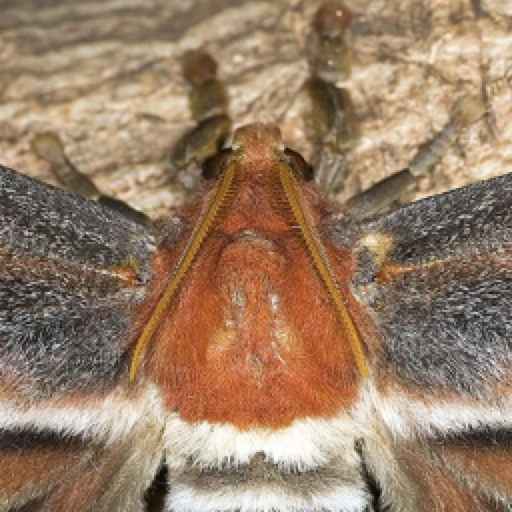} &
\includegraphics[width=\sotaImgW]{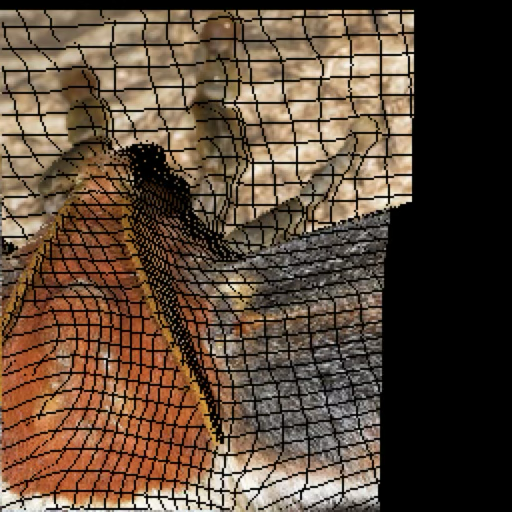} &
\includegraphics[width=\sotaImgW]{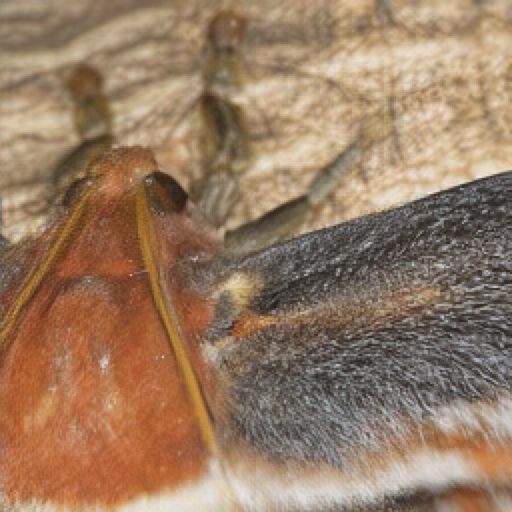} \\[\dimexpr\sotaGap-\dp\strutbox\relax]

\includegraphics[width=\sotaImgW]{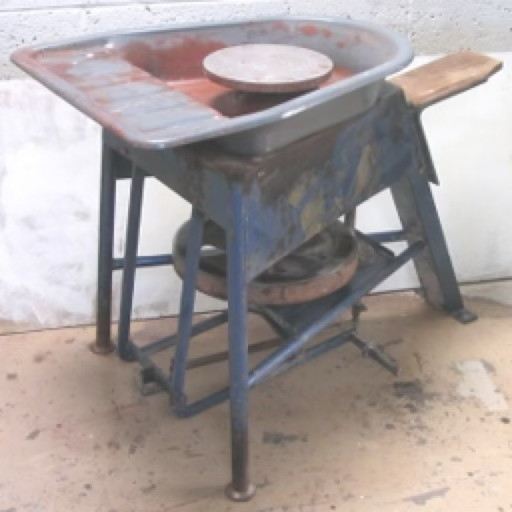} &
\includegraphics[width=\sotaImgW]{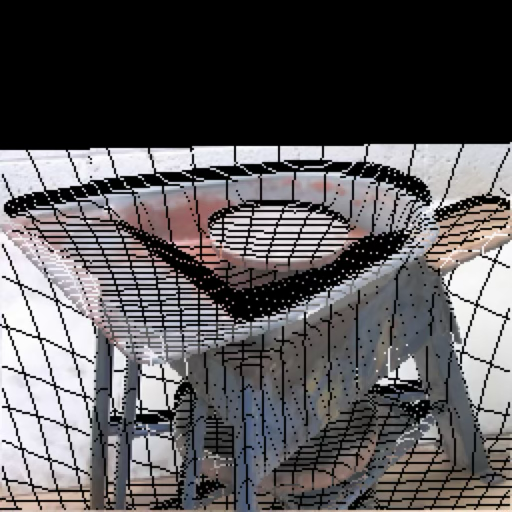} &
\includegraphics[width=\sotaImgW]{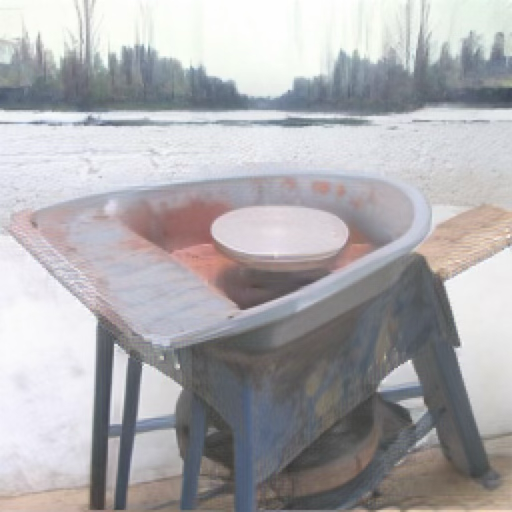} \\[\dimexpr\sotaGap-\dp\strutbox\relax]

\includegraphics[width=\sotaImgW]{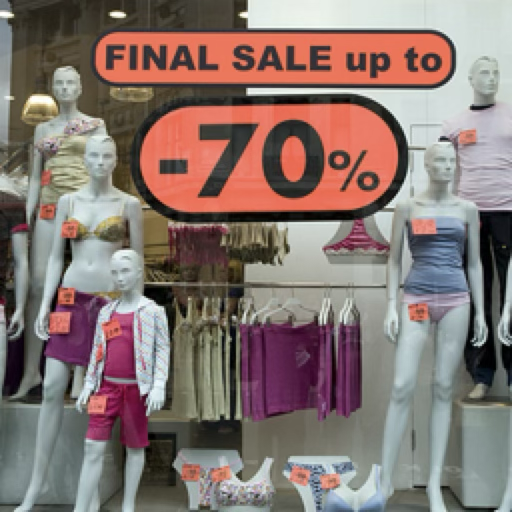} &
\includegraphics[width=\sotaImgW]{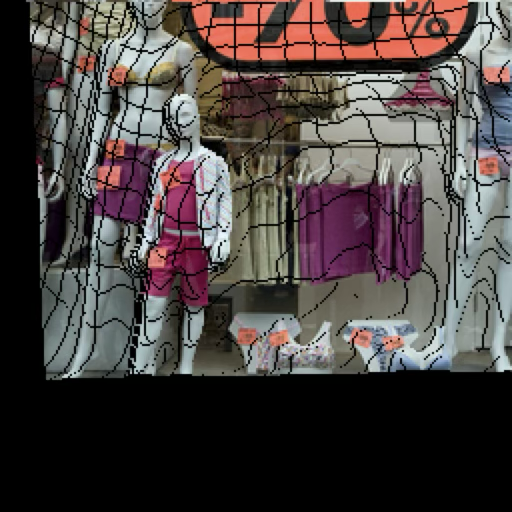} &
\includegraphics[width=\sotaImgW]{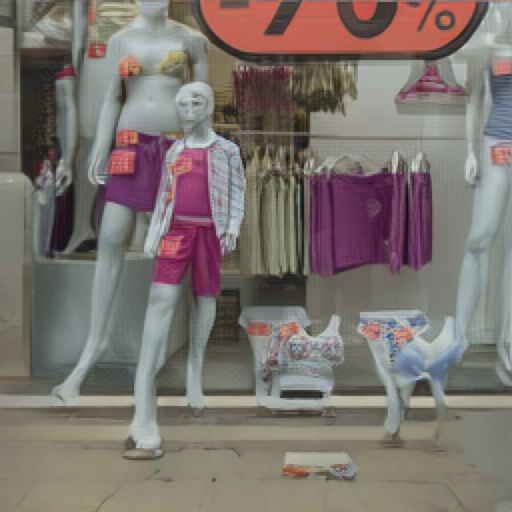}

\end{tabular}
\caption{\textbf{Comparison of source inputs, training pseudo-targets, and generated views.} During training, OVIE is supervised on pseudo-targets (middle) created by depth-lifting the source image (left) to a sampled pose. At inference, it generates novel views (right) from a source image and target pose. Here, the generated views are rendered at the same poses as their corresponding pseudo-targets.}
\label{fig:source_pc_gen_4}
\end{figure*}

\ifdefined\qualImgW\else\newlength{\qualImgW}\fi\setlength{\qualImgW}{1.8cm}
\ifdefined\qualGap\else\newlength{\qualGap}\fi\setlength{\qualGap}{1pt}
\ifdefined\qualSep\else\newlength{\qualSep}\fi\setlength{\qualSep}{3pt}

\begin{figure*}[t]
\centering
\resizebox{\textwidth}{!}{%
\setlength{\tabcolsep}{0pt}%
\begin{tabular}{
  @{}
  c @{\hspace{\qualGap}}
  c
  @{\hspace{\qualSep}}!{\color{black!25}\vrule width 0.4pt}@{\hspace{\qualSep}}
  c @{\hspace{\qualGap}}
  c @{\hspace{\qualGap}}
  c
  @{\hspace{\qualSep}}!{\color{black!25}\vrule width 0.4pt}@{\hspace{\qualSep}}
  c @{\hspace{\qualGap}}
  c
  @{}
}

\small Source &
\small Target &
\small GeoGPT &
\small PhotoNVS &
\small VIVID &
\small \mname (Ours)\\[\qualGap]

\includegraphics[width=\qualImgW]{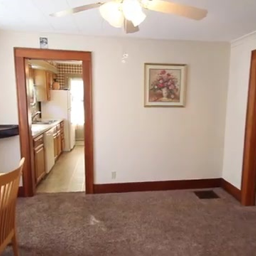} &
\includegraphics[width=\qualImgW]{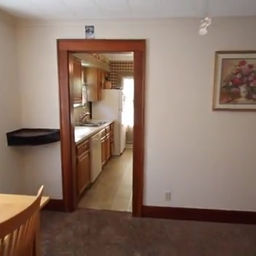} &
\includegraphics[width=\qualImgW]{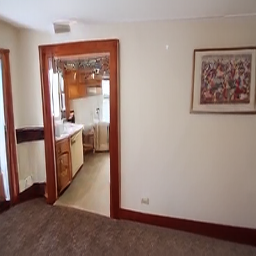} &
\includegraphics[width=\qualImgW]{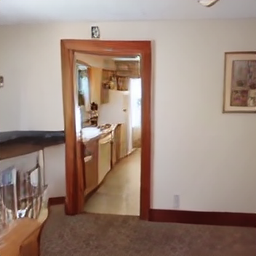} &
\includegraphics[width=\qualImgW]{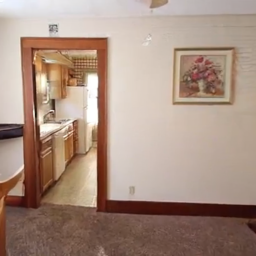} &
\includegraphics[width=\qualImgW]{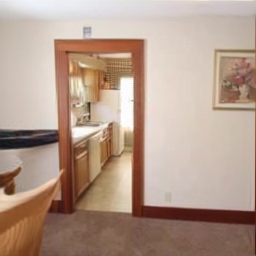} \\[\qualGap]

\includegraphics[width=\qualImgW]{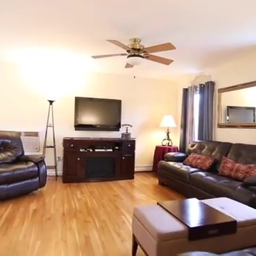} &
\includegraphics[width=\qualImgW]{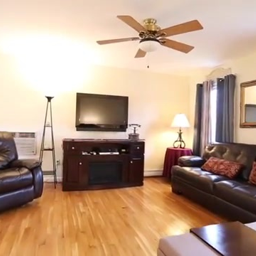} &
\includegraphics[width=\qualImgW]{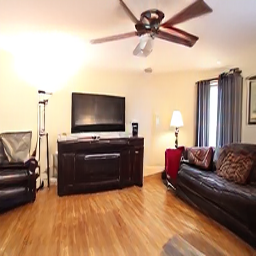} &
\includegraphics[width=\qualImgW]{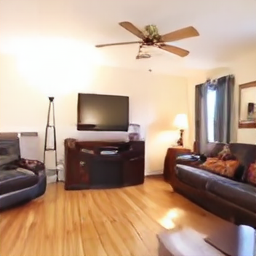} &
\includegraphics[width=\qualImgW]{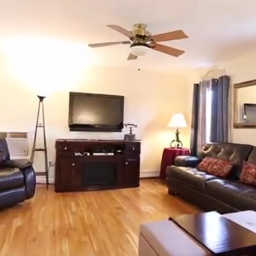} &
\includegraphics[width=\qualImgW]{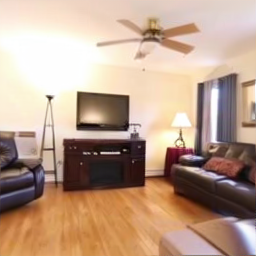} \\[\qualGap]

\includegraphics[width=\qualImgW]{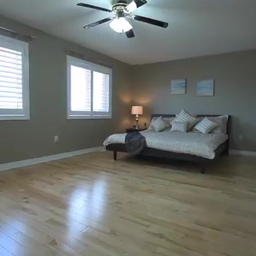} &
\includegraphics[width=\qualImgW]{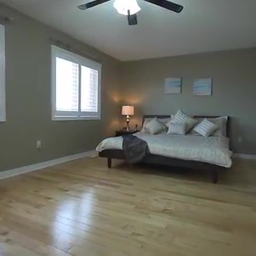} &
\includegraphics[width=\qualImgW]{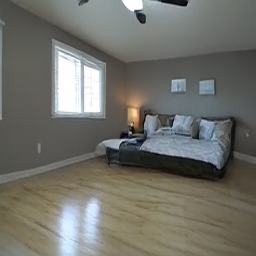} &
\includegraphics[width=\qualImgW]{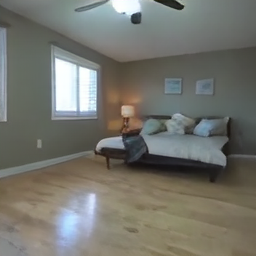} &
\includegraphics[width=\qualImgW]{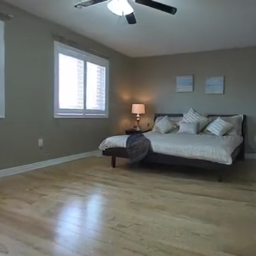} &
\includegraphics[width=\qualImgW]{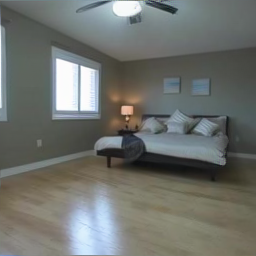} \\[\qualGap]

\includegraphics[width=\qualImgW]{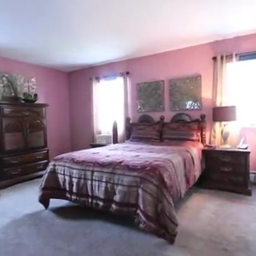} &
\includegraphics[width=\qualImgW]{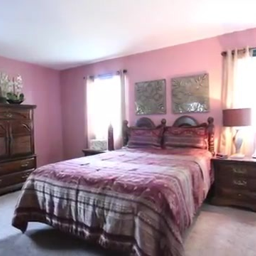} &
\includegraphics[width=\qualImgW]{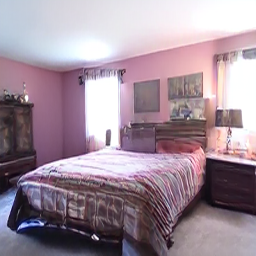} &
\includegraphics[width=\qualImgW]{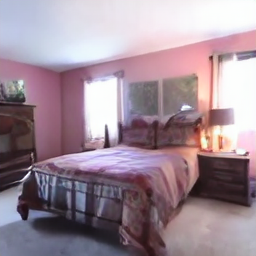} &
\includegraphics[width=\qualImgW]{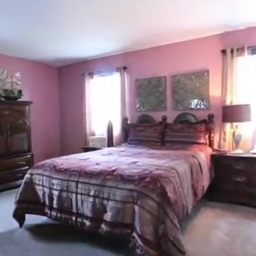} &
\includegraphics[width=\qualImgW]{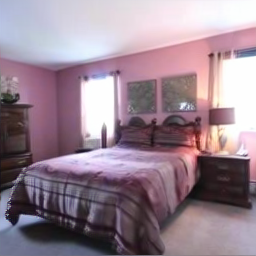} \\[\qualGap]

\includegraphics[width=\qualImgW]{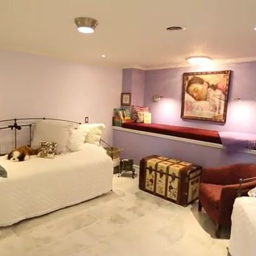} &
\includegraphics[width=\qualImgW]{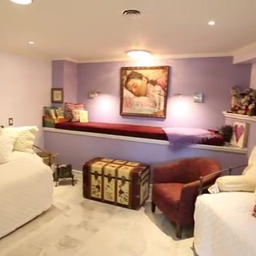} &
\includegraphics[width=\qualImgW]{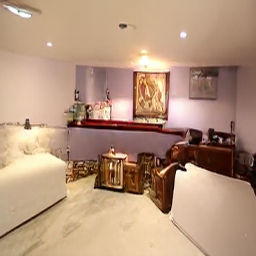} &
\includegraphics[width=\qualImgW]{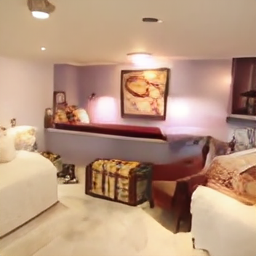} &
\includegraphics[width=\qualImgW]{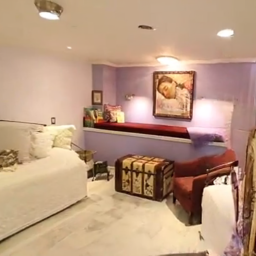} &
\includegraphics[width=\qualImgW]{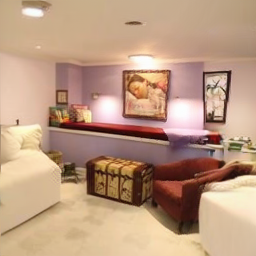} \\[\qualGap]

\includegraphics[width=\qualImgW]{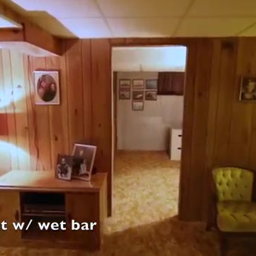} &
\includegraphics[width=\qualImgW]{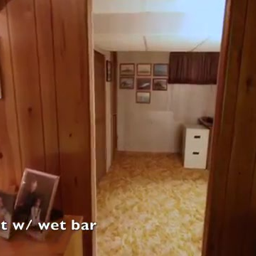} &
\includegraphics[width=\qualImgW]{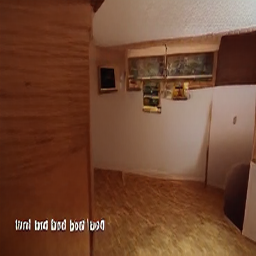} &
\includegraphics[width=\qualImgW]{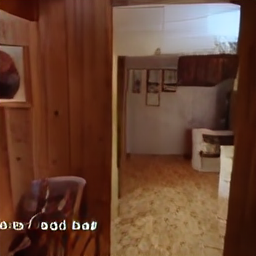} &
\includegraphics[width=\qualImgW]{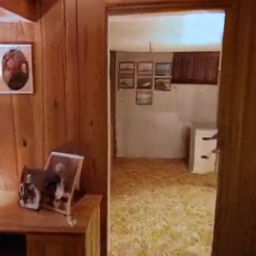} &
\includegraphics[width=\qualImgW]{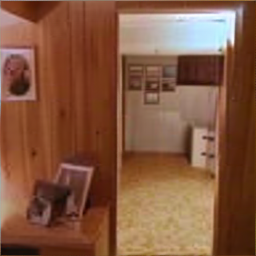} \\[\qualGap]

\includegraphics[width=\qualImgW]{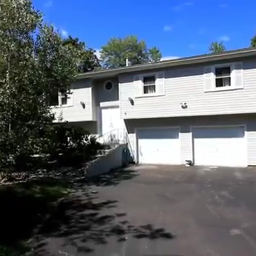} &
\includegraphics[width=\qualImgW]{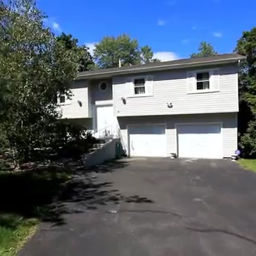} &
\includegraphics[width=\qualImgW]{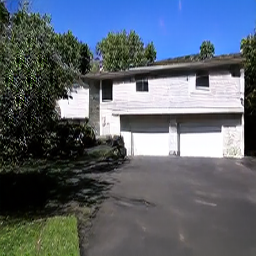} &
\includegraphics[width=\qualImgW]{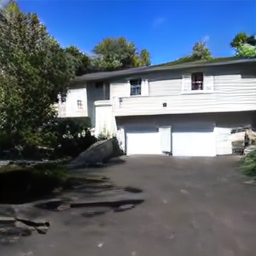} &
\includegraphics[width=\qualImgW]{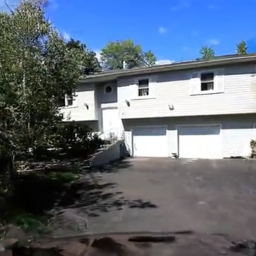} &
\includegraphics[width=\qualImgW]{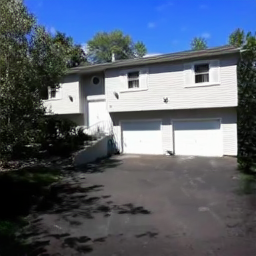} \\[\qualGap]

\includegraphics[width=\qualImgW]{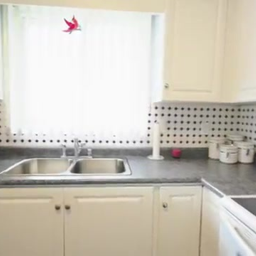} &
\includegraphics[width=\qualImgW]{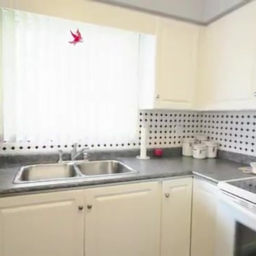} &
\includegraphics[width=\qualImgW]{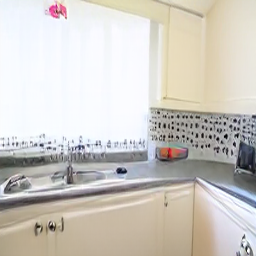} &
\includegraphics[width=\qualImgW]{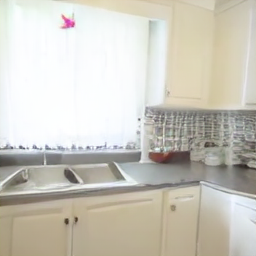} &
\includegraphics[width=\qualImgW]{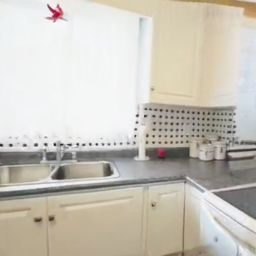} &
\includegraphics[width=\qualImgW]{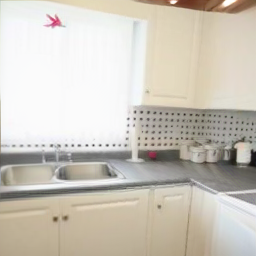}

\end{tabular}%
}%
\caption{\textbf{Qualitative comparison with state-of-the-art methods.}
Given a source image and target camera pose, each method synthesizes a novel view. Despite training on no multi-view data, \mname generates novel views that match or exceed the quality of concurrent methods.}
\label{fig:sup_comparison_1}
\end{figure*}

\ifdefined\qualImgW\else\newlength{\qualImgW}\fi\setlength{\qualImgW}{1.8cm}
\ifdefined\qualGap\else\newlength{\qualGap}\fi\setlength{\qualGap}{1pt}
\ifdefined\qualSep\else\newlength{\qualSep}\fi\setlength{\qualSep}{3pt}

\begin{figure*}[t]
\centering
\resizebox{\textwidth}{!}{%
\setlength{\tabcolsep}{0pt}%
\begin{tabular}{
  @{}
  c @{\hspace{\qualGap}}
  c
  @{\hspace{\qualSep}}!{\color{black!25}\vrule width 0.4pt}@{\hspace{\qualSep}}
  c @{\hspace{\qualGap}}
  c @{\hspace{\qualGap}}
  c
  @{\hspace{\qualSep}}!{\color{black!25}\vrule width 0.4pt}@{\hspace{\qualSep}}
  c @{\hspace{\qualGap}}
  c
  @{}
}

\small Source &
\small Target &
\small GeoGPT &
\small PhotoNVS &
\small VIVID &
\small \mname (Ours)\\[\qualGap]

\includegraphics[width=\qualImgW]{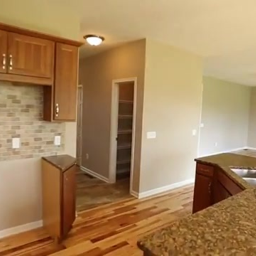} &
\includegraphics[width=\qualImgW]{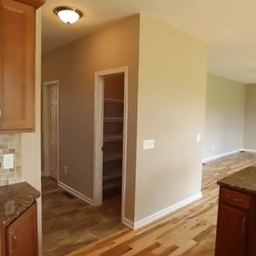} &
\includegraphics[width=\qualImgW]{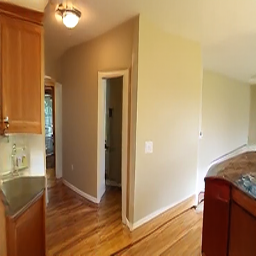} &
\includegraphics[width=\qualImgW]{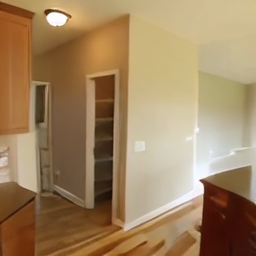} &
\includegraphics[width=\qualImgW]{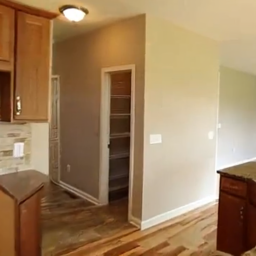} &
\includegraphics[width=\qualImgW]{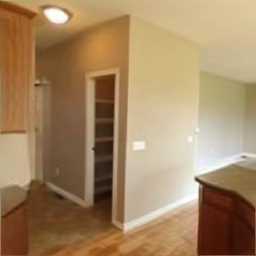} \\[\qualGap]

\includegraphics[width=\qualImgW]{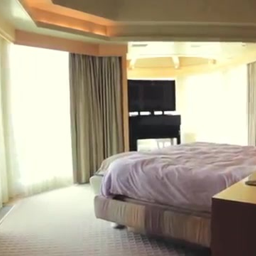} &
\includegraphics[width=\qualImgW]{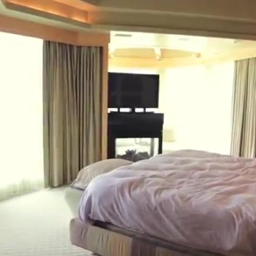} &
\includegraphics[width=\qualImgW]{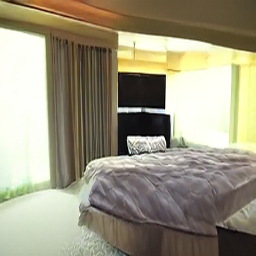} &
\includegraphics[width=\qualImgW]{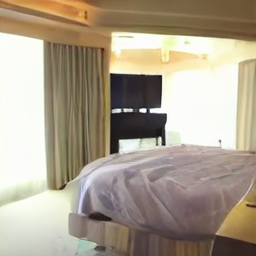} &
\includegraphics[width=\qualImgW]{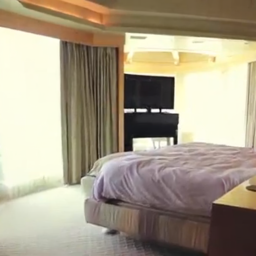} &
\includegraphics[width=\qualImgW]{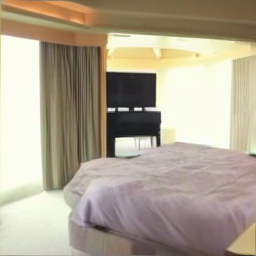} \\[\qualGap]

\includegraphics[width=\qualImgW]{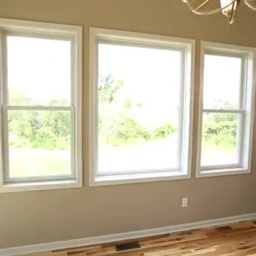} &
\includegraphics[width=\qualImgW]{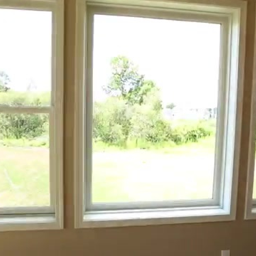} &
\includegraphics[width=\qualImgW]{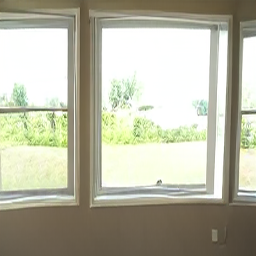} &
\includegraphics[width=\qualImgW]{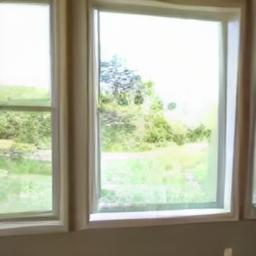} &
\includegraphics[width=\qualImgW]{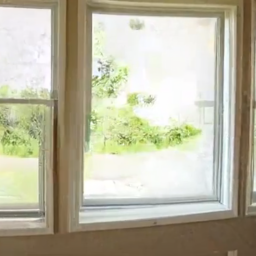} &
\includegraphics[width=\qualImgW]{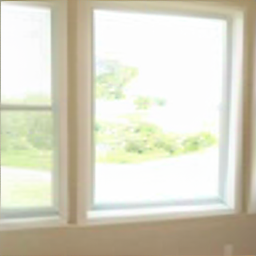} \\[\qualGap]

\includegraphics[width=\qualImgW]{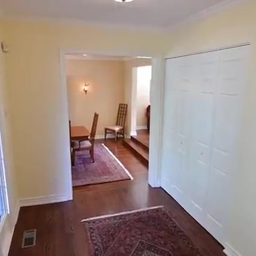} &
\includegraphics[width=\qualImgW]{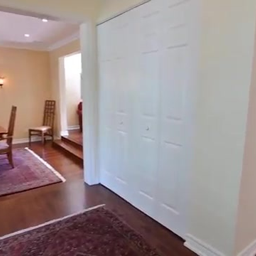} &
\includegraphics[width=\qualImgW]{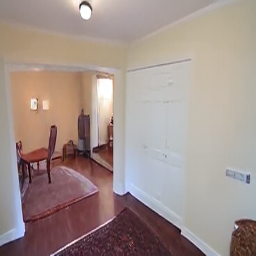} &
\includegraphics[width=\qualImgW]{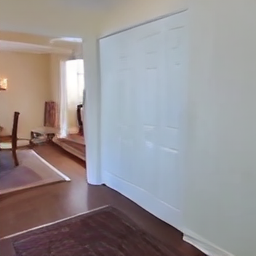} &
\includegraphics[width=\qualImgW]{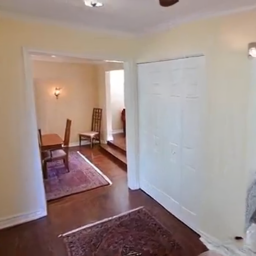} &
\includegraphics[width=\qualImgW]{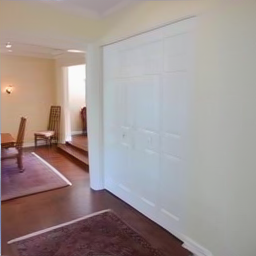} \\[\qualGap]

\includegraphics[width=\qualImgW]{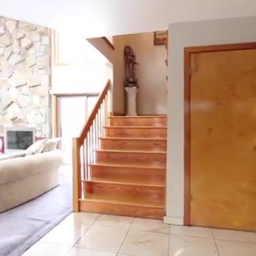} &
\includegraphics[width=\qualImgW]{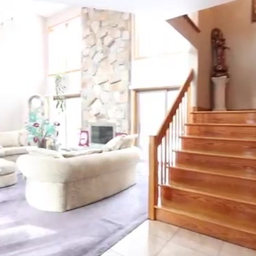} &
\includegraphics[width=\qualImgW]{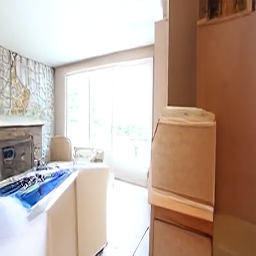} &
\includegraphics[width=\qualImgW]{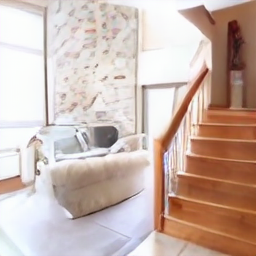} &
\includegraphics[width=\qualImgW]{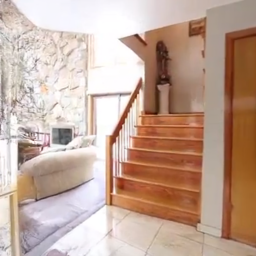} &
\includegraphics[width=\qualImgW]{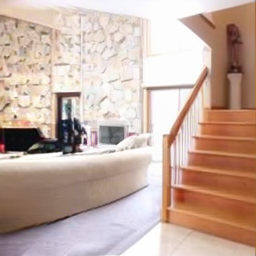} \\[\qualGap]

\includegraphics[width=\qualImgW]{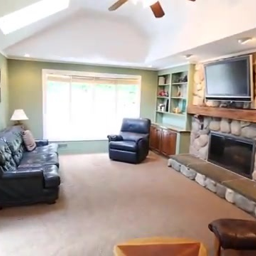} &
\includegraphics[width=\qualImgW]{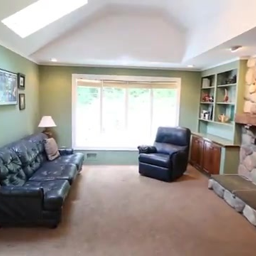} &
\includegraphics[width=\qualImgW]{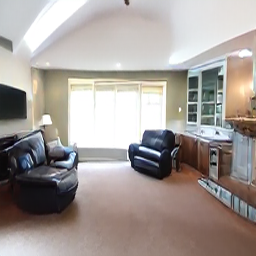} &
\includegraphics[width=\qualImgW]{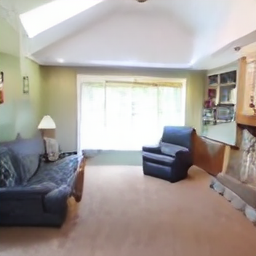} &
\includegraphics[width=\qualImgW]{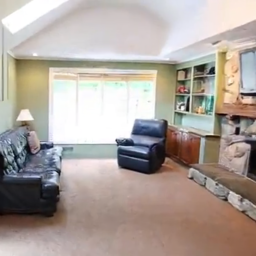} &
\includegraphics[width=\qualImgW]{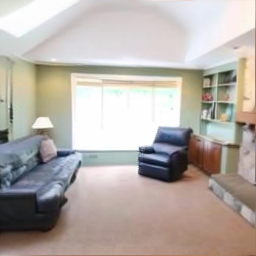} \\[\qualGap]

\includegraphics[width=\qualImgW]{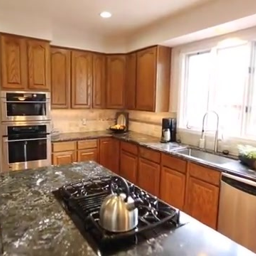} &
\includegraphics[width=\qualImgW]{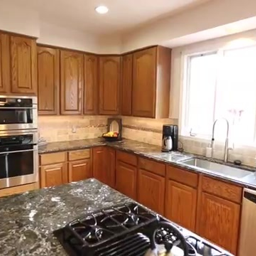} &
\includegraphics[width=\qualImgW]{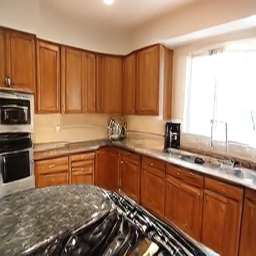} &
\includegraphics[width=\qualImgW]{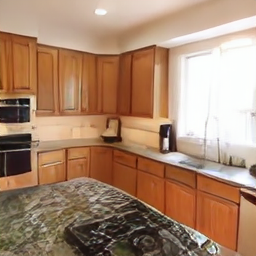} &
\includegraphics[width=\qualImgW]{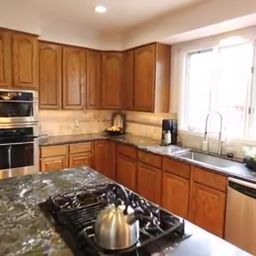} &
\includegraphics[width=\qualImgW]{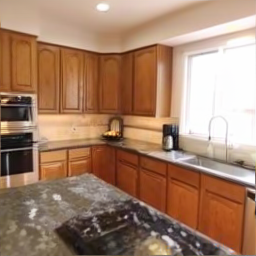} \\[\qualGap]

\includegraphics[width=\qualImgW]{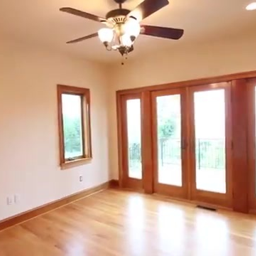} &
\includegraphics[width=\qualImgW]{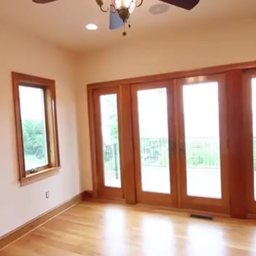} &
\includegraphics[width=\qualImgW]{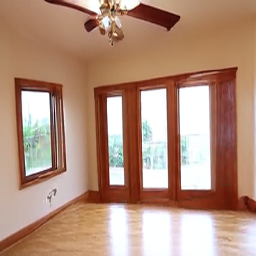} &
\includegraphics[width=\qualImgW]{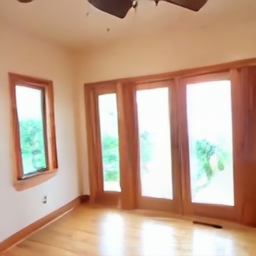} &
\includegraphics[width=\qualImgW]{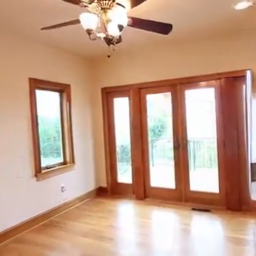} &
\includegraphics[width=\qualImgW]{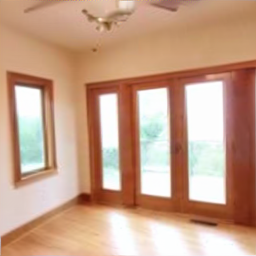}

\end{tabular}%
}%
\caption{\textbf{Qualitative comparison with state-of-the-art methods.}
Given a source image and target camera pose, each method synthesizes a novel view. Despite training on no multi-view data, \mname generates novel views that match or exceed the quality of concurrent methods.}
\label{fig:sup_comparison_2}
\end{figure*}

\ifdefined\qualImgW\else\newlength{\qualImgW}\fi\setlength{\qualImgW}{1.8cm}
\ifdefined\qualGap\else\newlength{\qualGap}\fi\setlength{\qualGap}{1pt}
\ifdefined\qualSep\else\newlength{\qualSep}\fi\setlength{\qualSep}{3pt}

\begin{figure*}[t]
\centering
\resizebox{\textwidth}{!}{%
\setlength{\tabcolsep}{0pt}%
\begin{tabular}{
  @{}
  c @{\hspace{\qualGap}}
  c
  @{\hspace{\qualSep}}!{\color{black!25}\vrule width 0.4pt}@{\hspace{\qualSep}}
  c @{\hspace{\qualGap}}
  c @{\hspace{\qualGap}}
  c
  @{\hspace{\qualSep}}!{\color{black!25}\vrule width 0.4pt}@{\hspace{\qualSep}}
  c @{\hspace{\qualGap}}
  c
  @{}
}

\small Source &
\small Target &
\small GeoGPT &
\small PhotoNVS &
\small VIVID &
\small \mname (Ours)\\[\qualGap]

\includegraphics[width=\qualImgW]{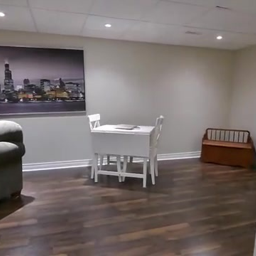} &
\includegraphics[width=\qualImgW]{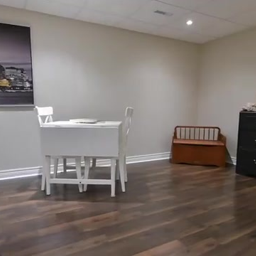} &
\includegraphics[width=\qualImgW]{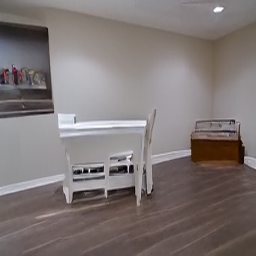} &
\includegraphics[width=\qualImgW]{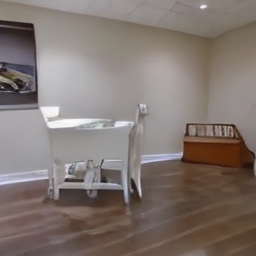} &
\includegraphics[width=\qualImgW]{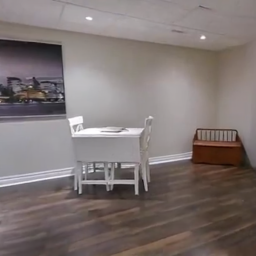} &
\includegraphics[width=\qualImgW]{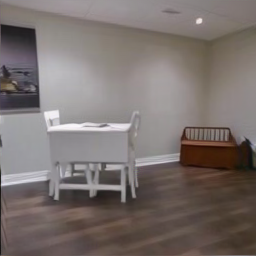} \\[\qualGap]

\includegraphics[width=\qualImgW]{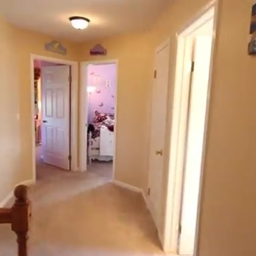} &
\includegraphics[width=\qualImgW]{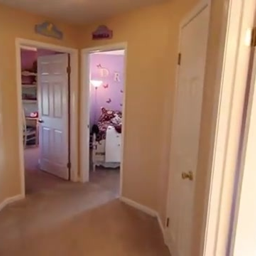} &
\includegraphics[width=\qualImgW]{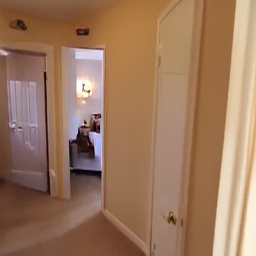} &
\includegraphics[width=\qualImgW]{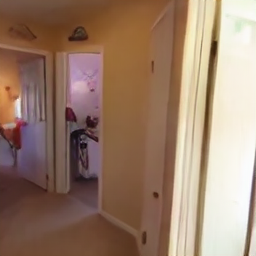} &
\includegraphics[width=\qualImgW]{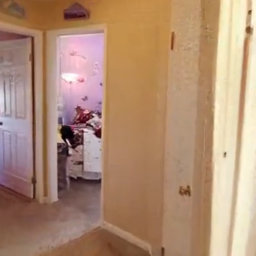} &
\includegraphics[width=\qualImgW]{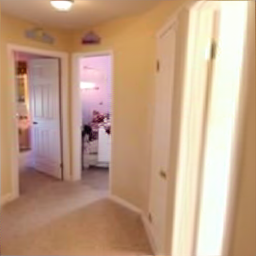} \\[\qualGap]

\includegraphics[width=\qualImgW]{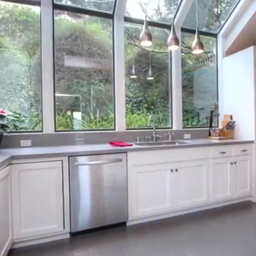} &
\includegraphics[width=\qualImgW]{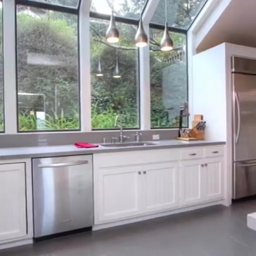} &
\includegraphics[width=\qualImgW]{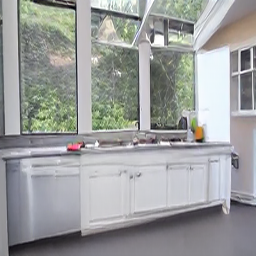} &
\includegraphics[width=\qualImgW]{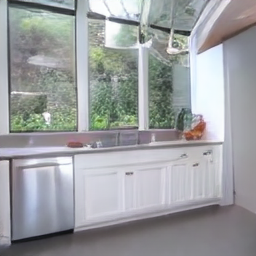} &
\includegraphics[width=\qualImgW]{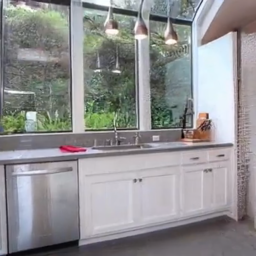} &
\includegraphics[width=\qualImgW]{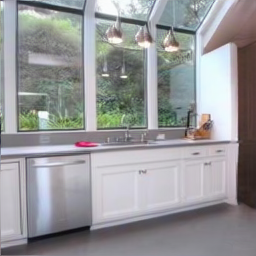} \\[\qualGap]

\includegraphics[width=\qualImgW]{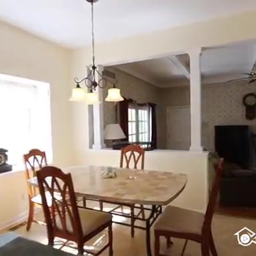} &
\includegraphics[width=\qualImgW]{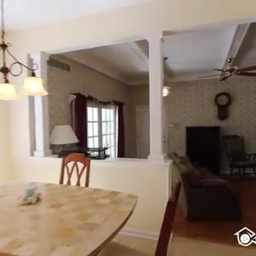} &
\includegraphics[width=\qualImgW]{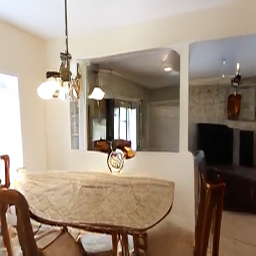} &
\includegraphics[width=\qualImgW]{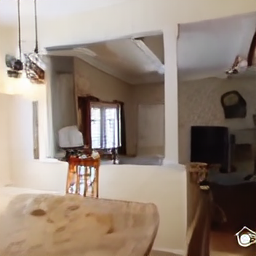} &
\includegraphics[width=\qualImgW]{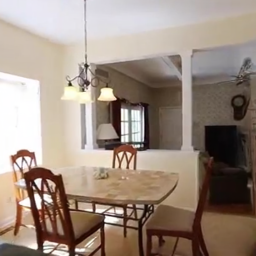} &
\includegraphics[width=\qualImgW]{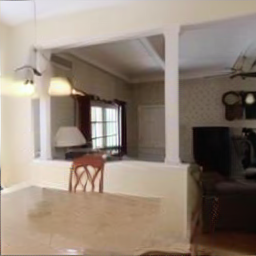} \\[\qualGap]

\includegraphics[width=\qualImgW]{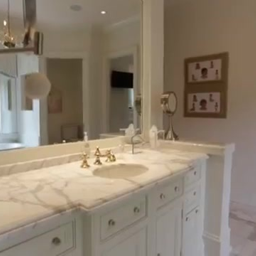} &
\includegraphics[width=\qualImgW]{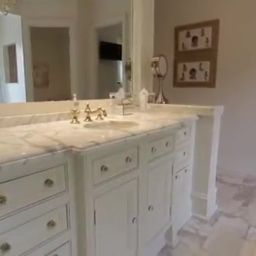} &
\includegraphics[width=\qualImgW]{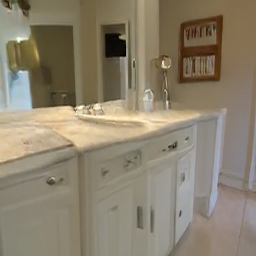} &
\includegraphics[width=\qualImgW]{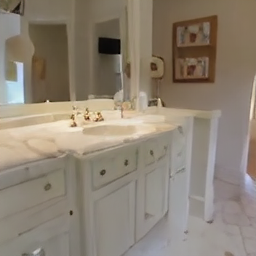} &
\includegraphics[width=\qualImgW]{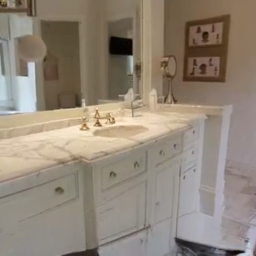} &
\includegraphics[width=\qualImgW]{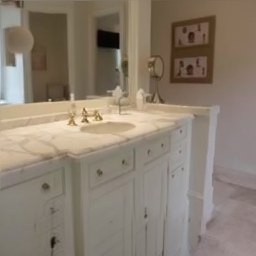} \\[\qualGap]

\includegraphics[width=\qualImgW]{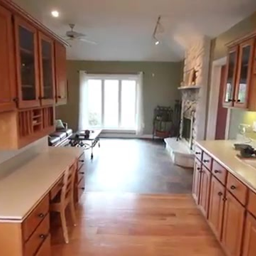} &
\includegraphics[width=\qualImgW]{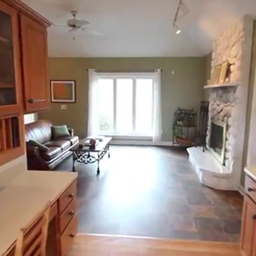} &
\includegraphics[width=\qualImgW]{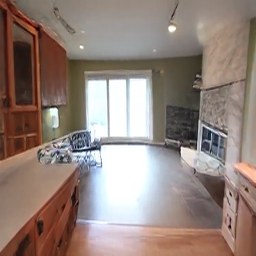} &
\includegraphics[width=\qualImgW]{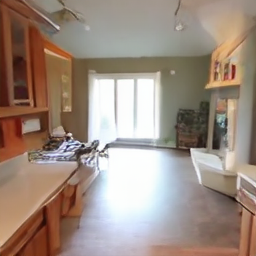} &
\includegraphics[width=\qualImgW]{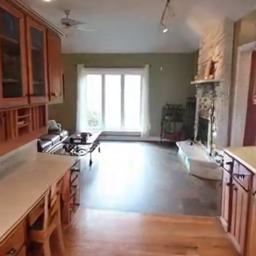} &
\includegraphics[width=\qualImgW]{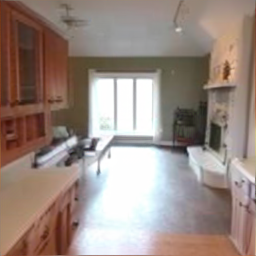} \\[\qualGap]

\includegraphics[width=\qualImgW]{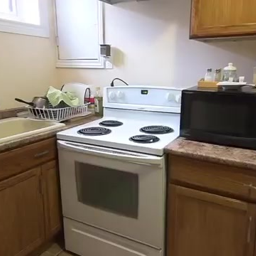} &
\includegraphics[width=\qualImgW]{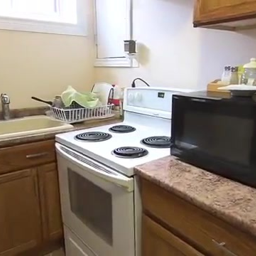} &
\includegraphics[width=\qualImgW]{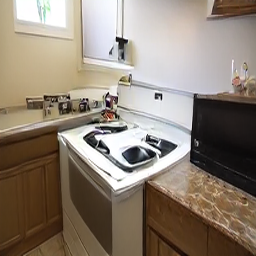} &
\includegraphics[width=\qualImgW]{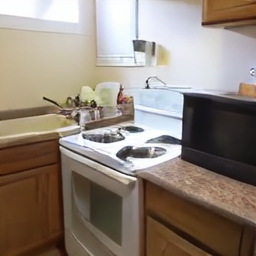} &
\includegraphics[width=\qualImgW]{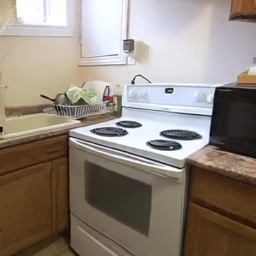} &
\includegraphics[width=\qualImgW]{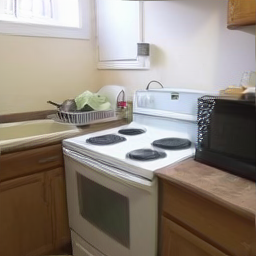} \\[\qualGap]

\includegraphics[width=\qualImgW]{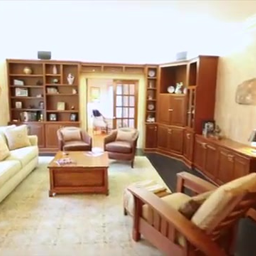} &
\includegraphics[width=\qualImgW]{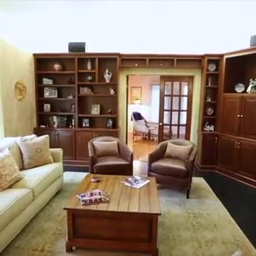} &
\includegraphics[width=\qualImgW]{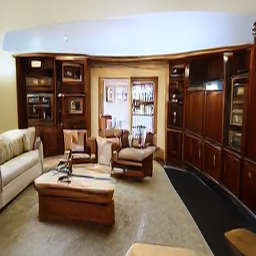} &
\includegraphics[width=\qualImgW]{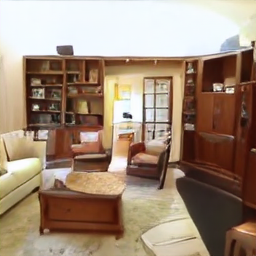} &
\includegraphics[width=\qualImgW]{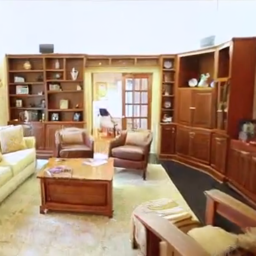} &
\includegraphics[width=\qualImgW]{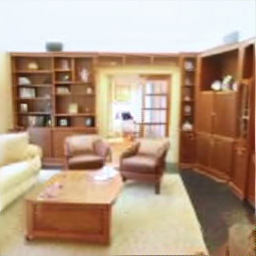}

\end{tabular}%
}%
\caption{\textbf{Qualitative comparison with state-of-the-art methods.}
Given a source image and target camera pose, each method synthesizes a novel view. Despite training on no multi-view data, \mname generates novel views that match or exceed the quality of concurrent methods.}
\label{fig:sup_comparison_3}
\end{figure*}

\section{Additional Comparison to InfiniteNature-Zero}
\label{sec:appendix_infinitenaturezero}

InfiniteNature-Zero~\cite{infinitenaturezero} learns perpetual view generation from 
single-view image collections, using a geometry-based render-refine-repeat pipeline 
that relies on monocular depth estimation and sky segmentation at inference. Like OVIE, 
it requires no posed multi-view data during training, making it a conceptually relevant 
comparison despite its fundamentally different, geometry-dependent approach.

To ensure a fair comparison, we use the public checkpoint of InfiniteNature-Zero and adapt its inference to our single-step 
evaluation protocol, generating each target view independently from the source frame 
rather than autoregressively. This grants the model access to the ground-truth source 
image and its depth estimate at every step, an inherent advantage over OVIE and other
geometry-free methods.

Despite this advantage, InfiniteNature-Zero's generation quality lags behind 
OVIE's. On RealEstate10K, it achieves an LPIPS of 0.395 and FID of 28.8, compared to 
OVIE's 0.279 and 6.74. On DL3DV, it scores an LPIPS of 0.472 and FID of 45.7 versus 
OVIE's 0.464 and 13.6, the highest FID among all compared methods on both benchmarks. 
As shown in Figure~\ref{fig:infinitenaturezero}, InfiniteNature-Zero 
produces blurry outputs, which explains its poor LPIPS and FID scores. Notably, its 
PSNR and SSIM remain comparable to OVIE's (19.4 and 0.642 on RealEstate10K versus 18.8 
and 0.602), a discrepancy attributable to the insensitivity of these metrics to 
perceptual sharpness, as discussed in \Cref{subsec:ablation}.

\ifdefined\sotaImgW\else\newlength{\sotaImgW}\fi\setlength{\sotaImgW}{0.24\linewidth}
\ifdefined\sotaGap\else\newlength{\sotaGap}\fi\setlength{\sotaGap}{2pt}
\begin{figure*}[t]
\centering
\setlength{\tabcolsep}{0pt}%
\setlength{\lineskip}{0pt}%
\begin{tabular}{
  @{}
  c @{\hspace{\sotaGap}}
  c @{\hspace{\sotaGap}}
  c @{\hspace{\sotaGap}}
  c
  @{}
}
\makebox[\sotaImgW][c]{\small Source} &
\makebox[\sotaImgW][c]{\small Ground-Truth} &
\makebox[\sotaImgW][c]{\small OVIE (ours)} &
\makebox[\sotaImgW][c]{\small InfiniteNature-Zero~\cite{infinitenaturezero}} \\[4pt]
\includegraphics[width=\sotaImgW]{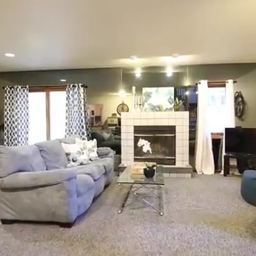} &
\includegraphics[width=\sotaImgW]{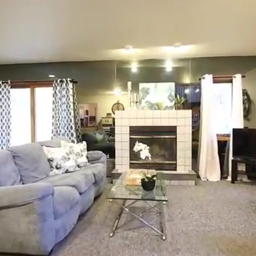} &
\includegraphics[width=\sotaImgW]{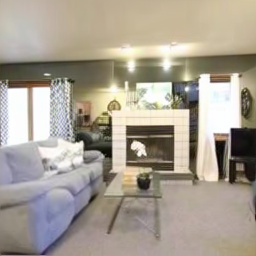} &
\includegraphics[width=\sotaImgW]{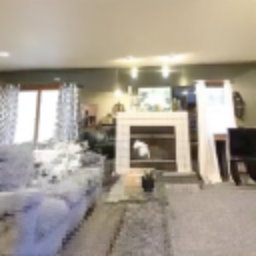} \\[\dimexpr\sotaGap-\dp\strutbox\relax]
\includegraphics[width=\sotaImgW]{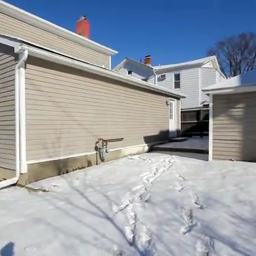} &
\includegraphics[width=\sotaImgW]{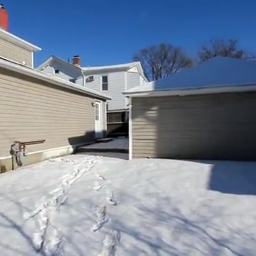} &
\includegraphics[width=\sotaImgW]{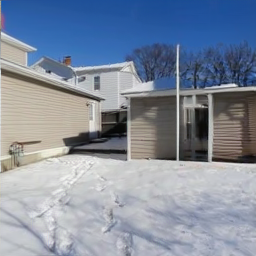} &
\includegraphics[width=\sotaImgW]{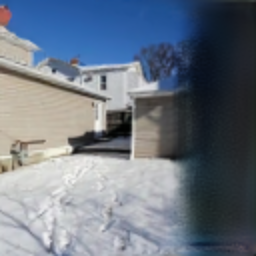} \\[\dimexpr\sotaGap-\dp\strutbox\relax]
\includegraphics[width=\sotaImgW]{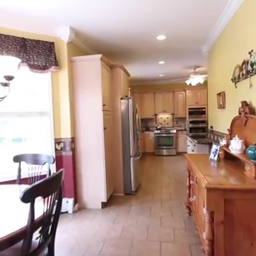} &
\includegraphics[width=\sotaImgW]{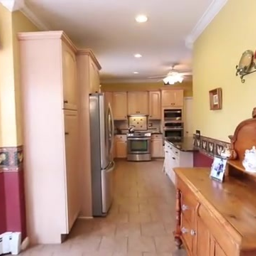} &
\includegraphics[width=\sotaImgW]{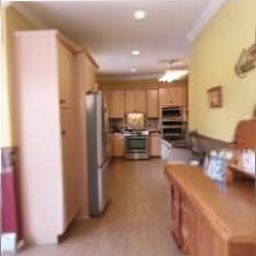} &
\includegraphics[width=\sotaImgW]{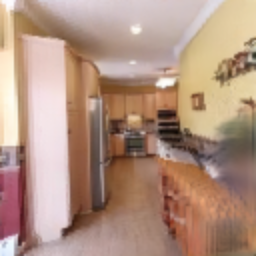} \\[\dimexpr\sotaGap-\dp\strutbox\relax]
\includegraphics[width=\sotaImgW]{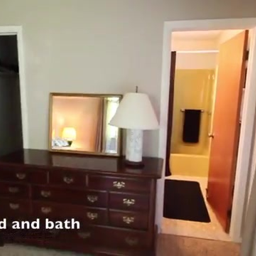} &
\includegraphics[width=\sotaImgW]{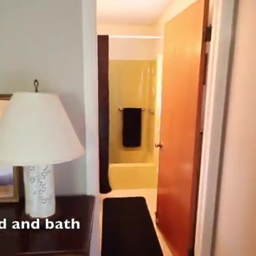} &
\includegraphics[width=\sotaImgW]{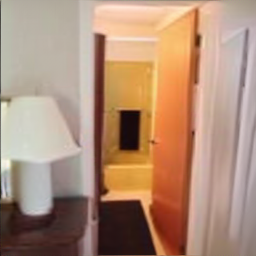} &
\includegraphics[width=\sotaImgW]{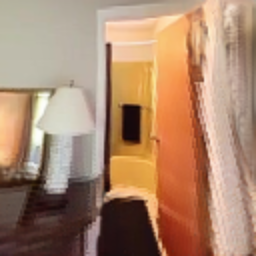}
\end{tabular}
\caption{\textbf{Qualitative comparison with InfiniteNature-Zero~\cite{infinitenaturezero}.} Given a source image (left), we show the ground-truth target view, the view generated by OVIE (ours), and the view generated by InfiniteNature-Zero. OVIE's generated views are sharper and more realistic than those of InfiniteNature-Zero.}
\label{fig:infinitenaturezero}
\end{figure*}

\end{document}